\documentclass[letterpaper]{article} 
\usepackage{pretty_aaai}  
\usepackage{times}  
\usepackage{helvet}  
\usepackage{courier}  
\usepackage[hyphens]{url}  
\usepackage{graphicx} 
\usepackage{natbib}  
\usepackage{caption} 
\usepackage{algorithm}
\usepackage{algorithmic}

\usepackage{newfloat}
\usepackage{listings}
\DeclareCaptionStyle{ruled}{labelfont=normalfont,labelsep=colon,strut=off} 
\floatstyle{ruled}
\newfloat{listing}{tb}{lst}{}
\floatname{listing}{Listing}
\usepackage{amsmath}
\usepackage{booktabs}
\usepackage{floatflt}
\usepackage{subcaption}
\usepackage{hyperref} 
\usepackage{cleveref}
\usepackage{dblfloatfix}
\usepackage{xspace}
\usepackage{url}            

\title{Adaptive Computation Modules: Granular Conditional Computation \\ for Efficient Inference}
\author{
    Bartosz Wójcik\textsuperscript{\rm 1, \rm 2}, Alessio Devoto\textsuperscript{\rm 3}, Karol Pustelnik\textsuperscript{\rm 4}, Pasquale Minervini\textsuperscript{\rm 5, \rm 6}, Simone Scardapane\textsuperscript{\rm 3}
}
\affiliations{
    \textsuperscript{\rm 1}IDEAS NCBR \\
    \textsuperscript{\rm 2}Jagiellonian University \\ 
    \textsuperscript{\rm 3}Sapienza University of Rome \\ \textsuperscript{\rm 4}University of Warsaw \\
    \textsuperscript{\rm 5}University of Edinburgh \\
    \textsuperscript{\rm 6}Miniml.AI\\

}

\begin{document}

\maketitle

\begin{abstract}
While transformer models have been highly successful, they are computationally inefficient.
We observe that for each layer, the full width of the layer may be needed only for a small subset of tokens inside a batch and that the ``effective" width needed to process a token can vary from layer to layer.
Motivated by this observation, we introduce the \emph{Adaptive Computation Module} (ACM), a generic module that dynamically adapts its computational load to match the estimated difficulty of the input on a per-token basis.
An ACM consists of a sequence of \emph{learners} that progressively refine the output of their preceding counterparts. An additional gating mechanism determines the optimal number of learners to execute for each token.
We also propose a distillation technique to replace any pre-trained model with an ``ACMized" variant. 
Our evaluation of transformer models in computer vision and speech recognition demonstrates that substituting layers with ACMs significantly reduces inference costs without degrading the downstream accuracy for a wide interval of user-defined budgets.
\end{abstract}

\section{Introduction}
\label{sec:intro}

Driven by their constantly improving capabilities, state-of-the-art neural networks have been experiencing a continued growth in size in the last decade \citep{pugliese2021machine}. This progress was made possible by algorithmic and model design improvements (in particular with the introduction of transformer models) and the increasing computational power of modern GPUs.
Scaling up the model architecture frequently results in improved performance \citep{devlin2018bert,zagoruyko2016wide}, causing the size and computational costs of state-of-the-art models to keep increasing steadily. These escalating costs limit applications in latency-sensitive and energy-constrained scenarios and contribute to higher carbon emissions, exacerbating environmental concerns~\cite{DBLP:journals/cacm/SchwartzDSE20,patterson2022carbon}. 

Several approaches from the literature aim to mitigate this problem.
For example, quantization reduces inference time by quantizing the weights and activations into low-precision floating-point \citep{courbariaux2014training} or integer values \citep{wu2020integer,dettmers2022llm}. Knowledge distillation methods \citep{hinton2015distilling,aguilar2020knowledge} can transfer knowledge from an ensemble or single larger model into a smaller network. Finally, sparsification methods \citep{lecun1989optimal,han2015learning,hoefler2021sparsity} yield either sparse weight or sparse activation tensors. 

\begin{figure}
    \centering
    \begin{subfigure}{.98\linewidth}
        \centering
        \begin{subfigure}[t]{\linewidth}
            \includegraphics[width=0.875\linewidth]{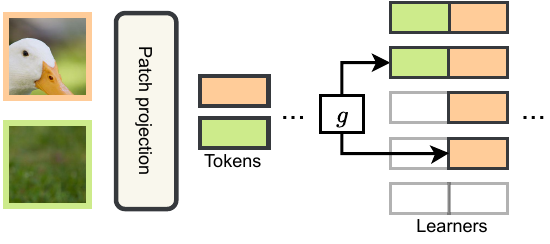}
        \end{subfigure}
        \label{fig:acm_teaser_concept}
        \caption{High-level overview of the Adaptive Computation Module}
    \end{subfigure}
    \begin{subfigure}{.98\linewidth}
        \centering
        \begin{subfigure}[t]{0.425\linewidth}
            \includegraphics[width=\linewidth]{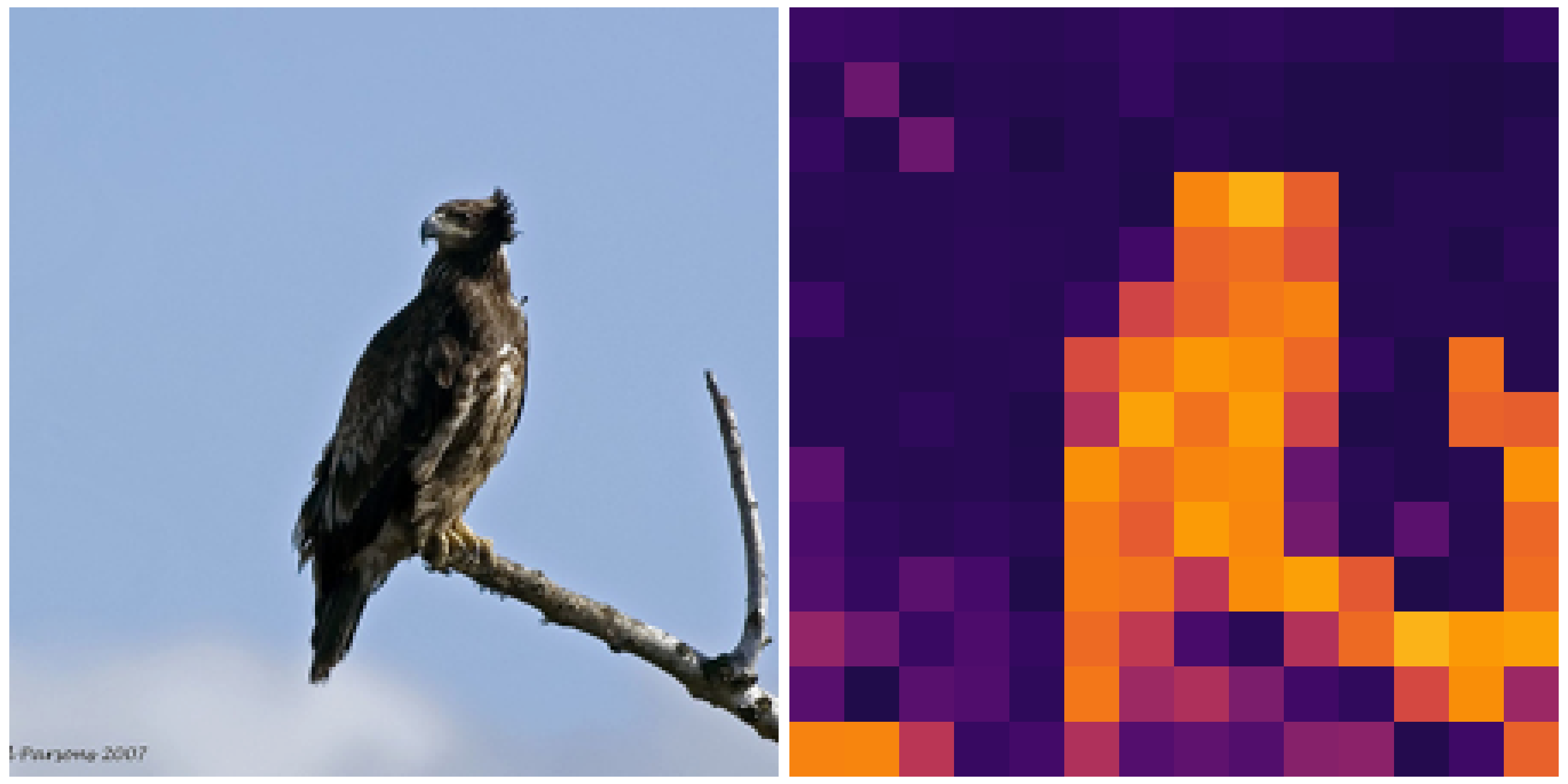}
        \end{subfigure}
        \begin{subfigure}[t]{0.425\linewidth}
            \includegraphics[width=\linewidth]{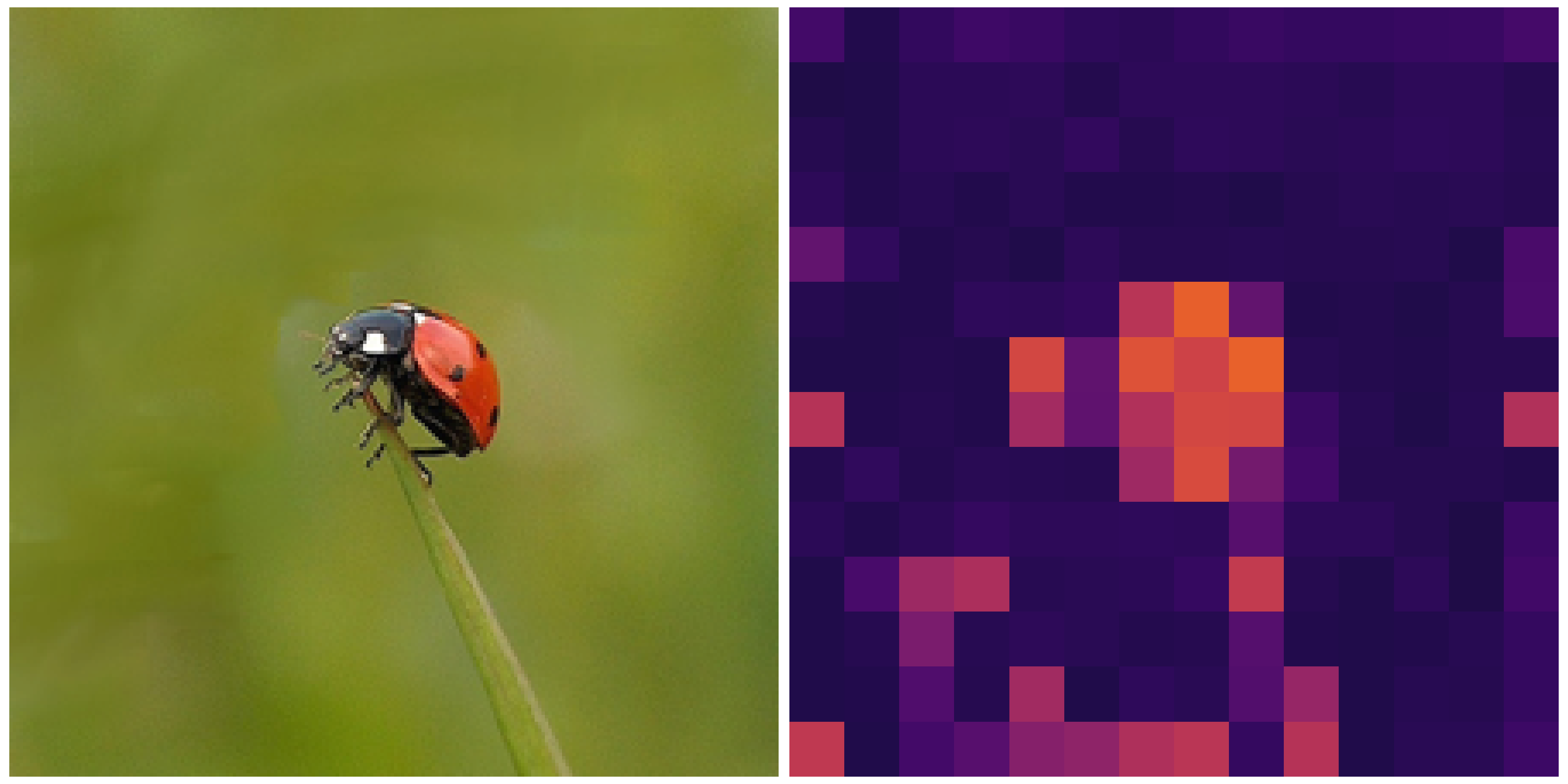}
        \end{subfigure}
        \label{fig:spatial_adaptability_teaser}
        \caption{Example images with computation spatial load maps}
    \end{subfigure}
    \centering
    \caption{ACMs adapt their computational load for each input on a per-token basis by selecting the number of \emph{learners} to execute via a trainable gate.
    In the example on the top, a background token (green) is allocated fewer learners than a content-rich token (orange). This results in a spatially varying computational load, as shown on the bottom.
    }
    \label{fig:acm_teaser}
\end{figure}

While they incur a slight decrease in downstream model performance, such methods show that some neural modules can often be replaced with computationally cheaper alternatives. Based on this observation, we argue that in transformer models~\cite{vaswani2017attention}, the representational capacity of a whole layer is not needed for every input token.
In other words, we hypothesize that the same transformation could be achieved in a more computationally efficient manner.
To explore this hypothesis, we introduce \emph{Adaptive Computation Modules} (ACMs), a neural network module that adapts its computational burden to match the difficulty of the current input token.
An ACM consists of a sequence of \emph{learners} and a single gating network. The task of each learner is to improve upon the combined output of previous learners, while the gating network determines the optimal number of learners to execute for each input token.
Since all token-level decisions are independent, the resulting model is a highly granular application of the conditional computation paradigm \citep{bengio2013deep,bengio2015conditional}, and thus allows for spatial adaptability in transformer models \citep{figurnov2017spatially,han2021dynamic}. 
\Cref{fig:acm_teaser} provides an outline of the proposed model.
To enable the use of a vast range of publicly available models, we propose a straightforward conversion procedure in which we: (1) substitute the selected blocks of the model with ACMs, (2) initialize the learners by distilling knowledge from the substituted blocks, (3) pre-train the gating networks using artificially generated labels, and (4) train the entire model in an end-to-end manner.

Our approach can significantly decrease the model's computational footprint without sacrificing performance.
We evaluate our method on the ImageNet-1k~\cite{ILSVRC15} dataset with Vision Transformer (ViT) models and on speech recognition with pre-trained Wav2Vec networks \cite{baevski2020wav2vec}, and show that in both cases we achieve a better performance-efficiency trade-off than other conditional computation methods.
We make the following contributions:
\begin{itemize}
    \item We show that individual layers in transformer models can be computationally inefficient and that their entire expressiveness is often needed only for a small subset of input tokens.
    
    \item We introduce ACM, an easy-to-train, modular, and general-purpose module that offers a granular approach for reducing the computational cost of transformers.
    
    \item We propose a smart initialization strategy for the parameters of ACMs that relies on module-wise knowledge distillation from pre-trained models.
    
    \item We provide an efficient GPU implementation to demonstrate that ACMs effectively speed up inference.
\end{itemize}

\section{Related Work}

We provide a brief overview of related works, focusing on our experimental comparison. A longer overview with a broader outlook can be found in the supplementary material. 

\subsection{Conditional Computation}
Conditional computation (CC) refers to the ability of a model to adapt its computational graph to its input. While CC can be used for problems such as continual learning \citep{lin2019conditional}, most CC methods adjust their execution on a per-input basis to significantly reduce their average computational cost while maintaining performance \citep{scardapane_condcomp}. 

In the following, we focus on CC algorithms that can be applied to any pre-trained transformer model with minimal modifications; architecture-specific CC algorithms (e.g., dynamic channel selection for CNNs \citep{chen2019you,li2021dynamic}) are discussed in the supplementary material. We broadly categorize the methods into three groups: early-exits (EEs) \citep{teerapittayanon2016branchynet,bolukbasi2017adaptive}, mixture-of-experts (MoEs) \citep{yuksel2012twenty,shazeer2017outrageously,fedus2022review}, and token dropping (TD) \citep{rao2021dynamicvit,yin2022vit,meng2022adavit,haurum2023tokens}. Finally, \citet{caiflextron} is a concurrent conditional computation work that is remarkably similar to the proposed ACMs. In particular, it also trains a router for each layer and converts a pre-trained dense model with a three-step method.

\subsection{Early-exits} In EE models, inputs are allowed to ``exit'' the architecture at intermediate layers via additional classifier heads that are trained together with the backbone network \citep{teerapittayanon2016branchynet}, or as a separate phase in a layerwise fashion \citep{han2021dynamic}, and the predictions of multiple heads can be merged with a number of different techniques \citep{teerapittayanon2016branchynet,wolczyk2021zero,scardapane2020differentiable}. In EE networks, the execution is stopped for the entire input at once, as opposed to individual tokens as is done in the proposed ACMs. In addition, designing and placing exit heads is itself a non-trivial problem \citep{teerapittayanon2016branchynet}, and very little work has been done outside standard computer vision and natural language processing classification tasks. In contrast, ACMs are designed to replace individual blocks in a transformer model in a plug-and-play fashion.

\subsection{Token Dropping} TD strategies either prematurely stop execution \citep{rao2021dynamicvit,yin2022vit,haurum2023tokens} or skip computation of selected layers \citep{meng2022adavit} for \textit{individual tokens} that are deemed less relevant or redundant. We can interpret them as dynamically allocating a variable depth to each token. Our proposed ACM can be seen as an orthogonal \textit{width} variant of token dropping, in which each token is allocated a variable \textit{width} for each layer in the network.

\subsection{Mixture-of-Experts} In MoEs, selected modules of a model are replaced by an independent set of blocks called experts, which are selectively activated for each token by an additional routing network \citep{puigcerver2023sparse}. MoEs have been successfully developed for replacing MLP layers in transformer models \citep{shazeer2017outrageously,riquelme2021scaling}, attention layers \citep{zhang2022mixture}, entire blocks \cite{tan2023sparse}, and adapters \citep{zadouri2023pushing}. Although MoEs are typically trained from scratch or fine-tuned from existing models \citep{zadouri2023pushing}, a small number of works have investigated \textit{moefication} procedures \citep{zhang2021moefication,qiu2023emergent}. Importantly, in MoEs, each token is allocated a fixed amount of compute depending on the routing strategy (e.g., top-$k$ routing \cite{riquelme2021scaling}).
ACM can be seen as a modification of MoEs in which the experts are ordered, and thus, the complexity of the routing problem is drastically reduced. The gating network decides \emph{how many} learners instead of \emph{which} experts to execute, therefore allowing for a variable amount of computation on a per-token basis. A concurrent work of \citet{jainmixture} defines nested experts, which results in a dynamic model similar to ACMs.

\section{Method}
\label{sec:methods}
\begin{figure}
    \centering
    \includegraphics[width=0.35\textwidth]{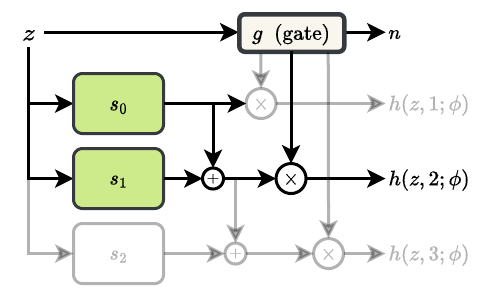}
    \caption{Architecture of an ACM block: the output is the sum of $k$ \emph{learners}, where $k$ is determined on a per-token basis by a small gating network $g$. The learners are executed in parallel. In the example, only the first two learners are executed, and the computation of the third (greyed out) is skipped.}
    \label{fig:acm_arch}
\end{figure}

An ACM is a conditional computation block that adapts its execution cost for each processed token via a trainable gating layer. In this paper, instead of training an ACM-based model from scratch, we focus on converting \textit{any} pre-trained transformer network into an equivalent ``ACMized'' version, i.e. one having similar accuracy while achieving a pre-defined computational budget on average. To this end, we propose an effective weight pre-initialization scheme. %
First, we substitute a subset of layers of the base model (e.g., MLPs or MHA projections) with an ACM of similar size. The ACMs are then trained independently but \textit{in parallel}, first with a per-layer reconstruction loss to initialize the learners and then using a cross-entropy loss to initialize the gates. The actual training consists of fine-tuning the model in an end-to-end manner to allow it to adapt its weight to the dynamic inference setting. In the supplementary material, we demonstrate that this setup significantly speeds up the training of the ACMized networks in all cases.

\subsection{Adaptive Computation Module}
\label{sec:acm_structure}

Our ACM module aims to allow the model to work well with any required computational budget while still ensuring efficient execution on GPUs, a property most of the dynamic width methods lack \citep{li2021dynamic}. Given an input token $z$, consider a set of $N$ homogeneous modules, $s_n(z; \phi_{(n)})$, $n \in \{1, ..., N\}$, each module having weights $\phi_{(n)}$. We refer to these modules as \emph{learners}. In an ACM, each learner progressively refines the prediction of the previous ones such that the $k$-th output of the ACM block, $k \in \{1, ..., N\}$, is given by:
\begin{equation}
    \label{eq:acm}
    h(z, k; \phi) = \sum_{n=1}^k s_n(z; \phi_{(n)})
\end{equation}
All intermediate outputs $h(z, 1; \phi), \ldots, h(z, N; \phi)$ are valid choices for the output of the ACM: a larger value for $k$ yields a more accurate result at the cost of a higher computational burden, while $k=1$ means that only a single learner is executed. Note that once $k$ is known for a token, the learners can be executed in parallel.

For any token $z$, $k$ should be chosen as the smallest possible value that guarantees good network performance. To this end, at the beginning of the ACM block, we add a small trainable gating network $g$ with weights $\omega$ to select the number of learners $k$ to be executed. In practice, the gating network returns $N$ real-valued outputs, which are then discretized into a one-hot gating choice vector with the Gumbel-Softmax trick \citep{jang2016categorical} to retain differentiability:
\begin{equation} \label{eq:gs}
    \nu_n = \frac{\exp((\log(g(z; \omega))_n + \gamma_n) / T)}{\sum_{n}\exp((\log(g(z; \omega))_n + \gamma_n) / T)}.
\end{equation}
In \Cref{eq:gs}, $\gamma_1,...,\gamma_N$ are i.i.d samples from $\mathtt{Gumbel}(0, 1)$, and $T$ is softmax temperature. In the forward pass, we discretize $\nu$ into a one-hot vector $\hat{\nu}$, but the continuous values are used directly for gradient computation in the backward pass. The complete ACM architecture is outlined in \Cref{fig:acm_arch}.

\paragraph{Design of the ACM blocks} Any network architecture with the required input and output dimensionality can play the role of a learner. For simplicity, in this paper, we always use two dense layers with a GELU activation function in between. Since we replace selected modules of a pre-trained model with ACMs, we always pick $N$ and the size of a single learner such that the entire ACM module has approximately the same computational cost and the number of parameters as the substituted block. This is straightforward to achieve as the cost of a learner scales linearly with its hidden dimensionality. However, the output dimension has to be the same as in the replaced module. This results with output layer biases taking a larger share of learner parameters as we increase $N$. To avoid this effect, we simply eliminate them from our architecture. 

For the gating network, we also use a two-layer network and determine its hidden dimensionality such that the total computational cost of gating is around $1\%$ of the overall model cost. When feasible -- such as the module being placed under a residual connection -- we set the minimum number of executable learners to $0$ so that even the computation of the first learner can be skipped.

\subsection{ACM weight initialization}
\label{sec:representation_distillation}
The costs of training large models are constantly increasing \citep{strubell2019energy}. On the other hand, there is a large number of publicly available pre-trained models, so the capability of adapting them is a desired property for new methods. Since ACMs are designed to replace individual modules, we initially train them by distilling the knowledge from the corresponding trained modules that are being substituted. Specifically, we propose the following scheme. A trained static model $f(\cdot)$ is cloned, and selected modules (e.g., every MLP module) are replaced with ACMs in that cloned model $f_\text{ACMized}(\cdot)$. Each learner from every ACM is initialized randomly. For each sample $x_i$ from a mini-batch $B$ forwarded through the original model, we save every input token $z_{i,j}^l$ and output token $o_{i,j}^l$ of each $l$-th module that was replaced in the copied model. Then, every ACM is trained independently and in parallel by minimizing the mean squared error (MSE) applied for every possible choice of $k$:
\begin{equation}
\mathcal{L}(\phi) = \frac{1}{|B|SLN} \sum_{i,j,l,n} \left\lVert h(z_{i,j}^l, n; \phi^l) - o_i^l \right\rVert ^2
\end{equation}
where $S$ is token sequence length, and $L$ is the number of substituted modules. Note that the gating network is not needed in this phase. The effectiveness of this training approach can be tested by setting a fixed $k$ for every ACM in the model and evaluating the model on the validation set.

With learners being able to reliably imitate the replaced modules, we subsequently freeze them and train the gating networks in a similar, layer-wise approach. We frame the problem as a classification task and generate artificial labels with the following heuristic. First, we consider the outputs of all learners and compute the distance to the original output:
\begin{equation}
    d_{i,j}^l(n) = \| h(z_{i,j}^l, n; \phi) - o_{i,j}^l \|_{2}
\end{equation}
The target label is then set as:
\begin{equation}
    t_{i,j}^l = \min \left\{ n  \in \{2, ... , N\} \,\middle\vert\,  \frac{d_{i,j}^l(n)}{d_{i,j}^l(n - 1)} \geq \tau \right\},
\end{equation}
\noindent where $\tau$ is a threshold hyperparameter. In other words, we select the smallest number of learners such that the relative improvement from adding one more learner is lower than the threshold $\tau$. With these labels, the gating networks are trained using standard cross-entropy loss:
\begin{equation}
    \mathcal{L}(\omega) = \frac{1}{|B|SLN}\sum_{i,j,l,n} -t_{i,j,n}^l \log(\nu_{i,j,n}^l).
\end{equation}

\subsection{End-to-end training}
\label{sec:end_to_end_training}
In the third phase, we finetune the entire model end-to-end to allow it to adapt its weights to the dynamic inference setting. To make the model more predictable in terms of its computational cost, we add an auxiliary loss term that penalizes for any deviation from the given target budget $\beta_{\text{target}}$ on average:
\begin{equation}
    \mathcal{L}_{\text{b}}(\theta) = \left\lVert\frac{1}{|B|}\sum_i\frac{\sum_{j}\sum_{l} k_{i,j}^l p^l}{\sum_{j}\sum_{l} N p^l} - \beta_{\text{target}}\right\rVert_1,
\end{equation}
\noindent where $p^l$ is the computational cost of a single learner from layer $l$ and $\beta_{\text{target}} \in (0, 1)$ is the targeted computational budget. This term still allows for the allocation of different amounts of computational resources to samples of varying complexity and only requires to be close to $\beta_{\text{target}}$ \emph{on average}. It also affects the computational cost of future training steps, potentially accelerating the training process.

While $\mathcal{L}_{\text{b}}$ does not prevent diversity of gating choices, it does not encourage it. In practice, we find that the gating networks collapse to a state where the same number of learners is chosen for every input token, effectively turning our dynamic model into a static one. To prevent this behavior and encourage diversity, we add two additional loss terms. The first auxiliary loss term maximizes the average normalized entropy of gating choices taken for tokens in a single image:
\begin{equation}
    \mathcal{L}_{\text{e}}(\theta) = \frac{1}{|B|L} \sum_{i,l} \frac{\sum_n a_n \log(a_n)}{\log(N)},
\end{equation}
\noindent where:
\begin{equation}
    a^l_{i,n} = \frac{\sum_j \hat{\nu}^l_{i,j,n}}{\sum_n \sum_j \hat{\nu}^l_{i,j,n}}
\end{equation}
represents a distribution between $N$ choices in batch $B$ for the $l$-th ACM aggregated for an entire sequence of tokens from sample $i$. The intuition behind this loss is that not every input region is equally important; hence, a non-uniform distribution of computation is required.

Finally, entire images may exhibit different difficulty levels, and enforcing diversity of gating choices for a single image at a time may not be enough. We address this with the second auxiliary loss, which is defined as:
\begin{equation}
    \mathcal{L}_{\text{d}}(\theta) = \frac{1}{{|B|}^2} \sum_i \sum_m \left\lVert b_i - b_m \right\rVert_1,
\end{equation}
where:
\begin{equation}
    b_i = \frac{\sum_j \sum_l k^l_{i,j}}{\sum_j \sum_l N}
\end{equation}
\noindent denotes the fraction of learners executed on sample $i$. It encourages the model to distribute its computational budget between easy and hard examples. 
The final loss that is minimized for classification tasks is given by:
\begin{equation}
\begin{aligned}
    \mathcal{L}(\theta) = & \frac{1}{|B|} \sum_i \sum_c -y_{i,c} \log(\hat{y}_{i,c}) + \alpha_{\text{b}}\mathcal{L}_{\text{b}}(\theta) \\ 
    & \qquad + \alpha_{\text{e}}\mathcal{L}_{\text{e}}(\theta) + \alpha_{\text{d}}\mathcal{L}_{\text{d}}(\theta),
\end{aligned}
\end{equation}
\noindent where $\{(x_i, y_i)\}_{i=1}^{|B|}$ are samples from the current mini-batch $B$, and $\alpha_{\text{b}}$, $\alpha_{\text{e}}$, $\alpha_{\text{d}}$ are hyperparameters for weighting the different terms. In practice, we always use $\alpha_{\text{b}} = 0.1$, $\alpha_{\text{e}} = 0.05$, and $\alpha_{\text{d}} = 0.05$. In the analysis section, we present an ablation study that demonstrates the necessity of applying the proposed auxiliary losses and shows the positive impact of their interplay. Note that the auxiliary losses are task-agnostic, allowing ACMs to be used for any other task than classification.

\section{Experiments}
Due to the widespread availability of pre-trained weights, we evaluate our method on popular image classification and speech recognition tasks. The aim of the evaluation is to compare the performance-efficiency trade-offs of different methods. To measure the computational efficiency, we track the number of FLOPs used by the model for each evaluated sample and report the averaged result. The source code for our experiments is available at \url{https://github.com/bartwojcik/adaptive_computation_modules}.

\begin{figure}[h!]
    \centering
    \includegraphics[width=0.35\textwidth]{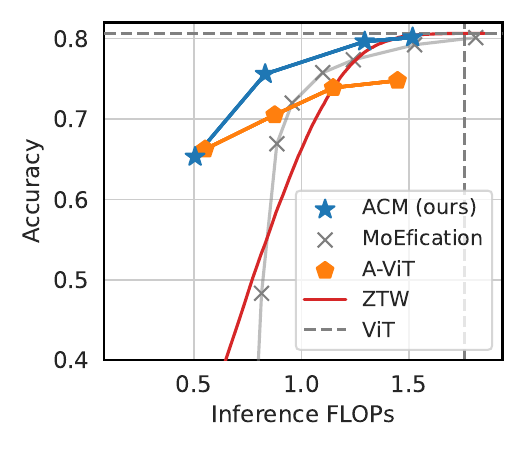}
    \caption{Performance-efficiency trade-offs of different conditional computation methods as measured on the ImageNet-1k dataset. ACM-based ViT-B achieves the Pareto frontier for a wide range of computational budgets.}
    \label{fig:cost_vs_acc_cv}
\end{figure}

\subsection{Computer Vision}
\label{sec:cv_experiments}

We select a ViT-B \citep{dosovitskiy2020image} model pre-trained on ImageNet-1k \citep{ILSVRC15} from the \verb|torchvison| library\footnote{\url{https://pytorch.org/vision/stable/models.html}} as the base model for all methods. We compare ACMized models with three conditional computation techniques, each one coming from a different group: A-ViT \citep{yin2022vit} (Token Dropping), MoEfication \citep{zhang2021moefication} (Mixture-of-Experts), and Zero Time Waste \citep{wolczyk2021zero} (Early Exiting). For the sake of a fair comparison, we assume the same training data budget of 100 epochs for every method. 

Since MoEfication requires models using ReLU activation functions, we first replace GELU activations with ReLUs and finetune the model for 80 epochs, as described by \citet{zhang2021moefication}. The remaining 20 epochs are used for training the routers. For Zero Time Waste, we train the early-exit heads for 95 epochs and then train the ensembles for 5 epochs. For the ACMized ViT, we replace every MLP module and every linear projection in each MHA with an ACM module (the self-attention mechanism itself is not affected). Module-wise representation distillation is performed for 2 epochs, followed by 1 epoch for pre-training of the gating networks and 97 epochs of end-to-end finetuning of the entire model. We set the number of learners $N$ to $4$ in every ACM. While MoEfication and Zero Time Waste can be evaluated with different numbers of selected experts and early exit confidence thresholds, A-ViT and ACMs need to be trained multiple times with different values of hyperparameters that approximately determine the final average computational budget. We emphasize that our A-ViT implementation includes fixes for two issues raised by GitHub users\footnote{\url{https://github.com/NVlabs/A-ViT/issues/4}}\footnote{\url{https://github.com/NVlabs/A-ViT/issues/13}}, which may affect the final performance of A-ViT. The authors of A-ViT have not yet addressed them at the time of writing this article.

We present the results in \Cref{fig:cost_vs_acc_cv}. As projections in MHA are responsible for around $30\%$ of the model cost, MoEfication seems to be hindered by the fact that it reduces only the cost of the MLP layers. A-ViT shows a gap in performance in relation to the pre-trained model, while Zero Time Waste is competitive only for higher computational budgets. ACMized ViT displays favorable performance for a wide range of computational budgets, and its advantage is especially significant below $12.5$ GFLOPs.

\begin{figure}[t]
    \centering
    \includegraphics[width=0.35\textwidth]{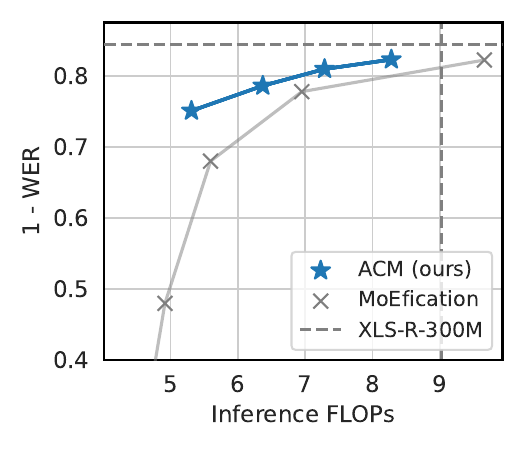}
    \caption{Performance-efficiency trade-offs of different conditional computation methods as measured on the CommonVoice-es dataset. The model's performance is reported in terms of Word Error Rate (WER). ACMs achieve lower WER for every evaluated computational budget.}
    \label{fig:cost_vs_acc_stt}
\end{figure}

\subsection{Speech-to-Text}
\label{sec:stt_experiments}

For speech recognition tasks, we use the XLS-R-300M \cite{babu2021xlsr}, a pre-trained Wav2Vec2 model \cite{baevski2020wav2vec}, fine-tuned on selected languages from the CommonVoice dataset \cite{ardila2020common}. Speech-to-text models introduce additional complexities for dynamic computation methods, as each token is decoded individually. Since A-ViT and Zero Time Waste were not designed for this setting, we restrict our baseline methods to MoEfication, the only task-agnostic method among those three.

We assume an equal training data budget of 10 epochs for all approaches. In the case of MoEfication, we substitute GELUs with ReLUs and train for eight epochs, followed by two epochs of router training.
For ACMs, we replace every MLP block within the model with an ACM module with $N=4$. We subsequently perform five epochs of module-wise distillation, one of training the gating networks, and four epochs of end-to-end finetuning. We present results in \Cref{fig:cost_vs_acc_stt}. We see that even if only the MLP blocks are ACMized, our method still obtains a better performance-efficiency trade-off than MoEfication.

\section{Analysis}
\label{sec:analysis}

Dynamic models allocate variable amounts of compute for different samples or input regions. In this section, we provide motivation for our method, examine the distribution of computational load, and show that the introduced auxiliary losses are needed. In the supplementary material, we show the impact of the proposed pre-training stages and investigate the effect of changing the number of learners $N$.

\begin{figure}[t]
  \centering
  \includegraphics[width=0.9\linewidth]{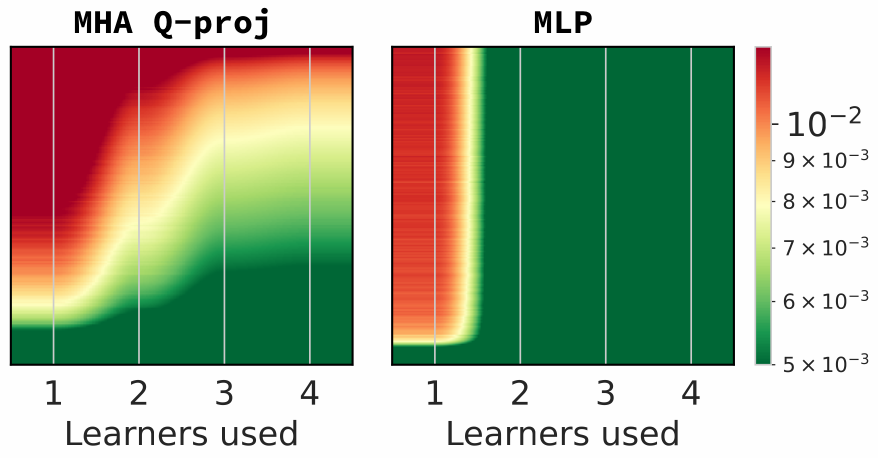}
  \caption
    {%
      Color-coded errors of a $4$-learner ACM plotted after performing module-wise representation distillation for modules from eight block of a ViT-B pre-trained model. Tokens are sorted along the Y-axis of the plot by their average error. For most input tokens, the same transformation can be performed by a considerably smaller module consisting of only two or three learners, thus justifying the use of ACMs.
      \label{fig:full_rep_distill}%
    }%
\end{figure}

\subsection{Computational Inefficiency of Static Modules} To justify the use of ACMs, we make the following experiment. First, we perform module-wise distillation from a selected static module of an ImageNet-1k pre-trained ViT-B into $4$ learners, just as we describe in the Method section. After convergence, we randomly select $5000$ sample images from the validation dataset and forward them through the original model to save every input token $z_{i,j}$ and output token $o_{i,j}$. For every possible number of learners used $n \in \{1, ..., N\}$, we are interested in how well the learners imitate the output of the original module. For this, as in training, we use MSE: $\left\lVert h(z_{i,j}^l, n; \phi^l) - o_i^l \right\rVert^2$.
The resulting tensor of MSE values has size $(5000, 197, 4)$, where $197$ is the sequence length specific to ViT-B with patch size set to $16$. Since ACMs process each token independently, we flatten the first two dimensions and then sort the tokens by the average error for readability. \Cref{fig:full_rep_distill} shows the resulting color-coded error visualizations for selected modules. We emphasize that the four learners have approximately the same cost and number of parameters as the original static module being substituted.

The results show that 1) only a small subset of tokens require the full processing power of the entire module, and for a majority of tokens, and 2) tokens usually exhibit varying levels of difficulty, thus warranting the use of conditional computation on a per-token basis.

\begin{figure}[t!]
    \centering
    \begin{subfigure}{0.45\columnwidth}
        \includegraphics[width=\columnwidth]{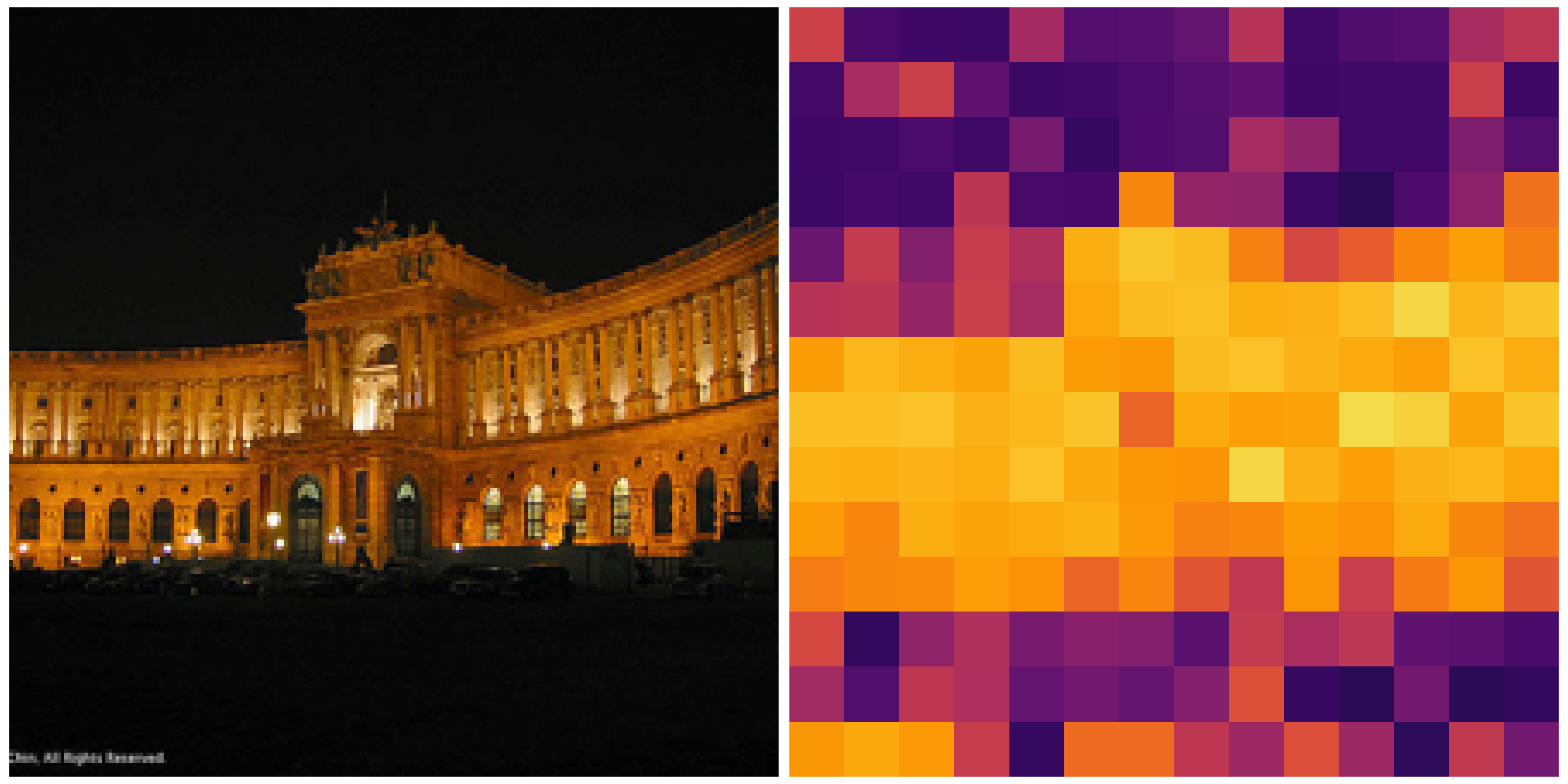}
    \end{subfigure}%
    \begin{subfigure}{0.45\columnwidth}
        \includegraphics[width=\columnwidth]{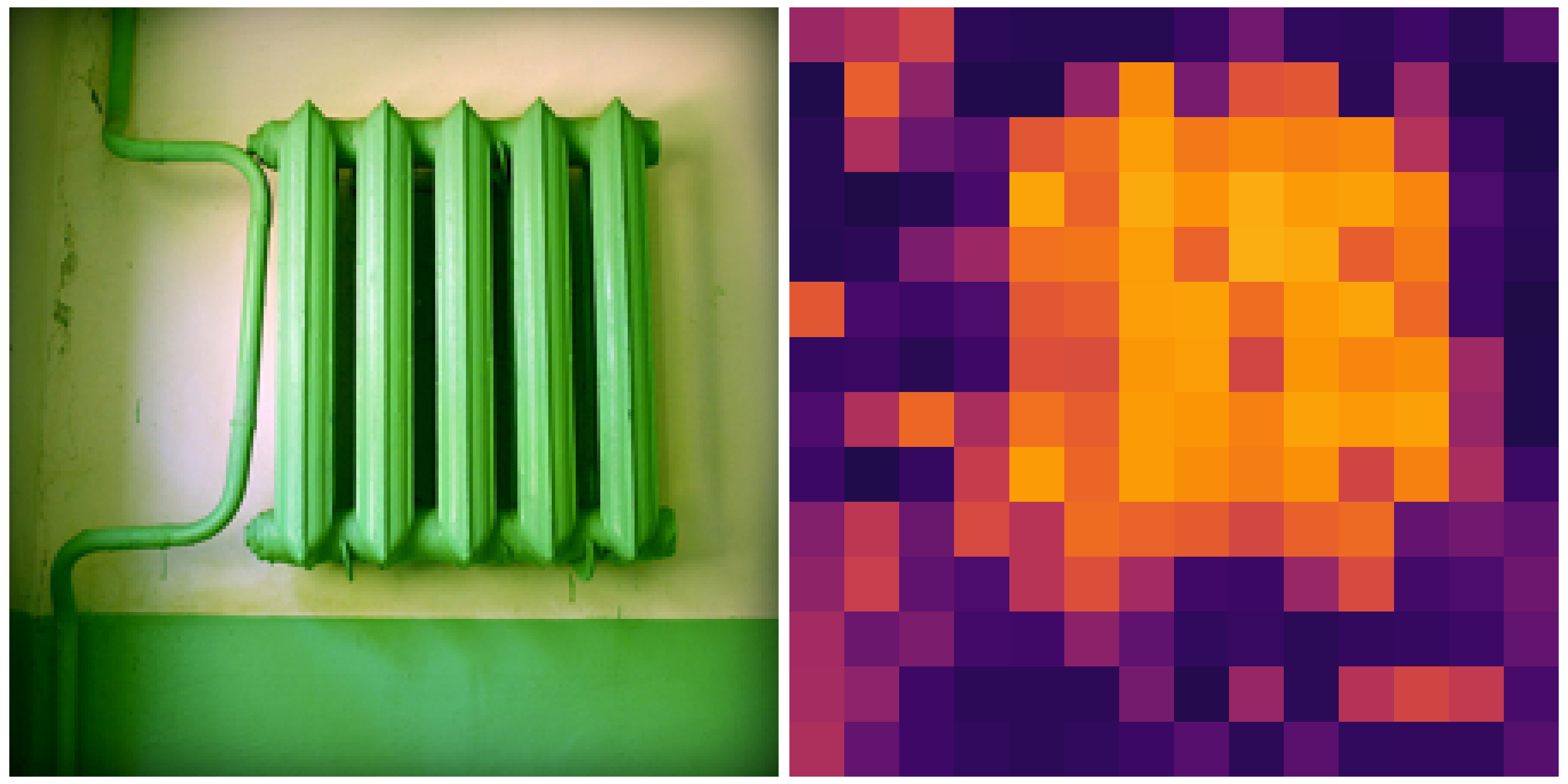}
    \end{subfigure}
    \begin{subfigure}{0.45\columnwidth}
        \includegraphics[width=\columnwidth]{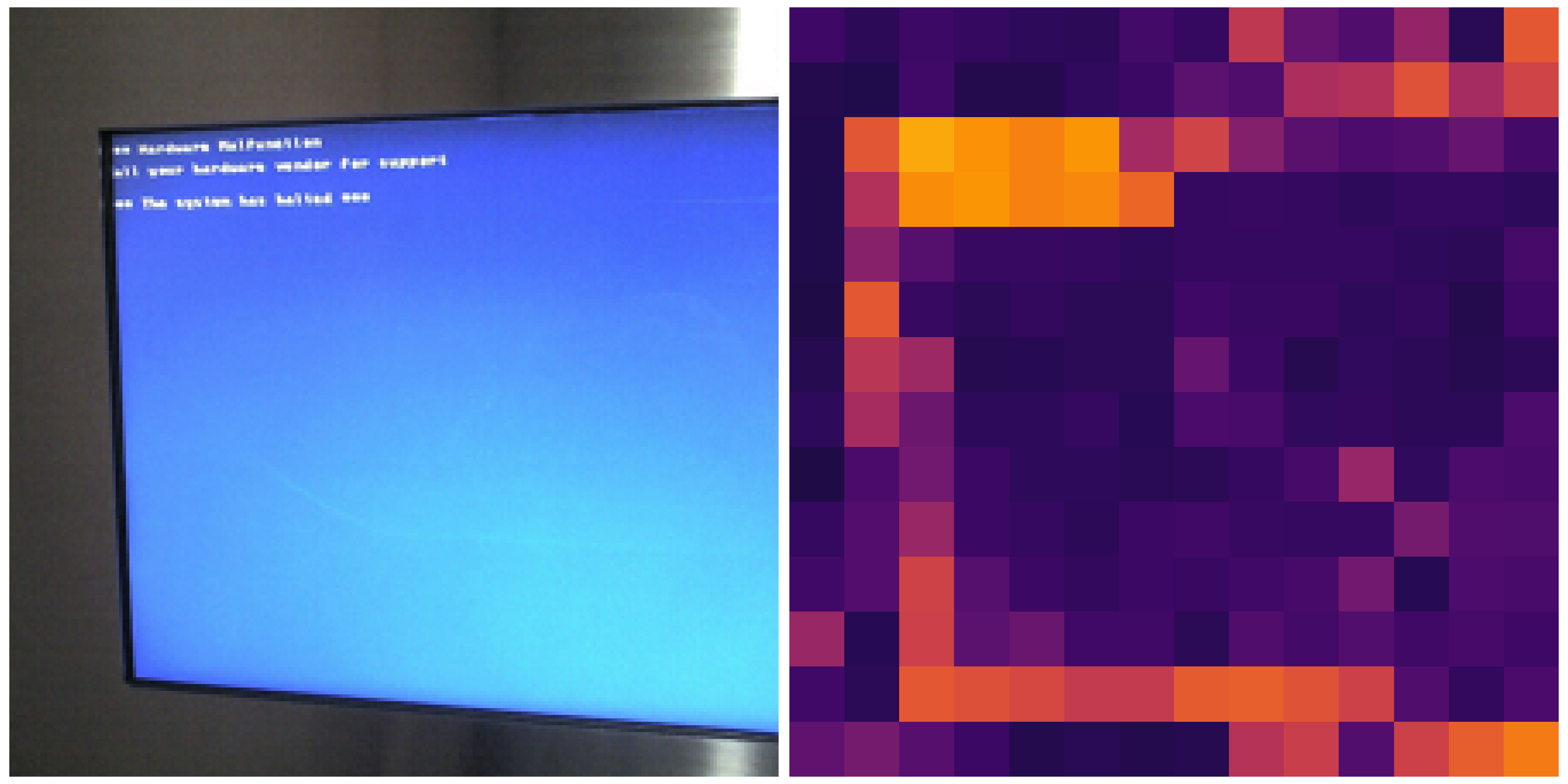}
    \end{subfigure}%
    \begin{subfigure}{0.45\columnwidth}
        \includegraphics[width=\columnwidth]{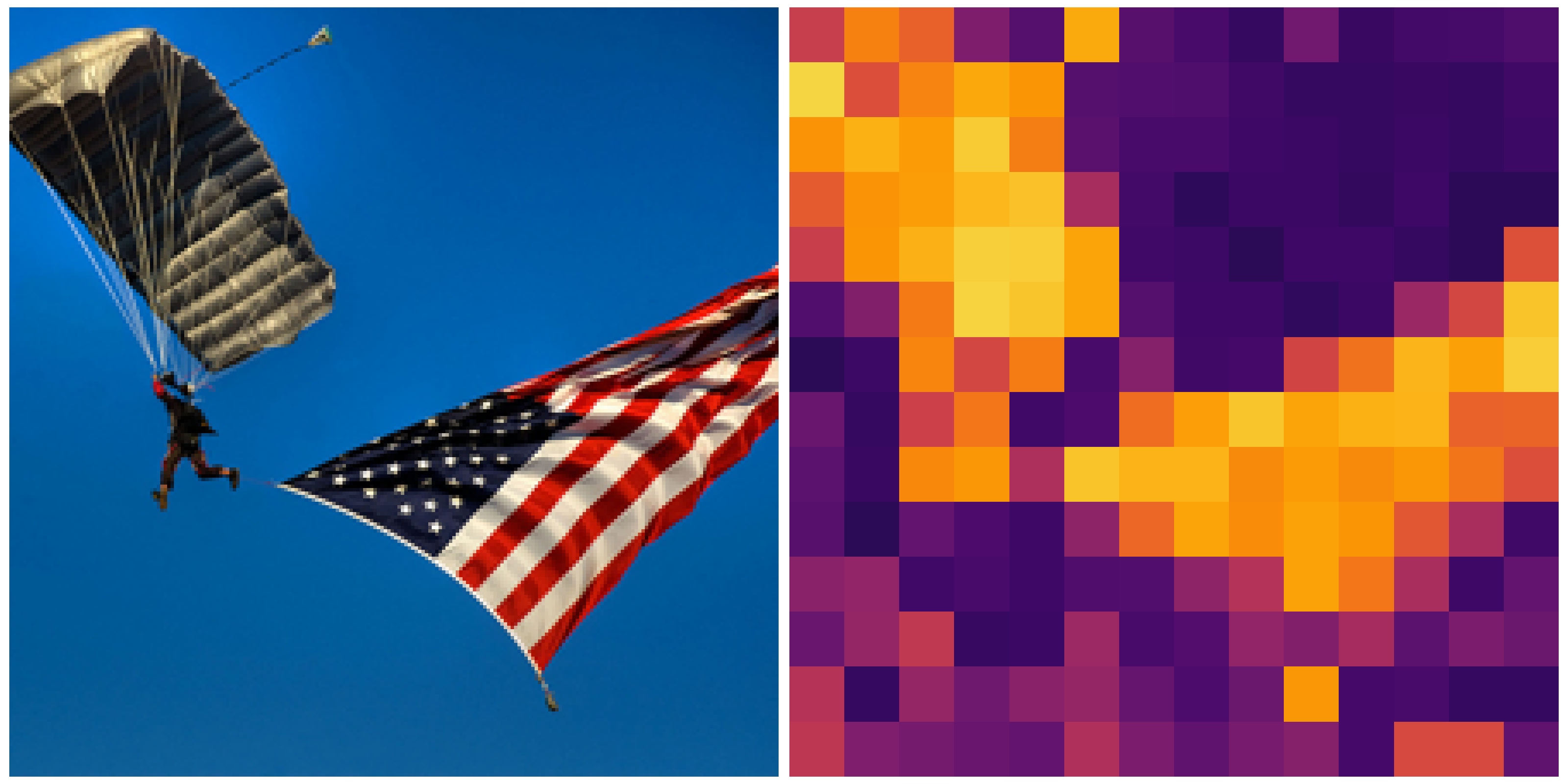}
    \end{subfigure}
    \begin{subfigure}{0.45\columnwidth}
        \includegraphics[width=\columnwidth]{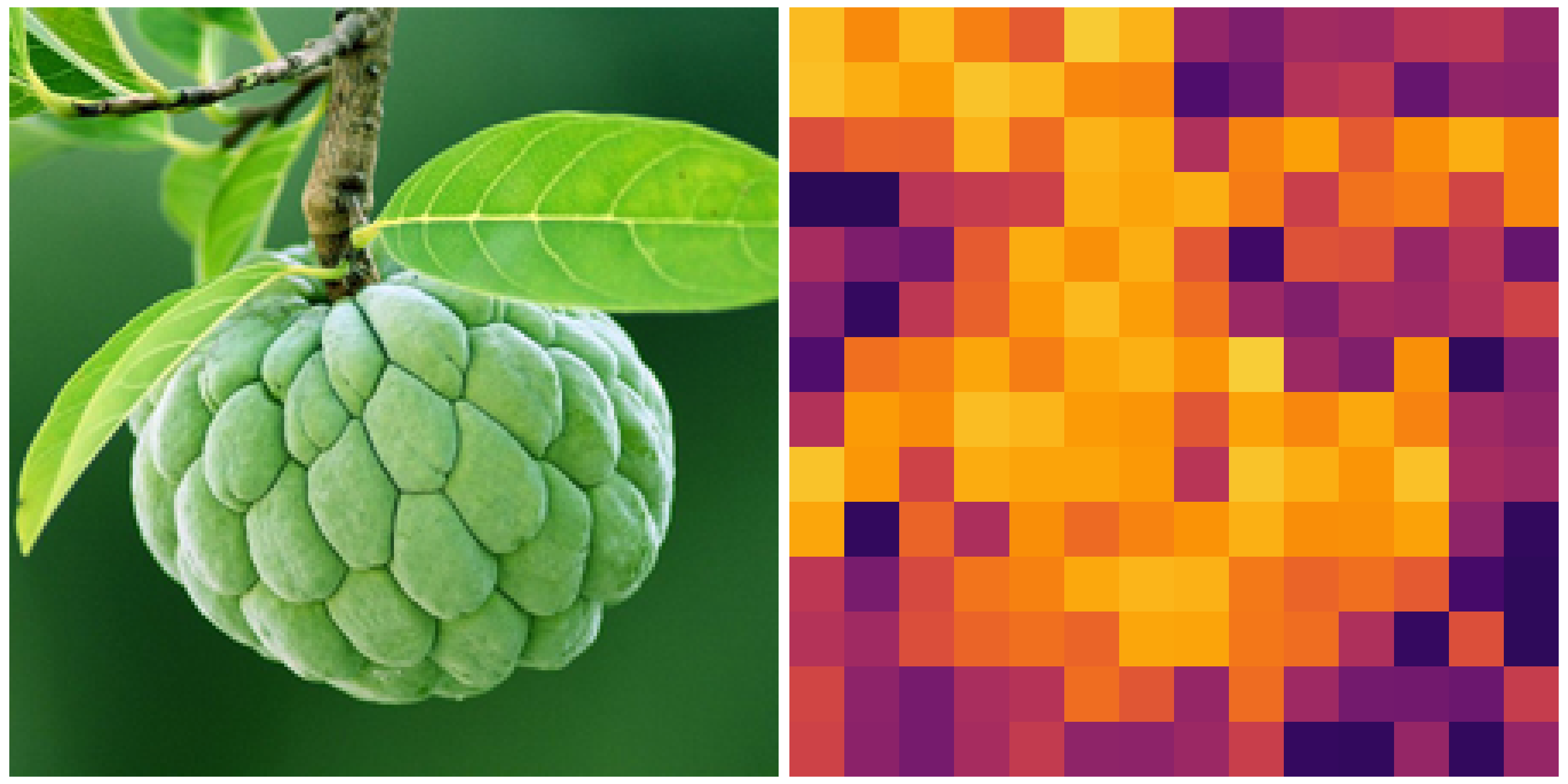}
    \end{subfigure}%
    \begin{subfigure}{0.45\columnwidth}
        \includegraphics[width=\columnwidth]{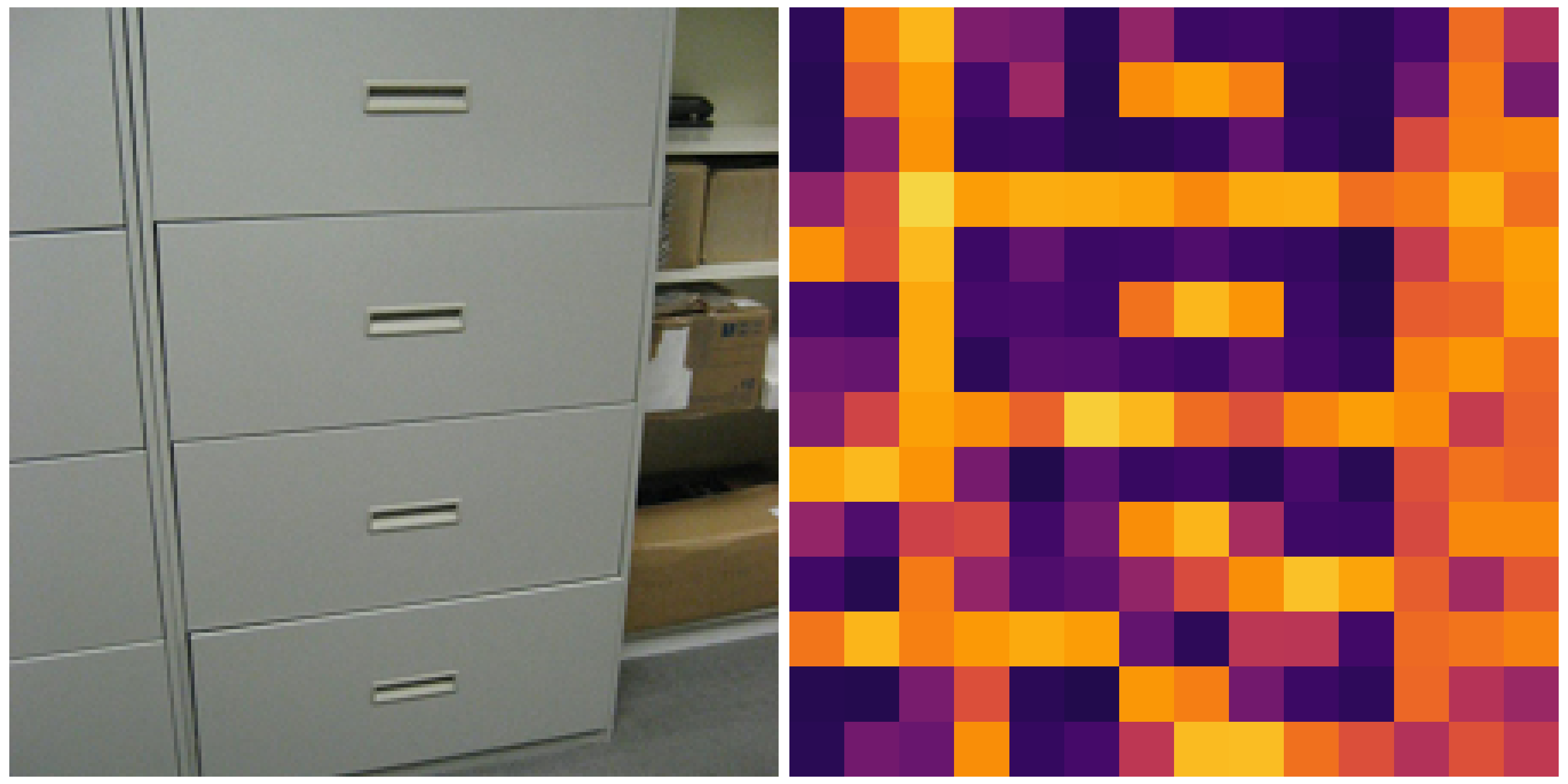}
    \end{subfigure}
    \begin{subfigure}{0.45\columnwidth}
        \includegraphics[width=\columnwidth]{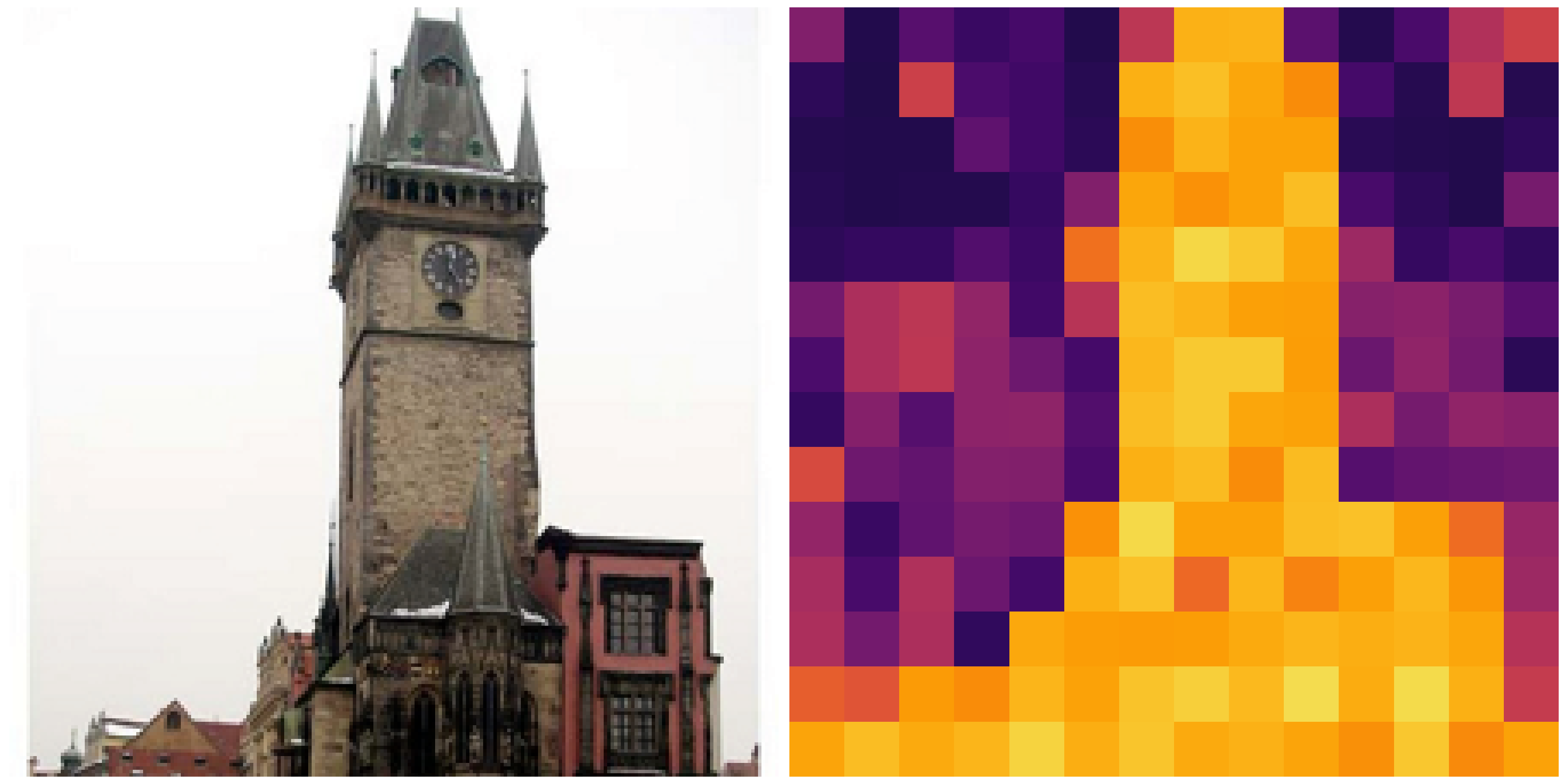}
    \end{subfigure}%
    \begin{subfigure}{0.45\columnwidth}
        \includegraphics[width=\columnwidth]{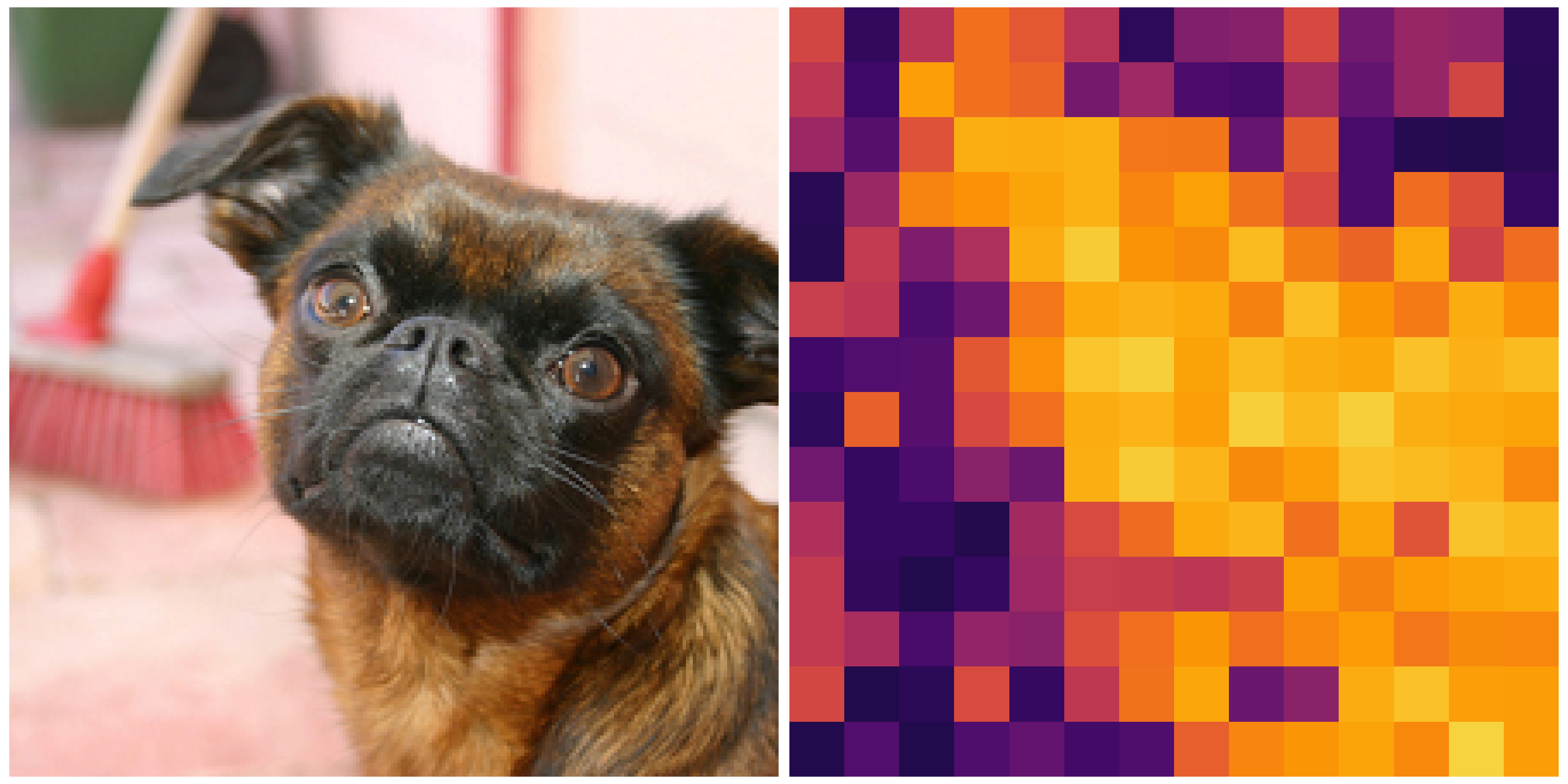}
    \end{subfigure}%
    
    \caption{Computational load heatmaps for the model trained with $\beta_{\text{target}} = 0.6$. The model allocates its computational budget to semantically meaningful patches.%
    \label{fig:heatmaps}}
\end{figure}

\begin{figure}[h!]
    \centering
    \begin{subfigure}{0.9\columnwidth}
        \includegraphics[width=\columnwidth]{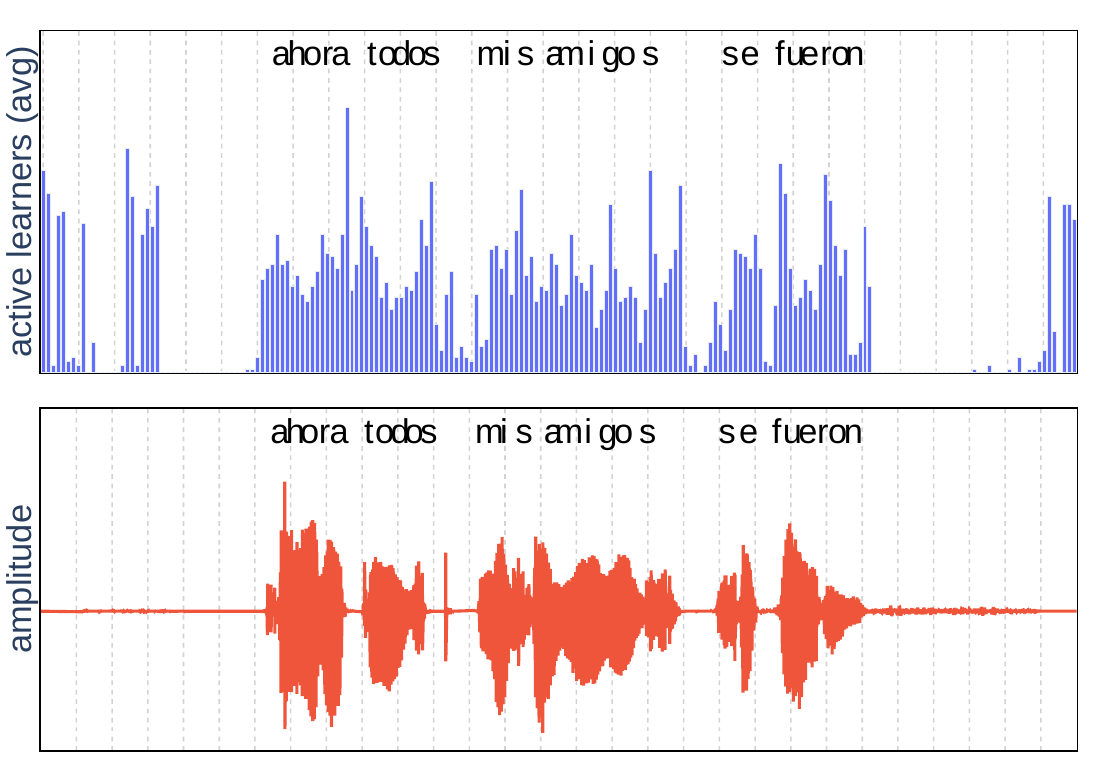}
    \end{subfigure}
    \caption{For each input audio token (red waveforms), we show the average number of learners that were activated in the ACMized model for $\beta_{\text{target}} = 0.25$ (blue bars). We can see that this model also learned to allocate its computational budget to important regions of the input.}
    \label{fig:audio_computational_load}
\end{figure}

\begin{figure*}[t!]
    \centering
    \begin{minipage}[c]{.18\linewidth}
        \centering
        {\hspace*{0.15\linewidth}\includegraphics[width=0.75\linewidth]{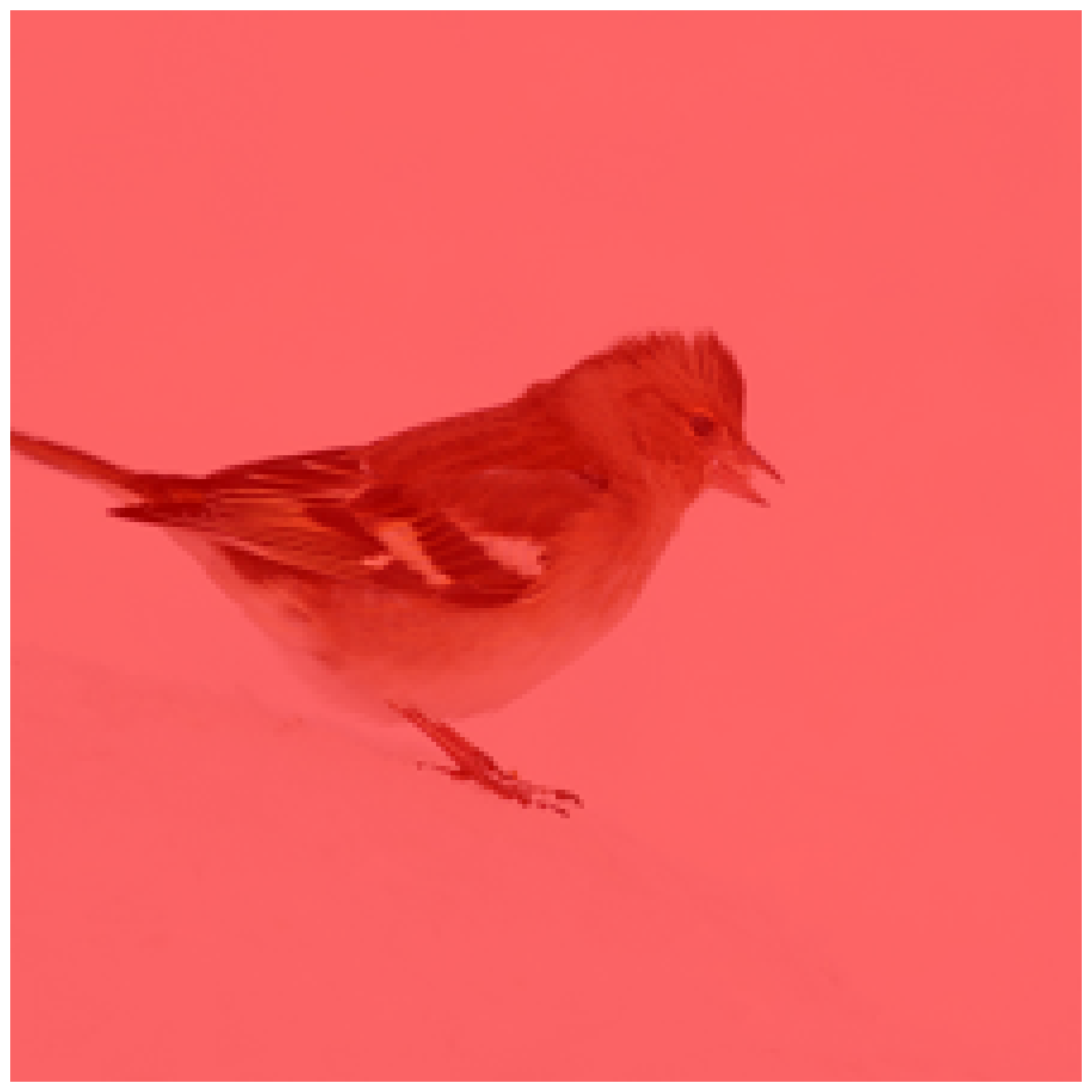}}%
    \end{minipage}
    \begin{minipage}[c]{.18\linewidth}
        \centering
        {\hspace*{0.15\linewidth}\includegraphics[width=0.75\textwidth]{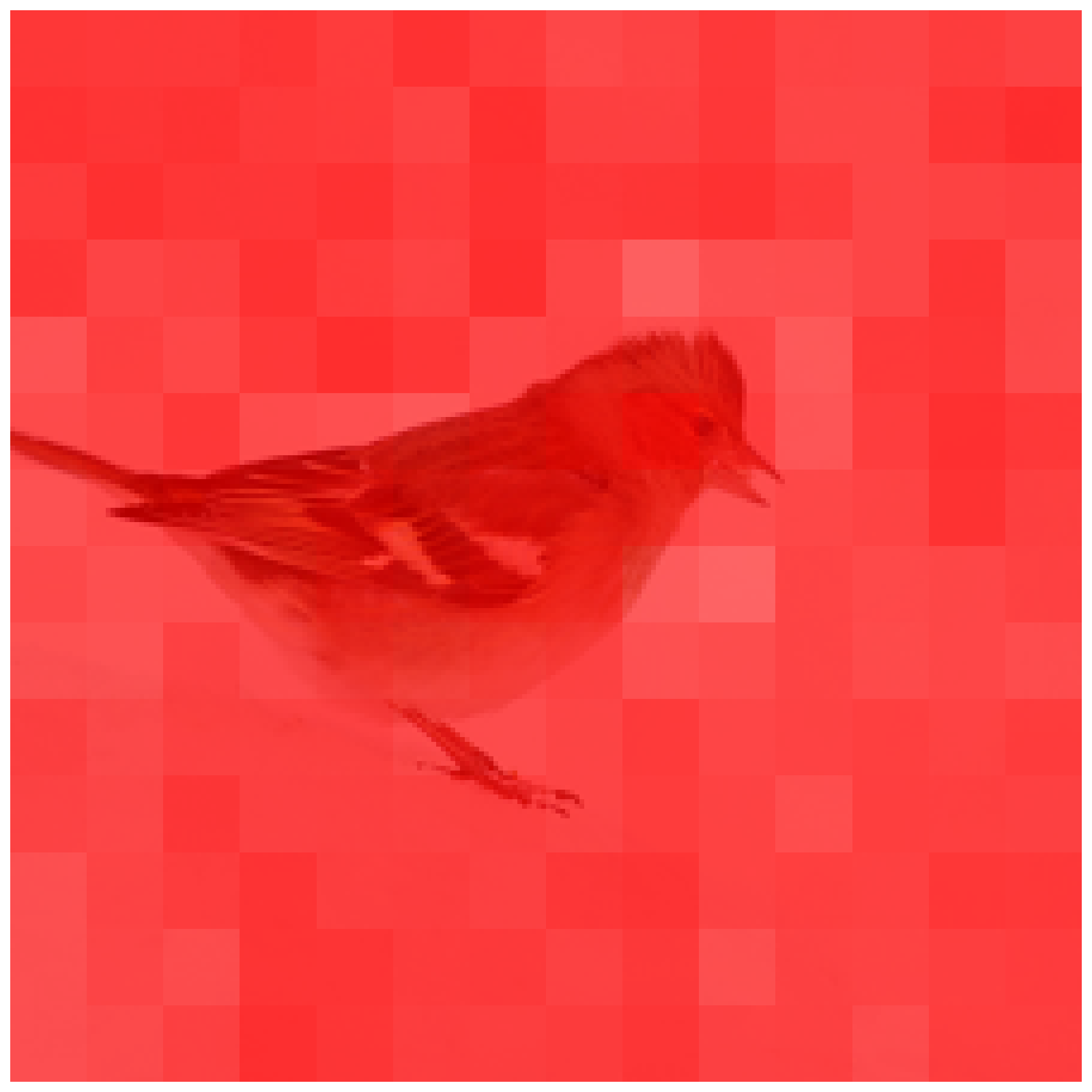}}
    \end{minipage}
    \begin{minipage}[c]{.18\linewidth}
        \centering
        {\hspace*{0.15\linewidth}\includegraphics[width=0.75\linewidth]{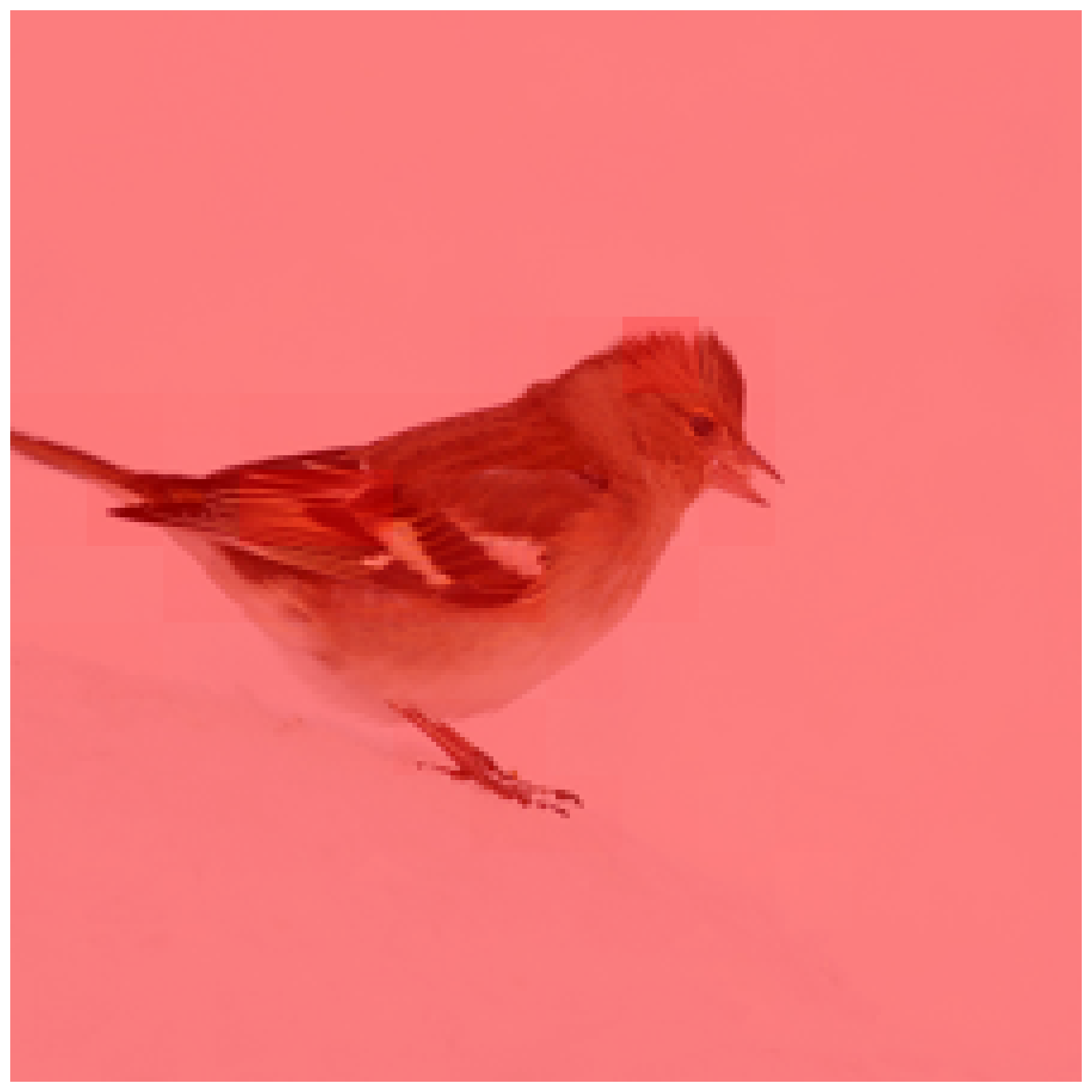}}%
    \end{minipage}
    \begin{minipage}[c]{.18\linewidth}
        \centering
        {\hspace*{0.15\linewidth}\includegraphics[width=0.75\textwidth]{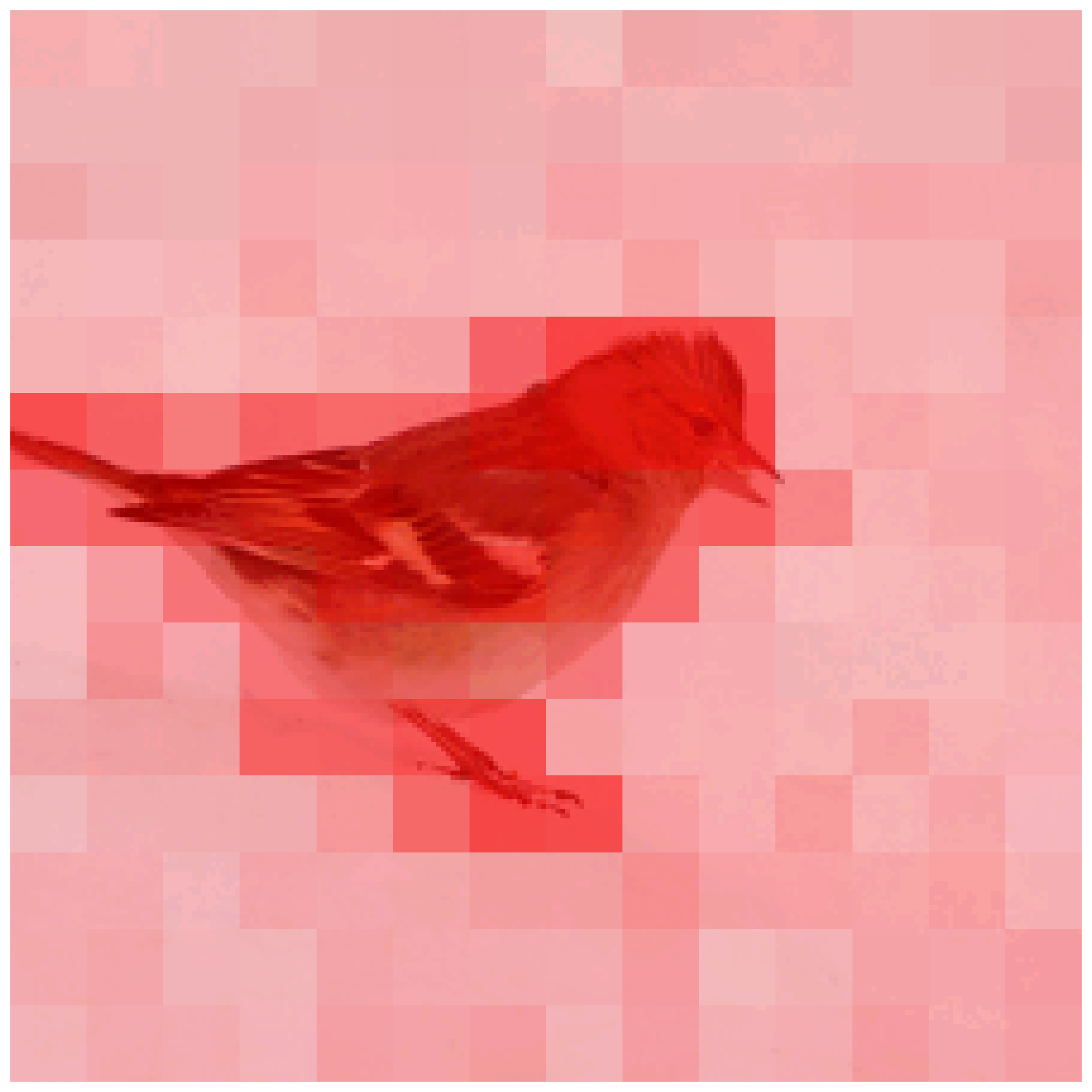}}
    \end{minipage}
    \\
    \begin{minipage}[c]{.18\linewidth}
        \centering\subfloat[Only $\mathcal{L}_{\text{b}}$]
        {\includegraphics[width=\linewidth]{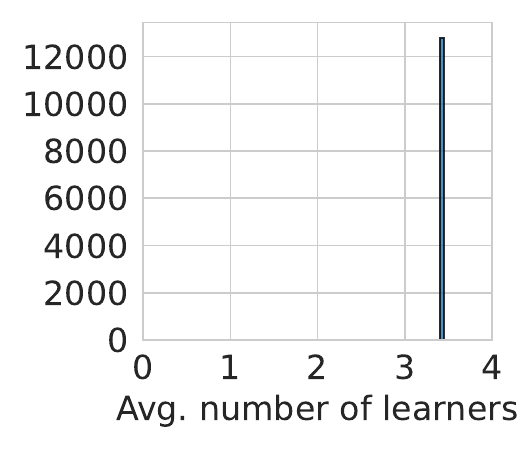}}%
    \end{minipage}
    \begin{minipage}[c]{.18\linewidth}
        \centering\subfloat[$\mathcal{L}_{\text{b}}$ and $\mathcal{L}_{\text{e}}$]
        {\includegraphics[width=\linewidth]{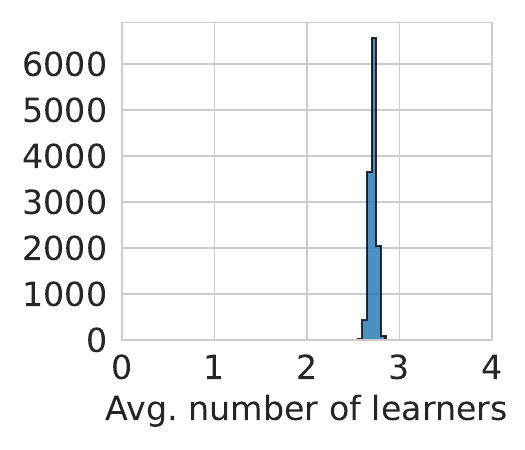}}
    \end{minipage}
    \begin{minipage}[c]{.18\linewidth}
        \centering\subfloat[$\mathcal{L}_{\text{b}}$ and $\mathcal{L}_{\text{d}}$]
        {\includegraphics[width=\linewidth]{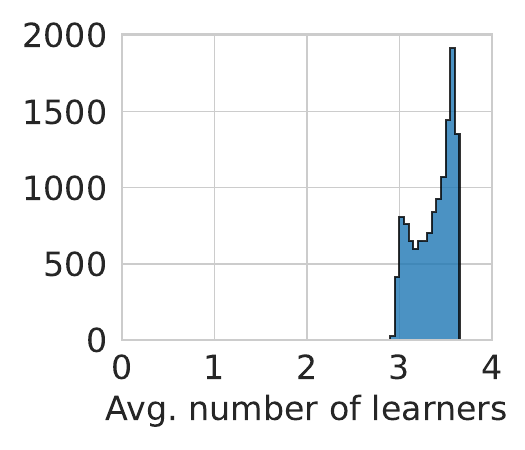}}%
    \end{minipage}
    \begin{minipage}[c]{.18\linewidth}
        \centering\subfloat[$\mathcal{L}_{\text{b}}$, $\mathcal{L}_{\text{e}}$ and $\mathcal{L}_{\text{d}}$]
        {\includegraphics[width=\linewidth]{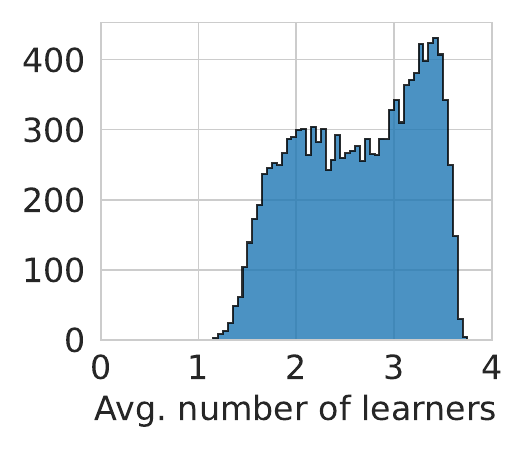}}
    \end{minipage}
    
    \caption{Effects of training with different combinations of enabled auxiliary losses. The computational load heatmap is rendered over the original image (top row). The histograms (bottom row) show the distribution of the total computational cost spent by the model on images from the validation set. Only combining all the proposed terms provides the desired diversity of the allocated computational budget.%
    \label{fig:loss_term_ablation}}
\end{figure*}

\subsection{Qualitative Analysis} A dynamical model should take advantage of the fact that images exhibit different difficulty levels for classification. By extracting the gating choices being done by every ACM for each patch, we can plot a heatmap that indicates which regions the model was focused the most on. This map should correlate with the meaningful regions of the image. \Cref{fig:heatmaps} shows that the model indeed learns to allocate its computational budget to those regions. We show that this effect is not exclusive to vision by performing a similar analysis for audio in \Cref{fig:audio_computational_load}. We provide additional examples in the supplementary material.

\subsection{Ablation Study} Being able to dynamically allocate computation when solving a task is the core feature of ACMized models. The auxiliary losses are necessary for the dynamism to emerge. We empirically demonstrate this in \Cref{fig:loss_term_ablation} by training multiple runs differing only in the loss terms applied during the end-to-end training phase. For each run, we visually analyze their intra-sample computational load distribution and generate a histogram of the total computation spent on every sample.

As expected, ACMs may converge to a solution in which always the same learners are used when only $\mathcal{L}_{\text{b}}$ is applied. The inclusion of $\mathcal{L}_{\text{e}}$ helps in diversifying between different regions of the input, but overall the computation spent on each image is mostly the same. Applying $\mathcal{L}_{\text{d}}$ instead of $\mathcal{L}_{\text{e}}$ diversifies the distribution of compute spent on every image, but the gating choices for different regions of the input are not diverse on its own. Only by enabling all the proposed losses can we achieve the desired effect of the model focusing on semantically meaningful patches and having a high variability of computational budget allocation.


\begin{figure}[h]
    \centering
    \includegraphics[width=0.7\linewidth]{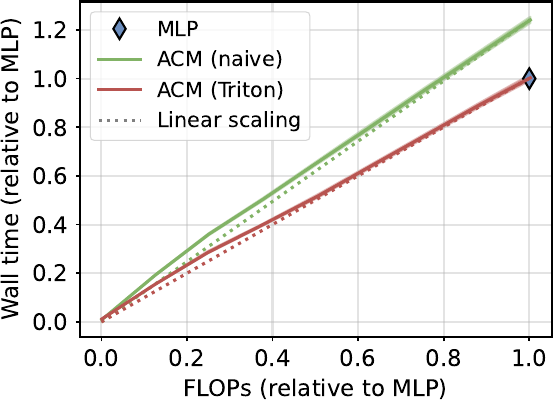}
    \caption{Latency of a $128$-sample batch processed by a single ACM layer on an A100 GPU. The gating network is included in the measurements, but we replace the gating choices with samples from a random distribution to achieve the desired average number of executed learners. ACMs have negligible overhead, and latency scales linearly with the average number of executed learners.}
    \label{fig:acm_layer_wallclock}
\end{figure}

\subsection{Hardware speedup} 
Due to their design, ACMs are inherently well-suited for parallel execution on GPUs. Specifically: (1) once gating decisions are determined, the execution of learners can occur in parallel; (2) tokens can be reordered without any additional copying when they are loaded from or stored to the main GPU memory by the kernel; (3) when tokens are sorted by the number of selected learners, the computation for each group is standard matrix multiplication for which GPUs are highly optimized for; (4) the hidden dimensionality of each learner is deliberately large to maximize GPU utilization. We implement the ACM forward pass with GPU kernels written in Triton \citep{tillet2019triton} and employ several optimizations including configuration auto-tuning and kernel fusion. 

In \Cref{fig:acm_layer_wallclock} we evaluate our implementation by measuring the wall-clock time of execution of a single ACM module from the ViT-B-based model. The overhead in respect to a static MLP is negligible. Moreover,  the wall clock time decreases linearly with the number of executed learners. 



\section{Conclusion}
In this work, we have demonstrated that large static modules lead to ineffective allocation of computational budget. The introduced ACM, a generic module that facilitates granular conditional computation, is designed for computational effectiveness, and models based on it achieve state-of-the-art results among the tested dynamic inference methods. Our training procedure distills a static network into an adaptive one, forcing it to allocate its computational budget to semantically meaningful input regions. 
Future work might explore the relationship between ACMs and pruning methods. Since our training procedure replaces individual modules with ACMs, one could also consider using quantized learners for further efficiency gains.

\section{Acknowledgements}

Bartosz Wójcik is supported by National Centre of Science (NCP, Poland) Grant No. 2023/49/N/ST6/02513. Simone Scardapane and Alessio Devoto are partially supported by Sapienza grant RG123188B3EF6A80 (CENTS).

We gratefully acknowledge Polish high-performance computing infrastructure PLGrid (HPC Center: ACK Cyfronet AGH) for providing computer facilities and support within computational grant no. PLG/2024/017202.

The contribution of Bartosz Wójcik to this research was conducted at the Faculty of Mathematics and Computer Science, and the Doctoral School of Exact and Natural Sciences of the Jagiellonian University.

\bibliography{refs}

\end{document}


\section{Extended Related Work}

\paragraph{Conditional computation} Conditional computation \citep{bengio2013estimating,bengio2015conditional} (CC) refers to a broad class of algorithms that make neural networks more efficient by conditioning their computational graph on the specific input, under the hypothesis that inputs with different complexity can require different amounts of compute. This is in contrast to methods such as quantization \citep{courbariaux2014training,wu2020integer,dettmers2023case}, pruning \citep{hoefler2021sparsity}, or knowledge distillation \citep{hinton2015distilling,aguilar2020knowledge,gou2021knowledge}, in which a network is replaced entirely by another network having a smaller cost, potentially hindering its performance whenever more flexibility is needed. In principle, networks with CC mechanisms can dynamically adapt their execution on a per-input or per-token basis, significantly reducing their average cost while maintaining performance. Notable examples of CC techniques include dynamic channel selection in CNNs \citep{chen2019you,li2021dynamic}, spatial masking of convolutions \citep{verelst2020dynamic}, variable-sized kernels \citep{verelst2022segblocks}, conditional execution of layers \citep{graves2016adaptive,ainslie2023colt5}, dynamic input resizing \citep{wang2021not,zhu2021dynamic}, or combinations of them \citep{xia2021fully,meng2022adavit,abnar2023adaptivity}. The majority of these techniques, however, rely on specific architectural choices, requiring a significant redesign of the model or a separate retraining step. In addition, theoretical speedups are challenging to achieve on modern hardware because the resulting sparsity patterns tend to be unstructured \citep{li2021dynamic}. In contrast, our proposed ACM execution is fully parallelized and provides significant gains without requiring complicated GPU kernels.

\paragraph{Early-exit Models} Thanks to their simplicity, early-exit (EE) models are a popular class of CC techniques \citep{teerapittayanon2016branchynet,bolukbasi2017adaptive}. In an EE model, inputs are allowed to `exit' the architecture at intermediate layers via additional classifiers taking as input the corresponding hidden representations. These auxiliary blocks can be trained together with the backbone network \citep{teerapittayanon2016branchynet}, or as a separate step in a layerwise fashion \citep{han2021dynamic}. Predictions at different levels can be differentiated based on their confidence \citep{teerapittayanon2016branchynet}, or combined via geometric \citep{wolczyk2021zero} or trainable \citep{scardapane2020differentiable} ensembling. EE models have been applied in a number of contexts including language modeling \citep{schuster2022confident} and vision-language applications \citep{tang2023you}. Finally, model cascades could be seen as a special case of an EE model \citep{lebovitz2023efficient,wangidk,enomoro2021learning}. While pre-trained models can be converted to EE models by separately training the auxiliary branches, designing and placing the exits is itself a non-trivial task \citep{teerapittayanon2016branchynet}, and the majority of solutions are specialized to classification problems. By contrast, the proposed ACM model can be applied seamlessly to any pre-trained Transformer architecture.

\paragraph{Token Dropping} Standard EE models stop the processing of the entire input (e.g., image) at once. Separately from this, multiple \textit{token dropping} strategies have been devised to prematurely stop execution on tokens that are deemed less relevant or redundant \citep{rao2021dynamicvit,yin2022vit,meng2022adavit,haurum2023tokens}. Roughly, most token dropping models dynamically allocate variable network depths to each token. Our proposed ACM can be seen as an orthogonal \textit{width} variant of token dropping, in which each token is allocated a variable \textit{width} for each layer in the network. Some authors have also investigated token merging (instead of dropping) to reduce the number of processed tokens at each block \cite{bolya2022token,bolya2023token,renggli2022learning}.

\paragraph{Mixture-of-Experts} Another closely connected CC technique is mixture-of-experts (MoEs) \citep{yuksel2012twenty,shazeer2017outrageously,fedus2022review}. In a MoE, specific parts of a model are replaced by a block-wise architecture and a routing module that selectively activates specific blocks (`experts') for each token \citep{puigcerver2023sparse}. MoEs have been successfully developed for replacing MLP layers in transformers \citep{shazeer2017outrageously,riquelme2021scaling}, attention layers \citep{zhang2022mixture}, entire blocks \cite{tan2023sparse}, and adapters \citep{zadouri2023pushing}. In the majority of cases routing is done by activating the top-$k$ most similar experts for each token \citep{riquelme2021scaling}, but many other choices have been proposed including linear assignment problems \citep{dai2022stablemoe}, expert choice routing \cite{zhou2022mixture}, BASE layers \citep{lewis2021base}, and even soft combinations of inputs \citep{puigcerver2023sparse} or experts' weights \citep{muqeeth2023soft}. Also, most MoE models are designed to be trained from scratch, although a few works investigated \emph{MoEfication} procedures \citep{zhang2021moefication,qiu2023emergent}, or fine-tuning with MoEs blocks \citep{zadouri2023pushing}. While MoEs and the proposed ACM bear some similarities, the design principles of ACMs are significantly different since each learner acts on the output of previous learners. As a result, in MoEs, the computational budget for each token is fixed and depends on the routing strategy (e.g., for top-$k$ routing, exactly $k$ experts will activate). At the same time, ACMs exploit the learners' ordering to allocate a variable number of them to each token. Finally, the simple design of ACM does not suffer from representation collapse issues \citep{chi2022representation}, a common problem in MoEs. As a side note, in the limit of replacing individual linear layers (e.g., key-value projections in attention) with smaller linear experts, our work connects to the broader topic of matrix decomposition using low-rank terms, which is gaining prominence in parameter-efficient fine-tuning \cite{hu2021lora,lialin2023stack}. However, we empirically validate that using small MLPs as learners is also beneficial for replacing linear blocks.

\paragraph{Discrete Latent Learning} In general, CC approaches (including EE and MoE layers) are examples of the emerging field called \textit{modular deep learning} \citep{pfeiffer2023modular}, aiming at designing neural networks with more flexible, decomposable architectures for improving efficiency, interpretability, and re-use. A fundamental task in modular NNs is how to take discrete decisions (e.g., sampling experts in MoEs) in a differentiable way for end-to-end training \citep{niculae2023discrete}. When sampling from a categorical distribution, this can be done quickly via the Gumbel-softmax estimator \citep{jang2016categorical} (also known as the concrete distribution \citep{maddison2016concrete}), with similar relaxations existing for top-$k$ sampling \citep{kool2019stochastic,ahmed2022simple}. However, sampling a generic element from the power set of a set is a combinatorial problem, requiring specialized solutions \citep{niepert2021implicit}. In the proposed ACM, we sidestep this issue by adequately ordering the learners so that sampling a subset of them can be cast as a more straightforward interval selection problem.

\section{Training Details}

In Tab. \ref{tab:vision_hyperparameters}, we list the hyperparameters that we used for computer vision experiments. We use the augmentation introduced by  \citet{touvron2022deit}, along with mixup ($\alpha_\text{mixup} = 0.8$) \citep{zhang2017mixup}, cutmix ($\alpha_\text{cutmix} = 0.8$) \citep{yun2019cutmix} and label smoothing ($\epsilon = 0.1$) \citep{szegedy2016rethinking}. Each ACM consists of $4$ learners and has approximately the same computational cost as the substituted module. We replace every linear projection of each MHA layer and each FFN module with an ACM. Due to residual connections, in every ACM that replaced a FFN module, we allow the gating network to activate 0 learners. In contrast, ACMs that replaced projections in MHA layers always have to execute at least one learner for each token.

\begin{table}[h!]
    \centering
    \caption{Hyperparameters used in computer vision experiments presented in the main paper.}
    \begin{tabular}{ccccc}
        \toprule
          & Phase I & Phase II & Phase III \\
         \midrule
         \verb|epochs| & 2 & 1 & 97 \\
         $|B|$ & 64 & 64 & 512 \\
         \verb|optimizer| & Adam & Adam & Adam \\
         $\eta$ & $10^{-3}$ & $10^{-2}$ & $10^{-4}$ \\
         \verb|scheduler| & cosine & cosine & cosine \\
         $\eta_\text{min}$ & $10^{-6}$ & $10^{-6}$ & $10^{-6}$ \\
         \verb|clip_grad_norm| & 1.0 & 1.0 & 1.0 \\
         \verb|T| & - & 0.8 & 0.8 \\
         $\tau$ & - & 1.2 & - \\
         $\beta_\text{target}$ & - & - & 
         \makecell{$\{0.25, 0.40, $ \\ $0.60, 0.75\}$}\\
         $\alpha_b$ & - & - & 0.1 \\
         $\alpha_d$ & - & - & 0.05 \\
         $\alpha_e$ & - & - & 0.05 \\
         \bottomrule
    \end{tabular}
    \label{tab:vision_hyperparameters}
\end{table}

\begin{table}[h!]
    \centering
    \caption{Hyperparameters used in speech-to-text experiments presented in the main paper.}
    \begin{tabular}{ccccc}
        \toprule
          & Phase I & Phase II & Phase III \\
         \midrule
         \verb|epochs| & 5  & 1  & 4  \\
         $|B|$ & 12 & 16  & 16 \\
         \verb|optimizer| & Adam & Adam & Adam \\
         $\eta$ & $3*10^{-4}$ & $3*10^{-3}$ & $3*10^{-4}$ \\
         \verb|scheduler| & cosine & cosine & cosine \\
         $\eta_\text{min}$ & $10^{-6}$ & $10^{-6}$ & $10^{-6}$ \\
         \verb|clip_grad_norm| & 1.0 & 1.0 & 1.0 \\
         \verb|T| & - & 0.8 & 0.8 \\
         $\tau$ & - & 1.2 & - \\
         $\beta_\text{target}$ & - & - & 
         \makecell{$\{0.25, 0.40, $ \\ $0.60, 0.75\}$}\\
         $\alpha_b$ & - & - & 0.1 \\
         $\alpha_d$ & - & - & 0.05 \\
         $\alpha_e$ & - & - & 0.05 \\
         \bottomrule
    \end{tabular}

    \label{tab:stt_hyperparameters}
\end{table}

We list the hyperparameters used for speech-to-text experiments in Tab. \ref{tab:stt_hyperparameters}. This time we replace only the FFN modules, but we also allow for 0 learners to be executed on some tokens.

\begin{figure*}
  \centering
  \begin{minipage}[c]{.5\linewidth}
    \centering\subfloat[End-to-end fine-tuning training curves]
      {\includegraphics[width=\linewidth]{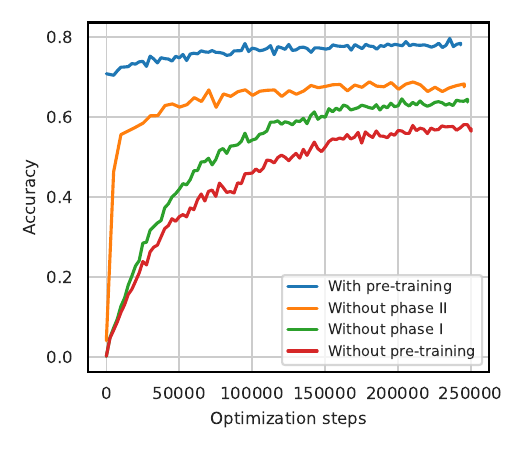}
      \label{fig:pretraining_impact_training_curves}}%
  \end{minipage}%
  \hfill
  \begin{minipage}[c]{.5\linewidth}
    \centering\subfloat[Performance-cost result]
      {\includegraphics[width=\linewidth]{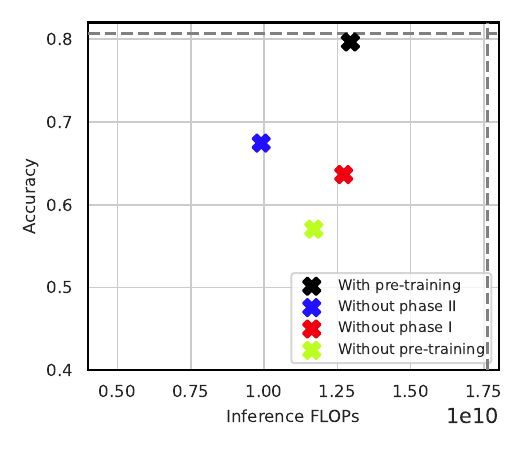}
      \label{fig:pretraining_cost_vs_acc}}%
  \end{minipage}%
  \caption
    {%
      Analysis of the impact of the proposed three-phase training. Applying both pre-training stages significantly accelerates training in comparison to training the model only with end-to-end fine-tuning.
      \label{fig:pretraining_impact}%
    }%
\end{figure*}

The experiments were run on A100 NVIDIA GPUs, NVIDIA Driver Version: \verb|525.125.06|, CUDA Version: \verb|12.0|, PyTorch \verb|2.3.1|, torchvision \verb|0.18.1|, Python \verb|3.11.9| on Ubuntu Linux 20.04.6.

Given the number of training runs for various hyperparameter values and the large model sizes, we conduct each experiment using a single seed.

\section{Impact of smart weight initialization}
One might hypothesize that the module-wise representation distillation and gating networks pre-training phases are not necessary. To show their usefulness, we compare our ACMized ViT-B training run with $\beta_\text{target} = 0.6$ from the main paper to three alternative training approaches: (1) end-to-end finetuning only, (2) without gating network pre-training, and (3) without module-wise representation distillation stage. In all cases a cosine learning scheduler was used, and the total number of training epochs remains constant (100 epochs). We always use 2 epochs for module-wise representation-distillation and 1 epoch for gating network pre-training, if they are present. In Fig. \ref{fig:pretraining_impact} we report results of this experiment. Pre-trained gating networks enable faster training in comparison to fine-tuning with randomly initialized parameters, and the same effect occurs with untrained learners. Both stages together provide the intended effect of significantly accelerated training.

\section{Choosing $N$}
\begin{figure}
    \centering
    \begin{minipage}[c]{\columnwidth}
    \centering
      {\includegraphics[width=\linewidth]{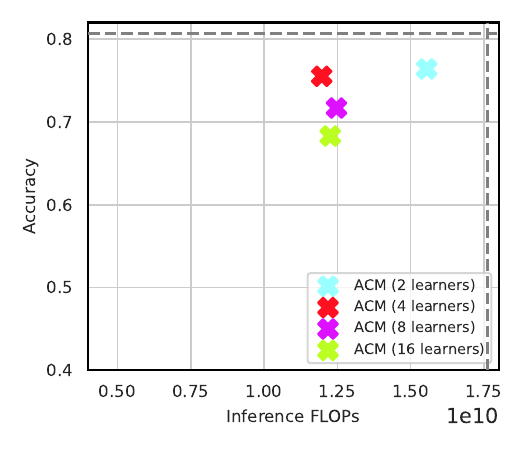}}
    \end{minipage}%
    \caption{Final performance-efficiency trade-offs of ACMized models with different $N$ after finetuning them for $5$ epochs.}
    \label{fig:cost_vs_acc_granularity}
\end{figure}

Due to space constraints, in the main paper we did not explain the reasons for our selection of number of learners $N$ hyperparameter, which we originally set to $4$. Here we present a short experiment justifying our selection of this value. We perform the conversion process for different number of learners $N \in \{2, 4, 8, 16\}$ on the same ViT-B model as in our main experiments, with the only difference being that we finetune (third stage) them only for $5$ epochs instead of $97$ epochs. As before, the total size in parameters and the computational cost of an ACM is always kept approximately close to the original substituted module. Tab. \ref{tab:n_choice_distillation_accuracies} lists the results after the module-wise representation distillation stage, while Fig. \ref{fig:cost_vs_acc_granularity} presents the final results after finetuning. We highlight two visible effects: (1) fine-grained learners (higher $N$) have downgraded performance in comparison to coarse-grained learners, and (2) too low $N$ reduces the flexibility of the model in terms of adjustment its computational budget. We pick $N=4$ as it exhibits the best trade-off that balances these two effects.

\begin{table}[]
    \centering
    \caption{ImageNet-1k accuracy scores of ACMized models after first phase of pre-training for different number of learners $N$ that each ACM consists of.}
    \begin{tabular}{ccccc}
        \toprule
         N & $\frac{k}{N}=0.25$ & $\frac{k}{N}=0.5$ & $\frac{k}{N}=0.75$ & $\frac{k}{N}=1.0$ \\
         \midrule
         2 & - & 68.37\% & - & 75.18\% \\
         4 & 41.49\% & 65.37\% & 70.65\% & 71.98\% \\
         8 & 37.62\% & 62.53\% & 67.43\% & 68.84\% \\
         16 & 33.38\% & 56.61\% & 62.05\% & 63.78\% \\
         \bottomrule
    \end{tabular}
    \label{tab:n_choice_distillation_accuracies}
\end{table}

\section{Efficient GPU implementation details}
In Listing 1 we present the pseudocode for our efficient implementation of the forward pass along with the pseudocode for the kernel of the first layer of a learner. We skip the pseudocode of the kernel for the second layer as it is almost identical. Full source code is also provided in this supplementary material.

\definecolor{codegreen}{rgb}{0,0.6,0}
\definecolor{codegray}{rgb}{0.5,0.5,0.5}
\definecolor{codepurple}{rgb}{0.58,0,0.82}
\definecolor{backcolour}{rgb}{0.95,0.95,0.92}

\lstdefinestyle{mystyle}{
    backgroundcolor=\color{backcolour},   
    commentstyle=\color{codegreen},
    keywordstyle=\color{magenta},
    numberstyle=\tiny\color{codegray},
    stringstyle=\color{codepurple},
    basicstyle=\ttfamily\footnotesize,
    breakatwhitespace=false,         
    breaklines=true,                 
    captionpos=b,                    
    keepspaces=true,                 
    numbers=left,                    
    numbersep=5pt,                  
    showspaces=false,                
    showstringspaces=false,
    showtabs=false,                  
    tabsize=2
}

\lstset{style=mystyle}

\begin{figure*}
    \begin{lstlisting}[language=Python, caption=Code for selected fragments our efficient GPU implementation of ACM]
def forward_triton(self, x, gating_network_outputs):
    sorted_gating_decisions, sort_indices = self.sort_indices(gating_network_outputs)
    intermediate_acts = AcmFirstLayerImplementation.apply(x, self.w1, self.b1, sorted_gating_decisions, sort_indices)
    final_out = AcmSecondLayerImplementation.apply(intermediate_acts, self.w2, sorted_gating_decisions, sort_indices)
    return final_out

@torch.compile(fullgraph=True, dynamic=True, mode='max-autotune')
def sort_indices(self, gating_network_outputs):
    with torch.no_grad():
        gating_decisions = gating_network_outputs.argmax(dim=-1)
        sorted_gating_decisions, sort_indices = gating_decisions.sort(dim=0)
    return sorted_gating_decisions, sort_indices

@triton.jit
def acm_first_layer_blocked_kernel(x_ptr, weight_ptr, bias_ptr, output_ptr, gating_decisions_ptr, sort_indices_ptr, sample_dim, hidden_dim, learner_dim, NUM_LEARNERS: tl.constexpr, BLOCK_SIZE_BD: tl.constexpr, BLOCK_SIZE_LD: tl.constexpr, BLOCK_SIZE_HD: tl.constexpr):
    # get program IDs
    pid_bd, pid_ld = ...
    # bd - batch dim, hd - hidden dim, nld - num learners dim, ld - learner dim
    ...
    # load gating decisions for the current set of input tokens
    gating_decisions = tl.load(gating_decisions_ptr + offs_bd, mask=..., other=0) - 1
    # load the actual (unsorted) positions of input tokens in the input tensor
    in_data_indices = tl.load(sort_indices_ptr + offs_bd, mask=..., other=-1)
    # what is the maximum number of selected learners for the current tile?
    max_gating_decision = tl.max(gating_decisions)
    # iterate over learners up to the max
    for learner_index in range(0, max_gating_decision + 1):
        # select only those input tokens that need to have the current learner executed
        samples_for_learner_mask = (learner_index <= gating_decisions)
        # using the above, calculate the memory location of inputs and weights
        x_ptrs = ...
        w_ptrs = ...
        # define a tensor variable for the result
        accumulator = tl.zeros((BLOCK_SIZE_BD, BLOCK_SIZE_LD), dtype=tl.float32)
        # tile-by-tile matrix multiplication
        for k in range(0, tl.cdiv(hidden_dim, BLOCK_SIZE_HD)):
            # load values from GPU global memory
            x = tl.load(x_ptrs, mask=..., other=0.0)
            w = tl.load(w_ptrs, mask=..., other=0.0)
            # matrix-multiply
            accumulator += tl.dot(x, w)
            # advance tensor of pointers after each tile
            x_ptrs += ...
            w_ptrs += ...
        # calculate the memory location of biases
        b_ptrs = ...
        # load and add biases to the result
        accumulator += tl.load(b_ptrs, mask=..., other=0.0).to(tl.float32)
        # apply the activation function
        accumulator = gelu(accumulator)
        ...
        # calculate the position for writing to output
        out_ptrs = ...
        # store the results to GPU global memory
        tl.store(out_ptrs, out, mask=...)
    \end{lstlisting}    
\end{figure*}

\section{Model wall-clock measurements}
\begin{figure}
    \centering
    \begin{minipage}[c]{\columnwidth}
    \centering
      {\includegraphics[width=0.8\linewidth]{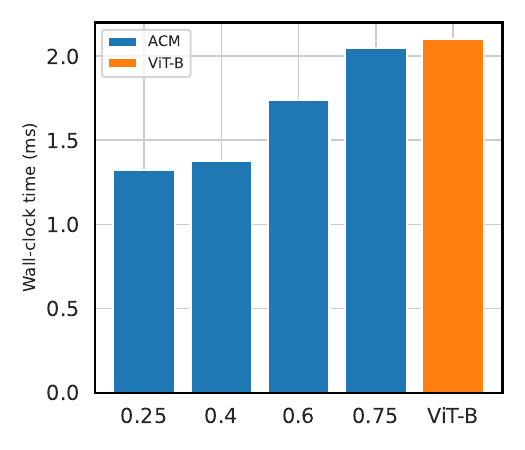}}
    \end{minipage}%
    \caption{Average wall-clock time of processing a ImageNet-1k sample for ACM-based ViT-B models from the main paper. Measurements taken on an A100 GPU using a batch size of $256$, and the $\beta_{\text{target}}$ hyperparameter value is displayed on the x-axis. ACMs reduce the overall average latency of the entire model.}
    \label{fig:wall_clock_model}
\end{figure}

In the main paper we measured latency of a single ACM layer in isolation. To show how ACMs affect the latency of the entire model, we present the wall-clock measurements of the four ACM-based ViT-B models in \Cref{fig:wall_clock_model}. ACMs can significantly reduce the wall-clock time of execution of a model.

\section{Robustness to distribution shifts}
In this section we explore how ACMized models behave under input data distribution shift. For this purpose we utilize the ImageNet-C \citep{hendrycks2019benchmarking} dataset, which contains images modified with multiple types of corruptions of different severity levels. 

\begin{figure*}
    \begin{minipage}[c]{\linewidth}
        \centering
        \includegraphics[width=0.5\linewidth]{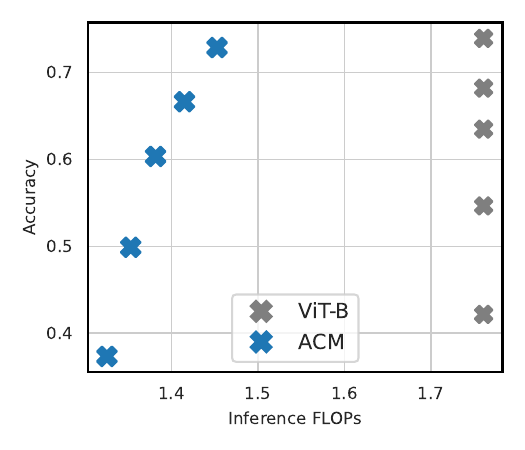}
    \end{minipage}
    \caption{FLOPs vs accuracy results of ViT-B and ACMized ViT-B on different ImageNet-C severity levels. Average compute and accuracy decrease with the increase of the severity of corruptions.}
    \label{fig:robustness_results}
\end{figure*}

\begin{figure*}
    \begin{minipage}[c]{\linewidth}
        \centering
        \includegraphics[width=0.33\linewidth]{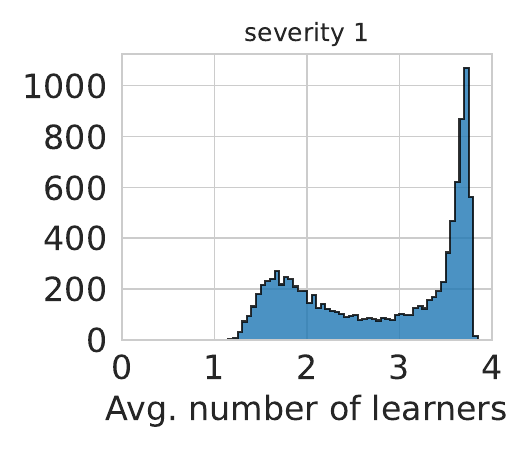}
        \hfill
        \includegraphics[width=0.33\linewidth]{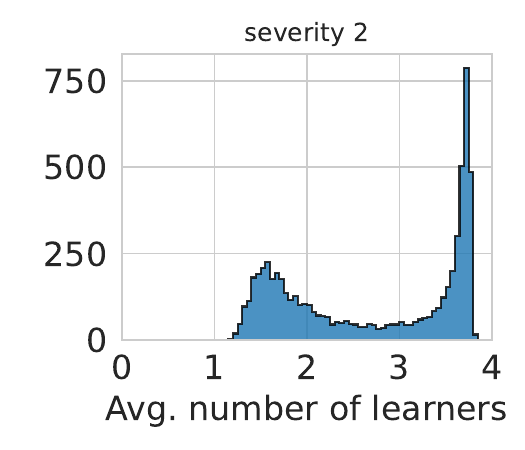}
        \hfill
        \includegraphics[width=0.33\linewidth]{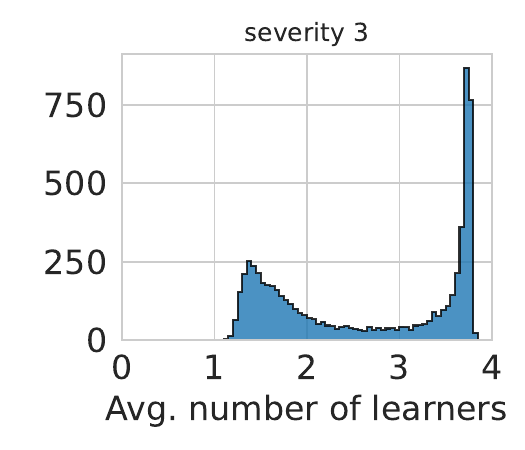}
    \end{minipage}
    \begin{minipage}[c]{\linewidth}
        \centering
        \includegraphics[width=0.33\linewidth]{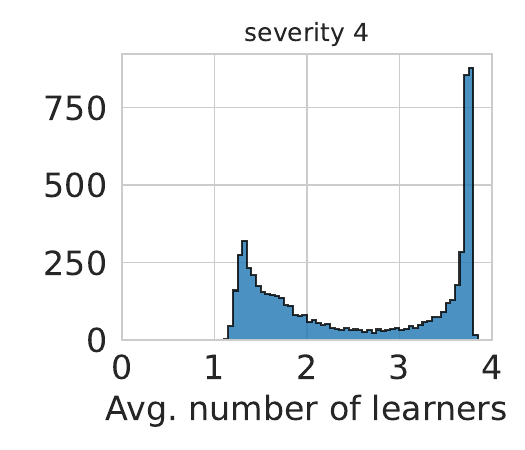}
        \includegraphics[width=0.33\linewidth]{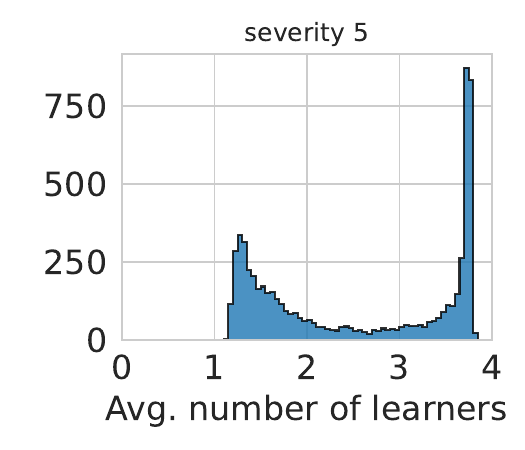}
        \hfill
    \end{minipage}
    \caption{Distribution of gating choices of an ACMized ViT-B model for different severity levels of distortions of samples from ImageNet-C. Distribution shift causes the gating networks to select either maximum or minimum number of learners more often.}
    \label{fig:robustness_gating}
\end{figure*}

The accuracy of both the static and ACMized models decreases as the distortions are heavier, which is shown in Figure \ref{fig:robustness_results}. While this is not surprising, the average computational effort of the dynamic model also changes.
In Figure \ref{fig:robustness_gating} we present the distribution of gating choices for different distortion severity levels. We can see that the dynamic model is sensitive to distributions shifts as distortions affect the routing choices of the gating networks, which are more extreme.

\section{Image classification computational load}
\begin{figure*}
    \begin{minipage}[c]{\linewidth}
        \centering
        \includegraphics[width=0.243\linewidth]{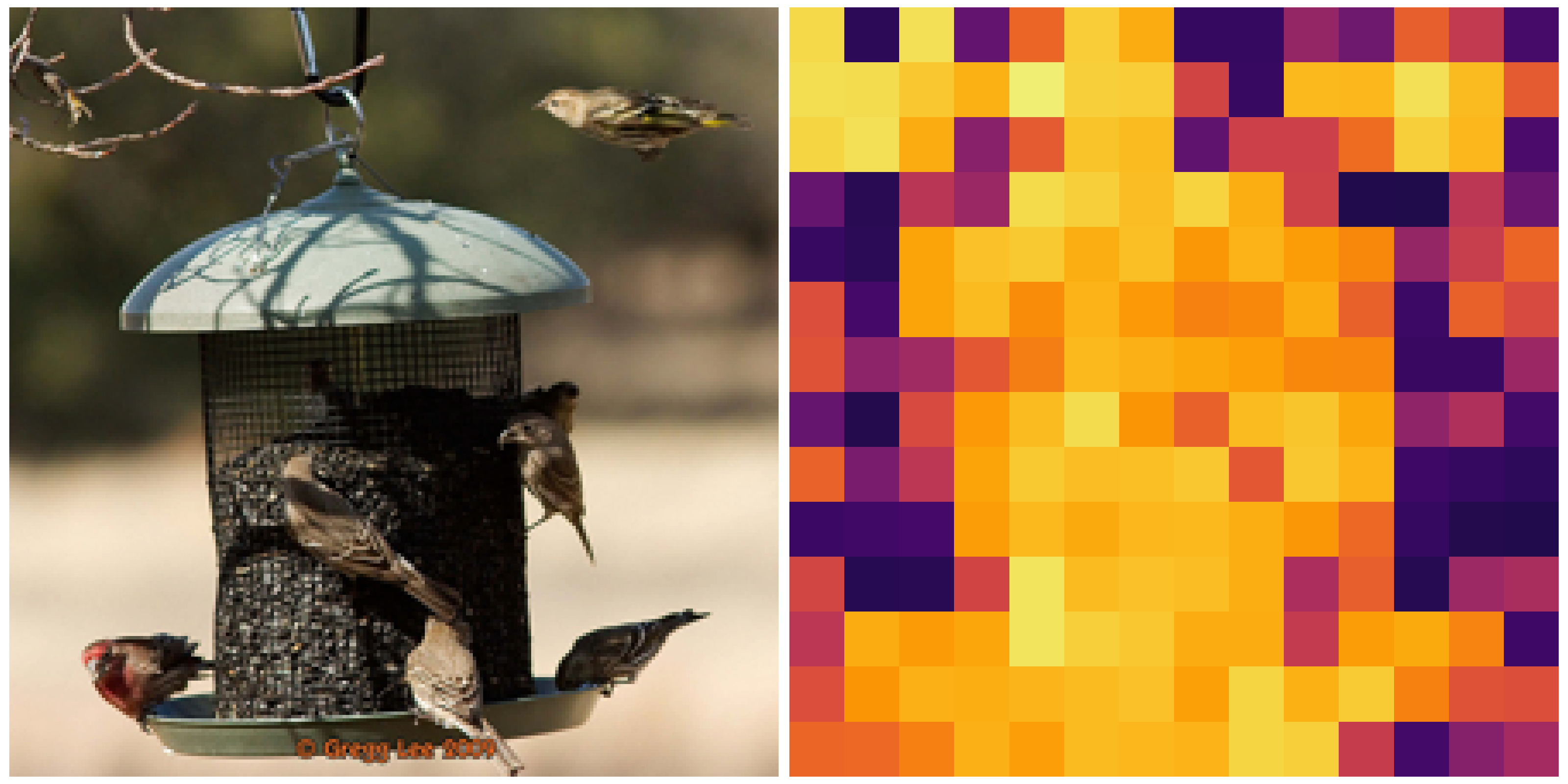}
        \hfill
        \includegraphics[width=0.243\linewidth]{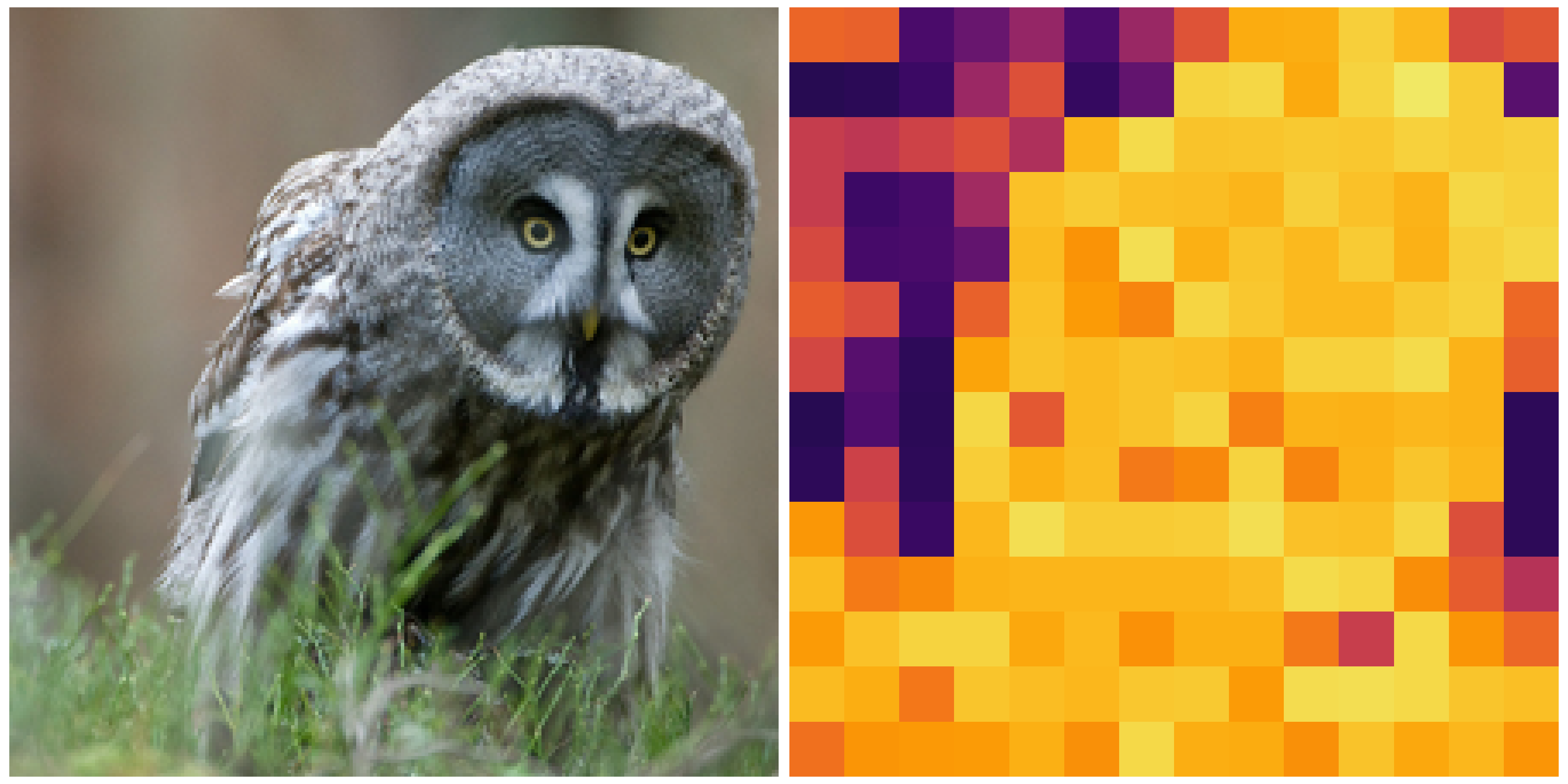}
        \hfill
        \includegraphics[width=0.243\linewidth]{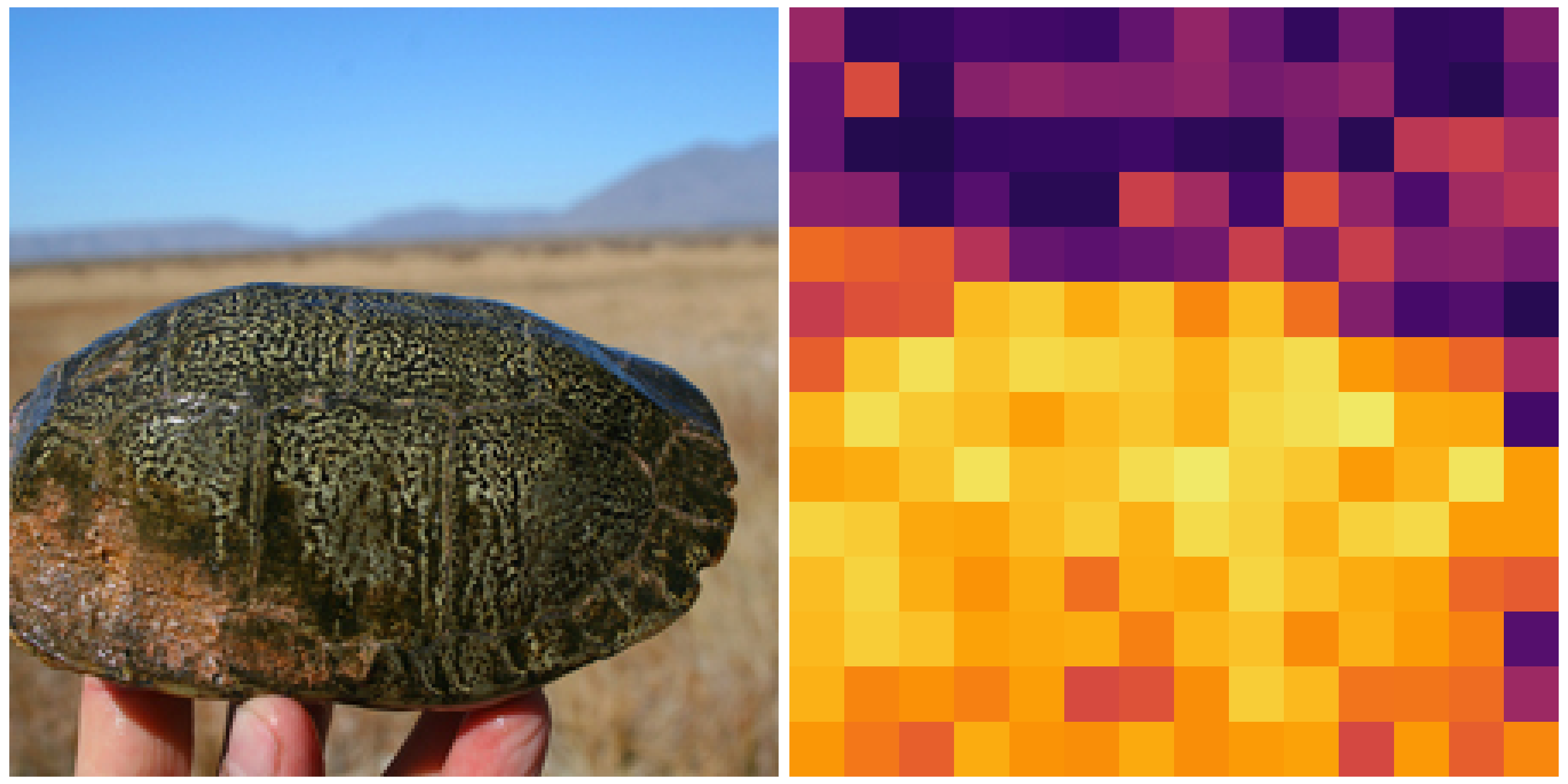}
        \hfill
        \includegraphics[width=0.243\linewidth]{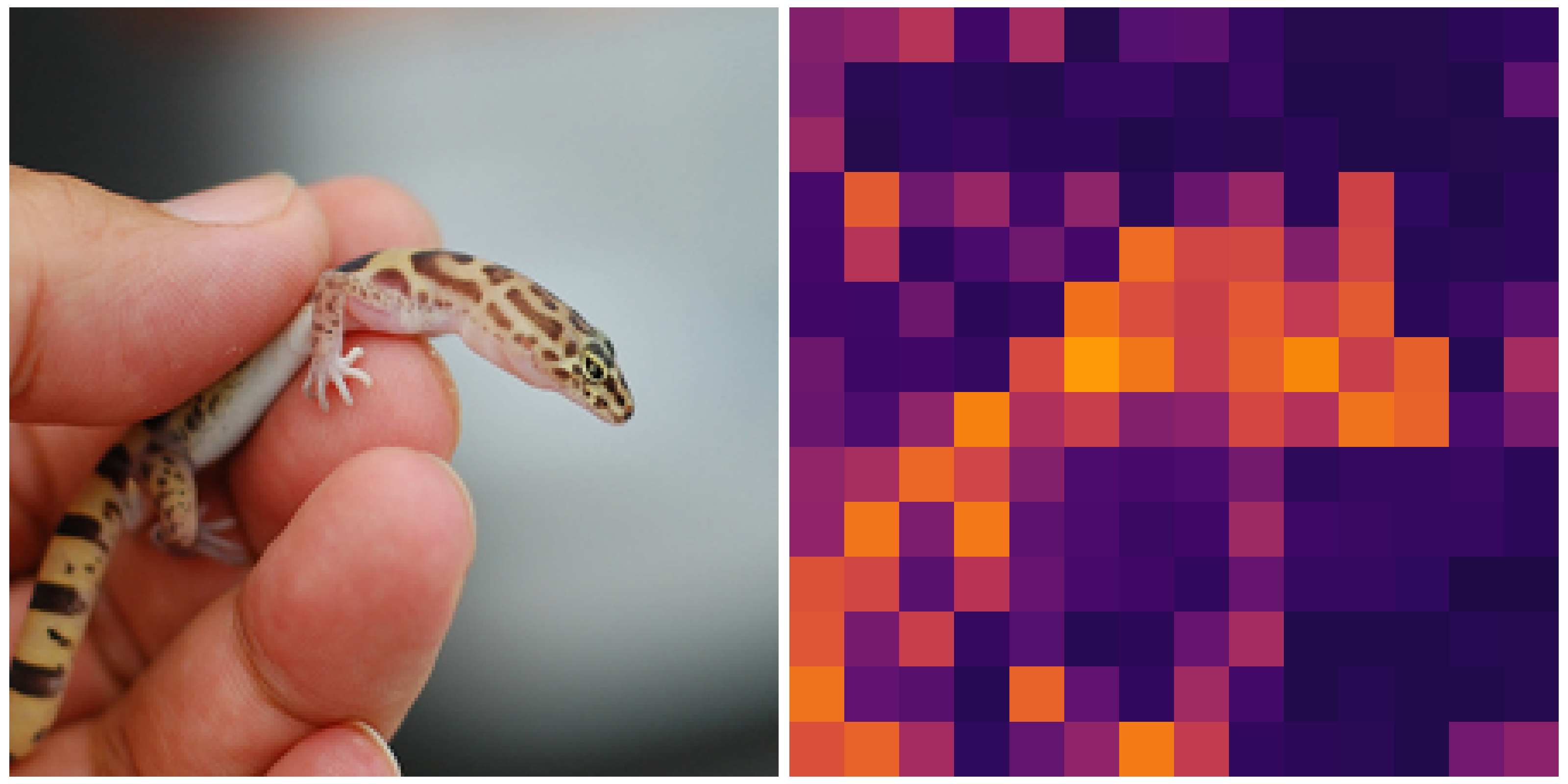}
    \end{minipage}
    \begin{minipage}[c]{\linewidth}
        \centering
        \includegraphics[width=0.243\linewidth]{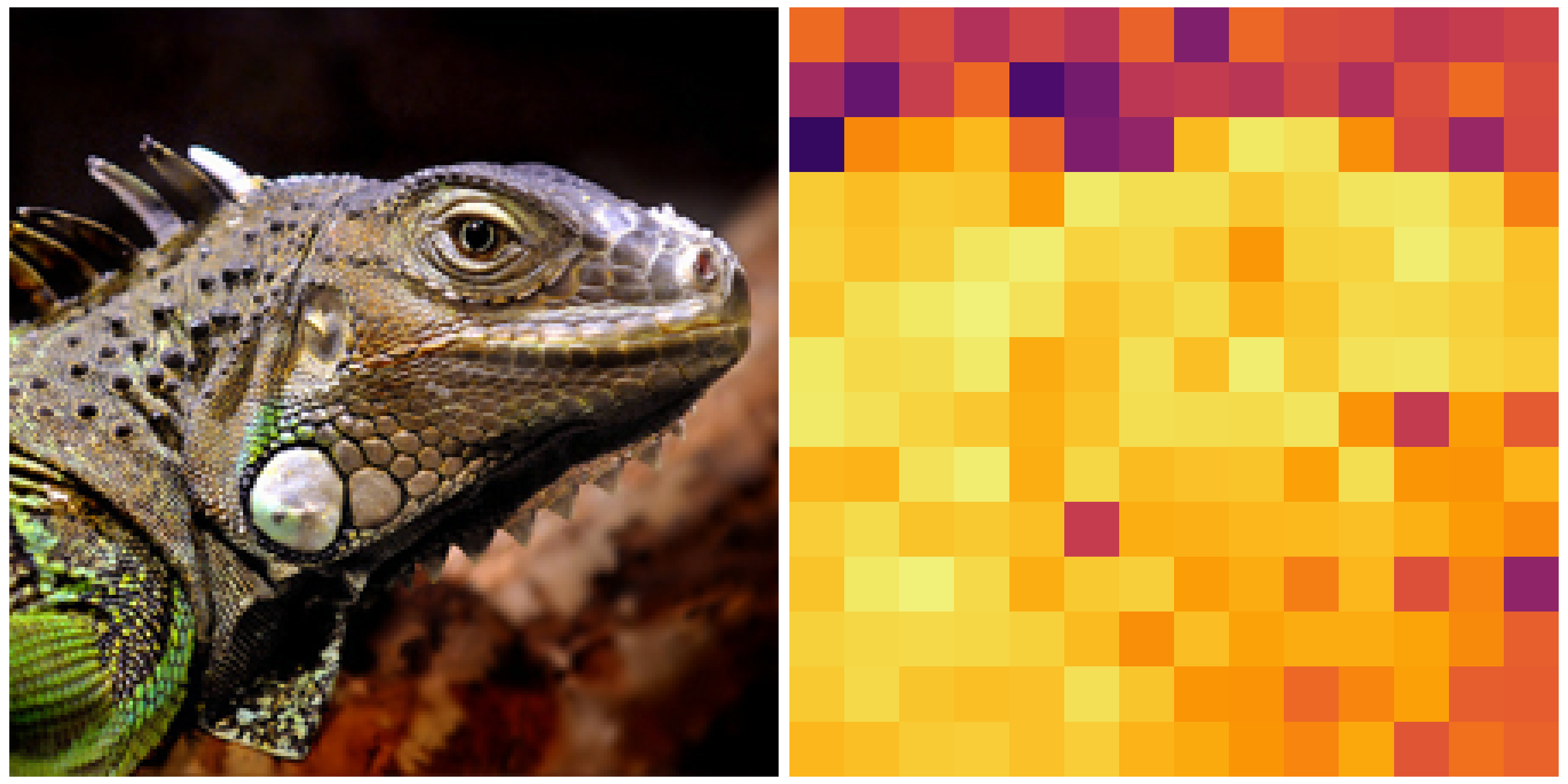}
        \hfill
        \includegraphics[width=0.243\linewidth]{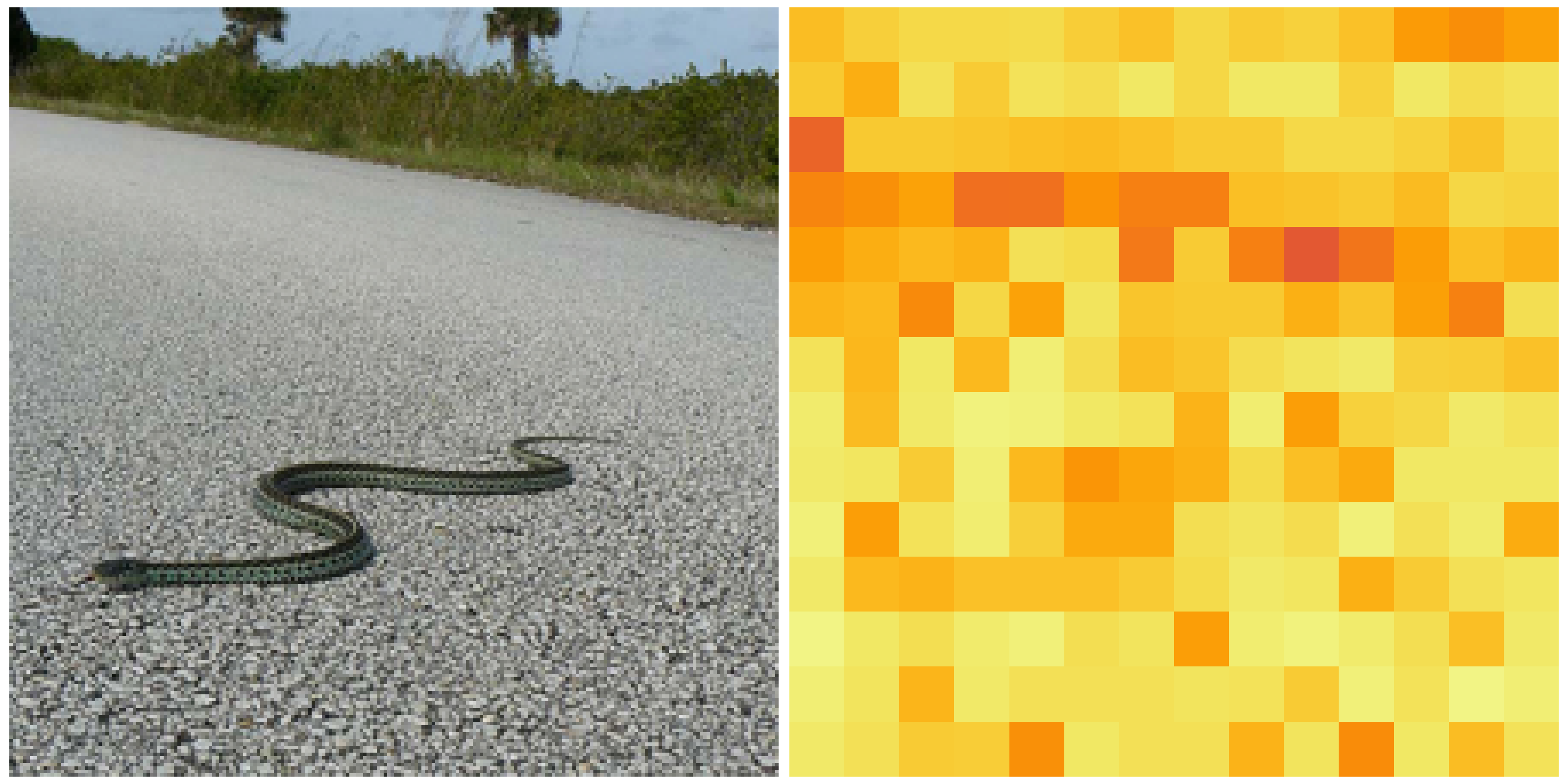}
        \hfill
        \includegraphics[width=0.243\linewidth]{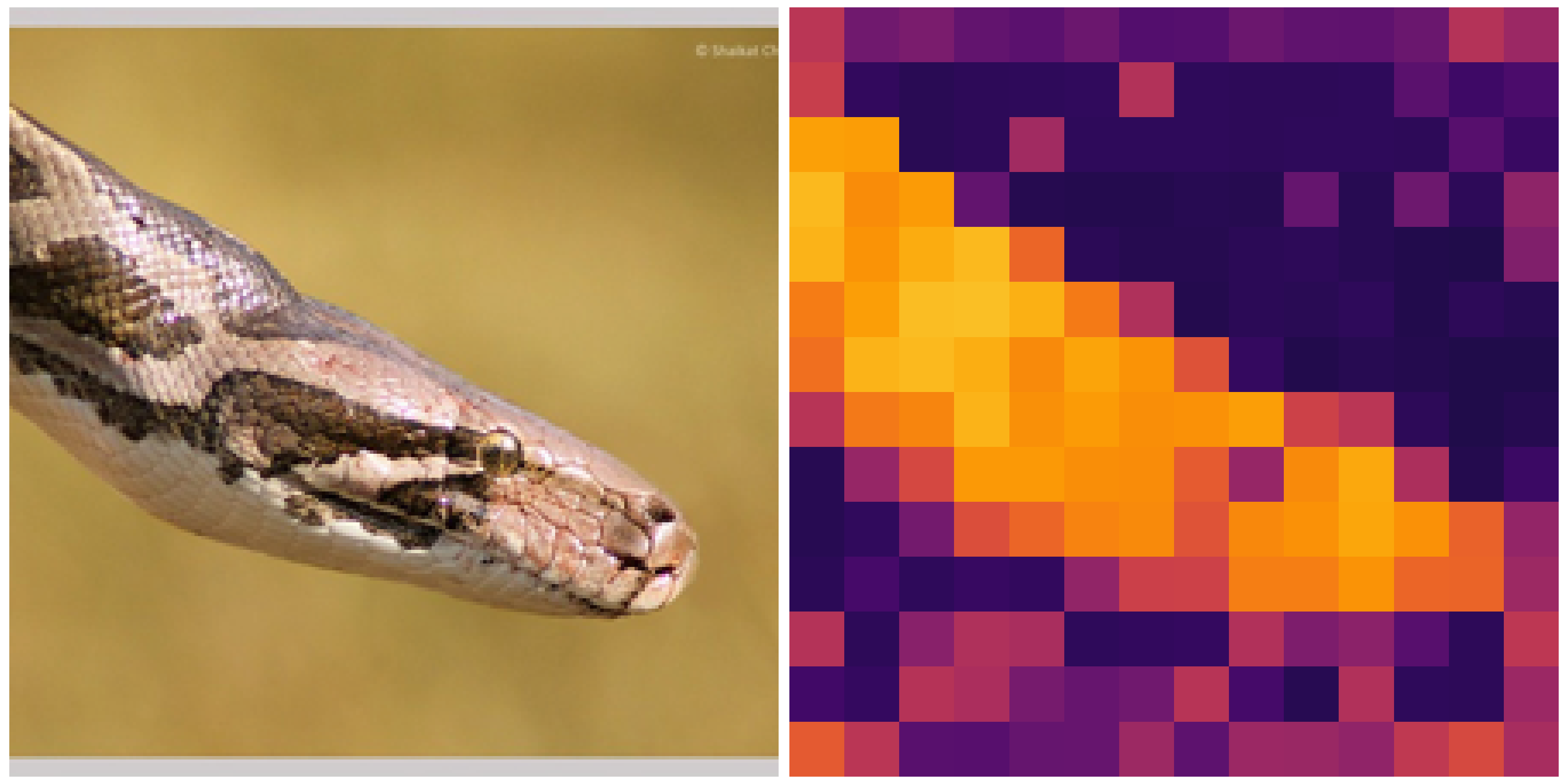}
        \hfill
        \includegraphics[width=0.243\linewidth]{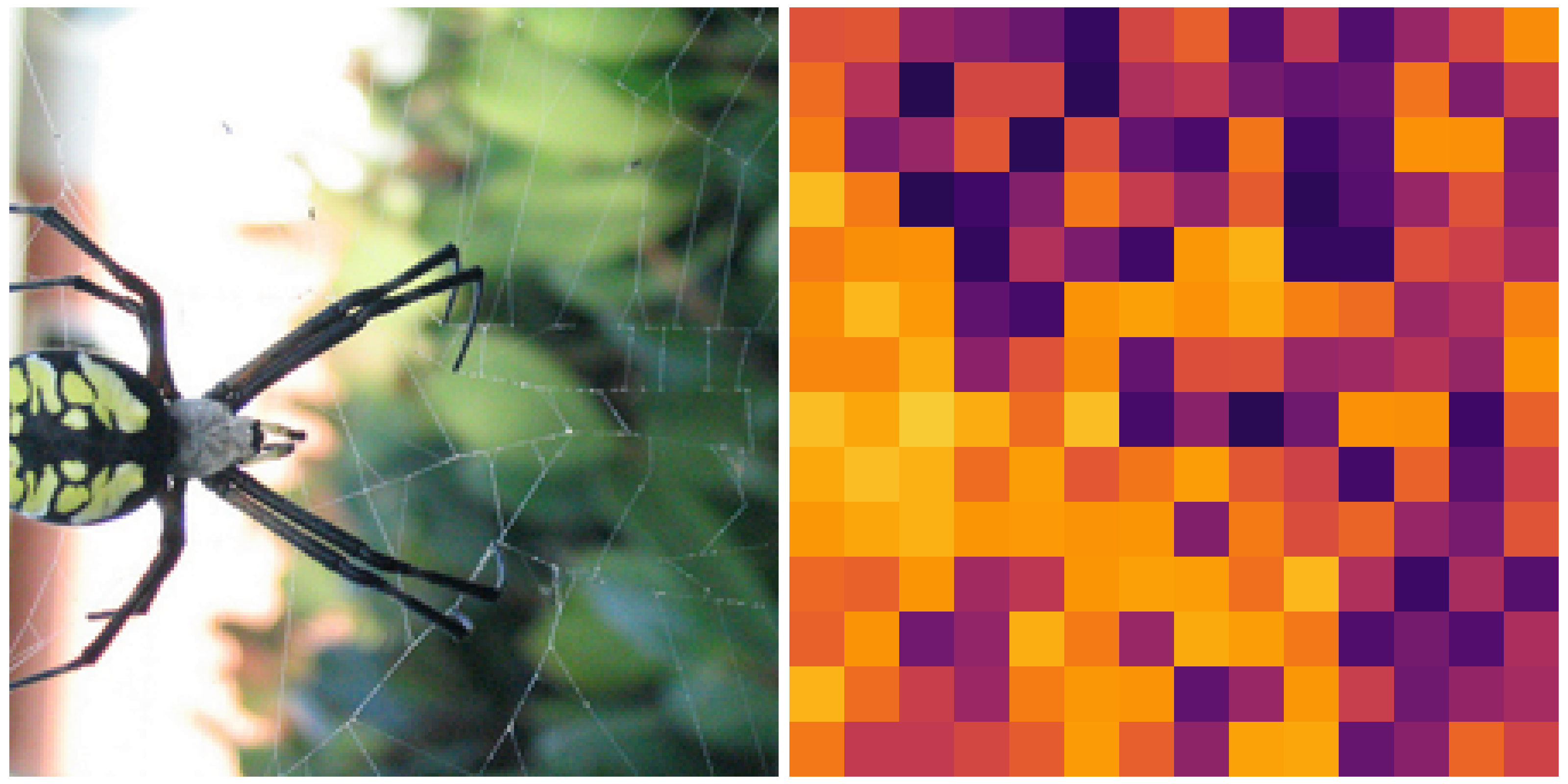}
    \end{minipage}
    \begin{minipage}[c]{\linewidth}
        \centering
        \includegraphics[width=0.243\linewidth]{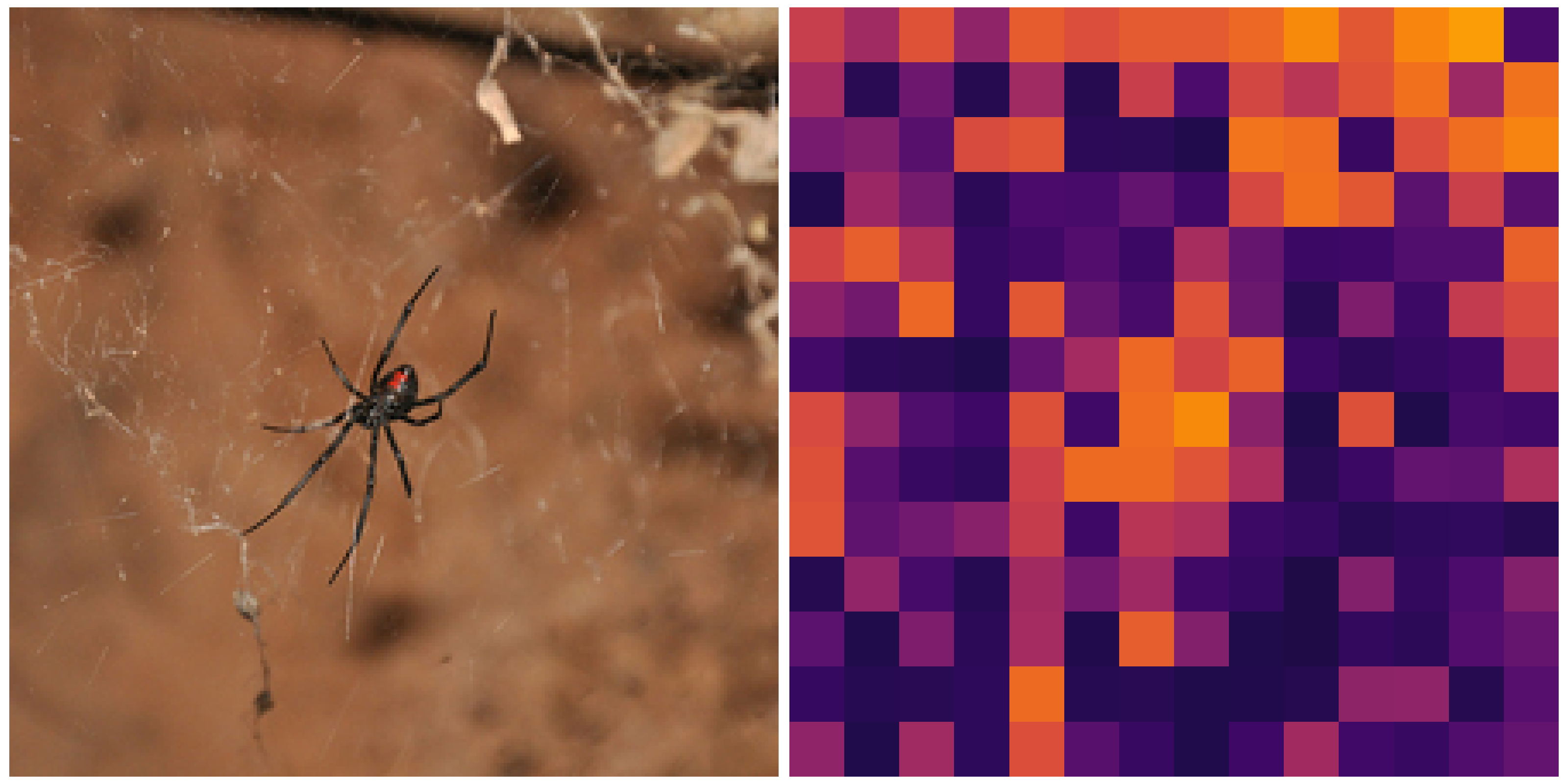}
        \hfill
        \includegraphics[width=0.243\linewidth]{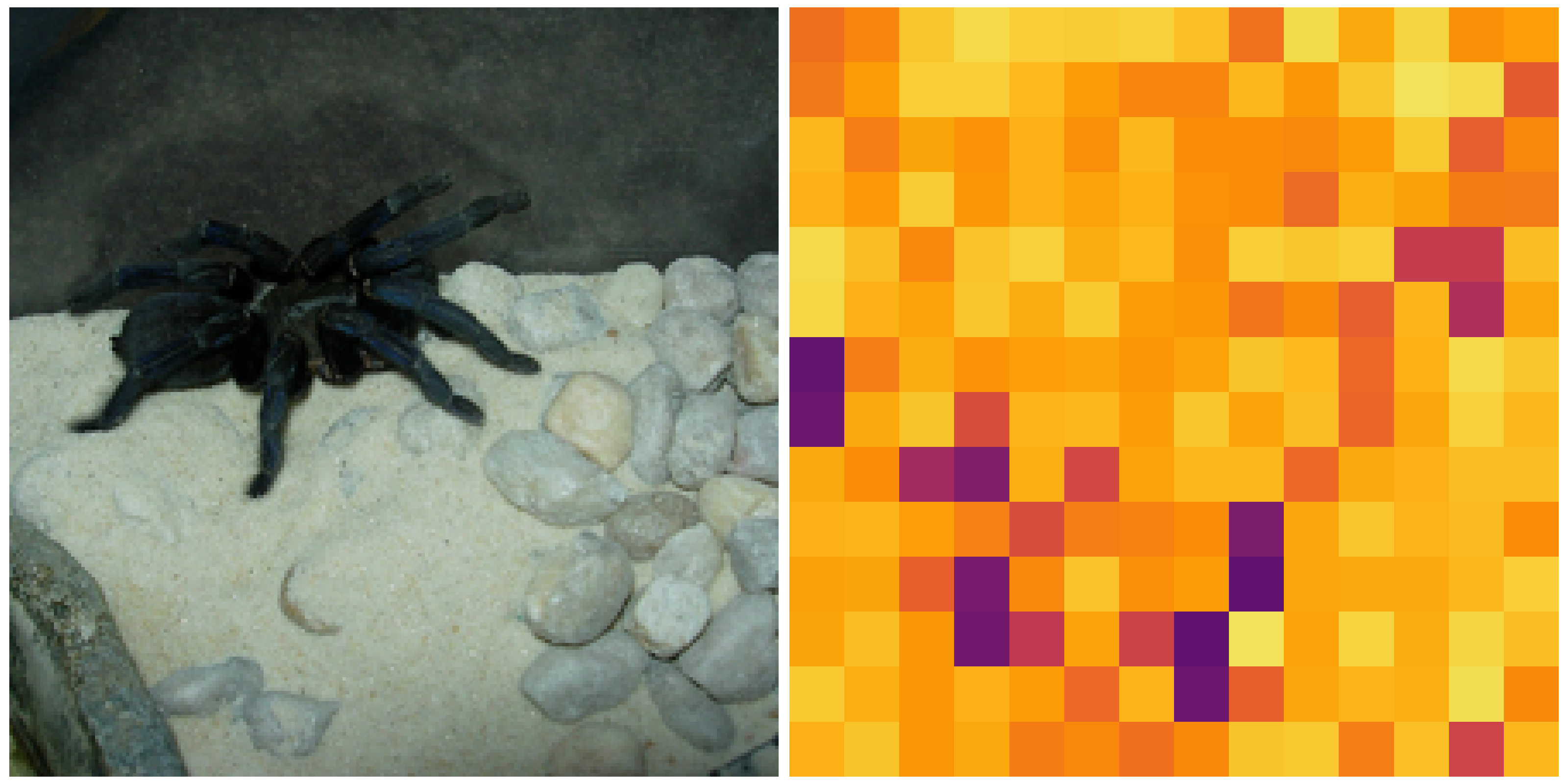}
        \hfill
        \includegraphics[width=0.243\linewidth]{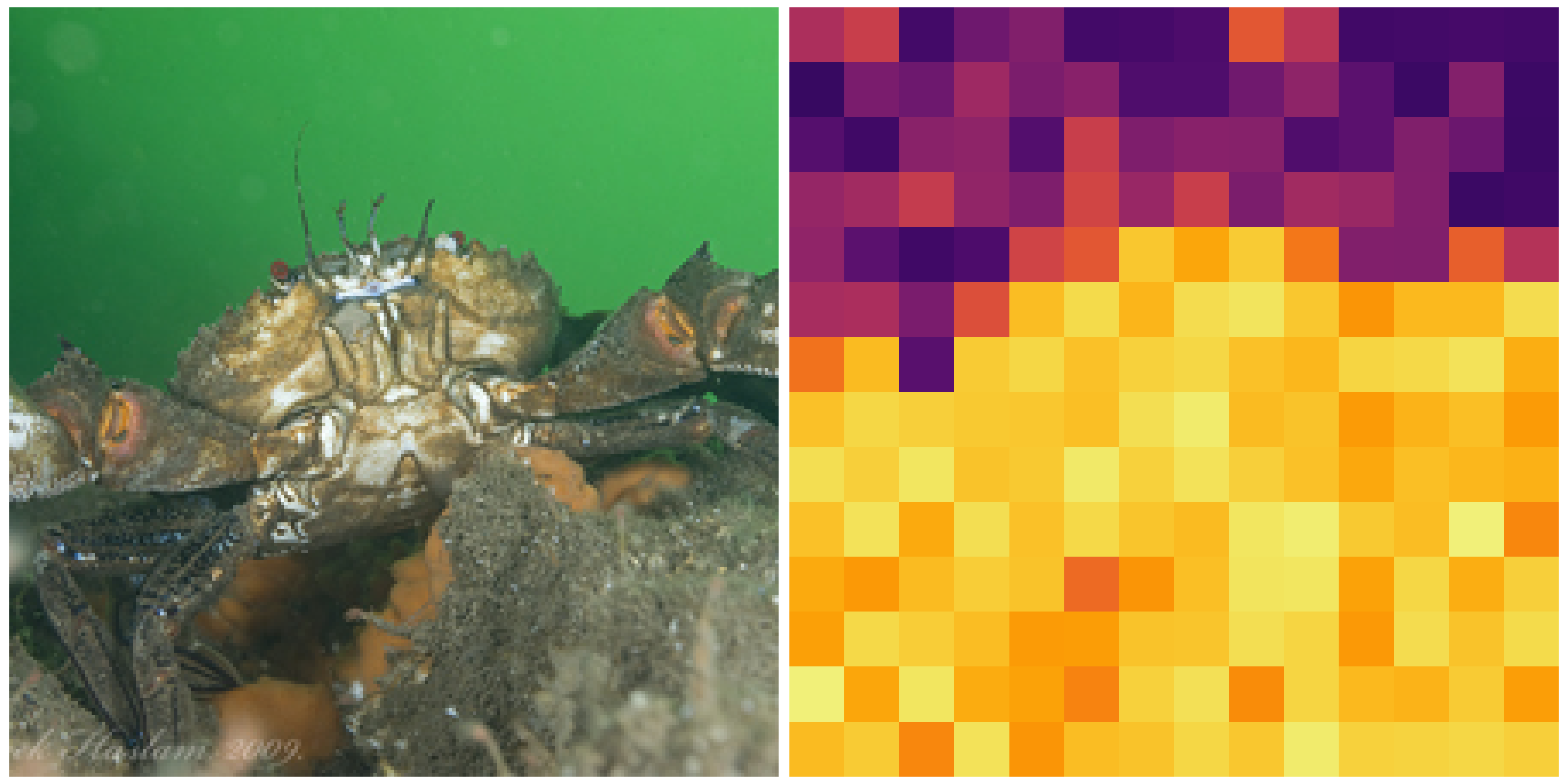}
        \hfill
        \includegraphics[width=0.243\linewidth]{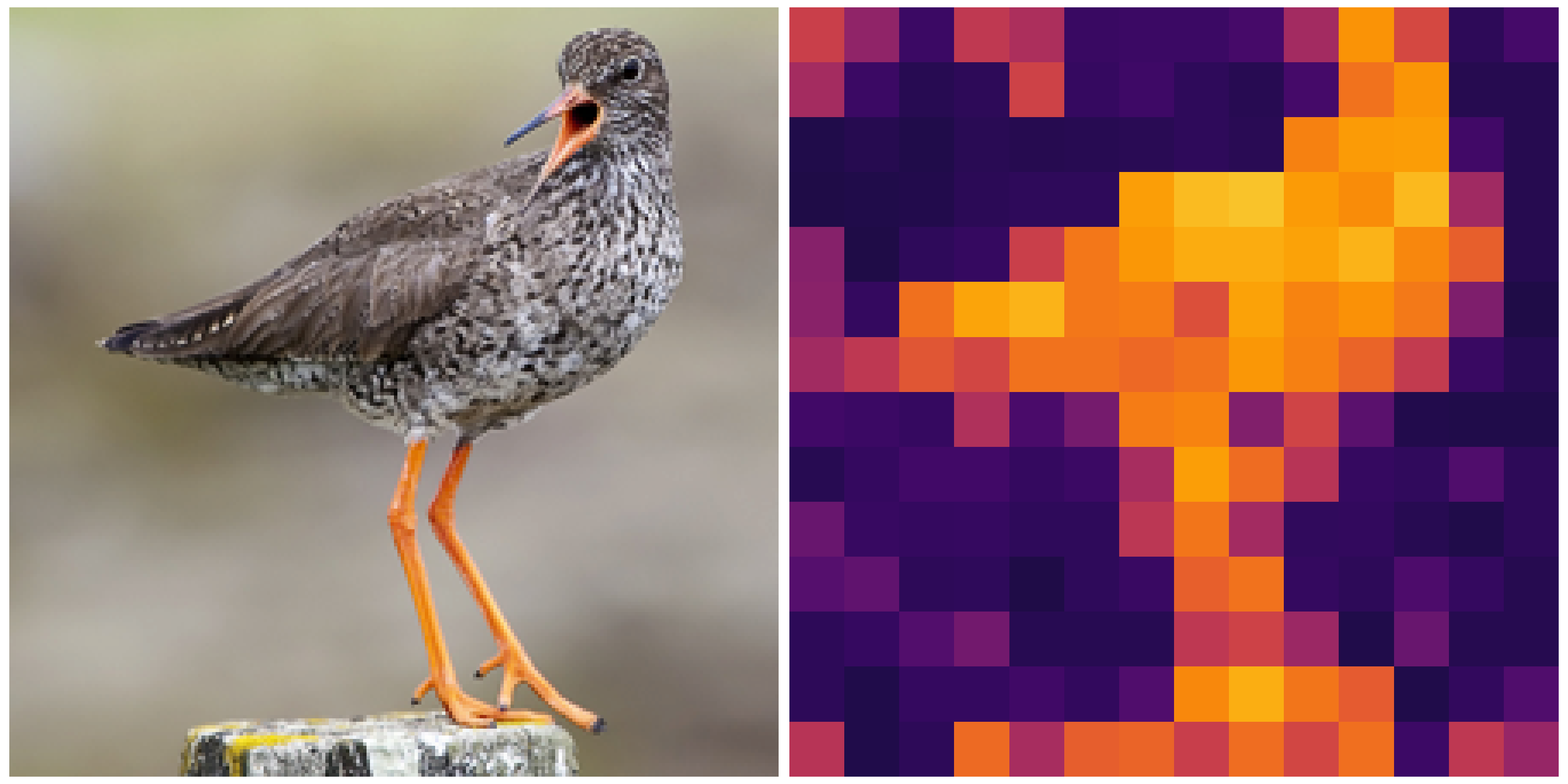}
    \end{minipage}
    \begin{minipage}[c]{\linewidth}
        \centering
        \includegraphics[width=0.243\linewidth]{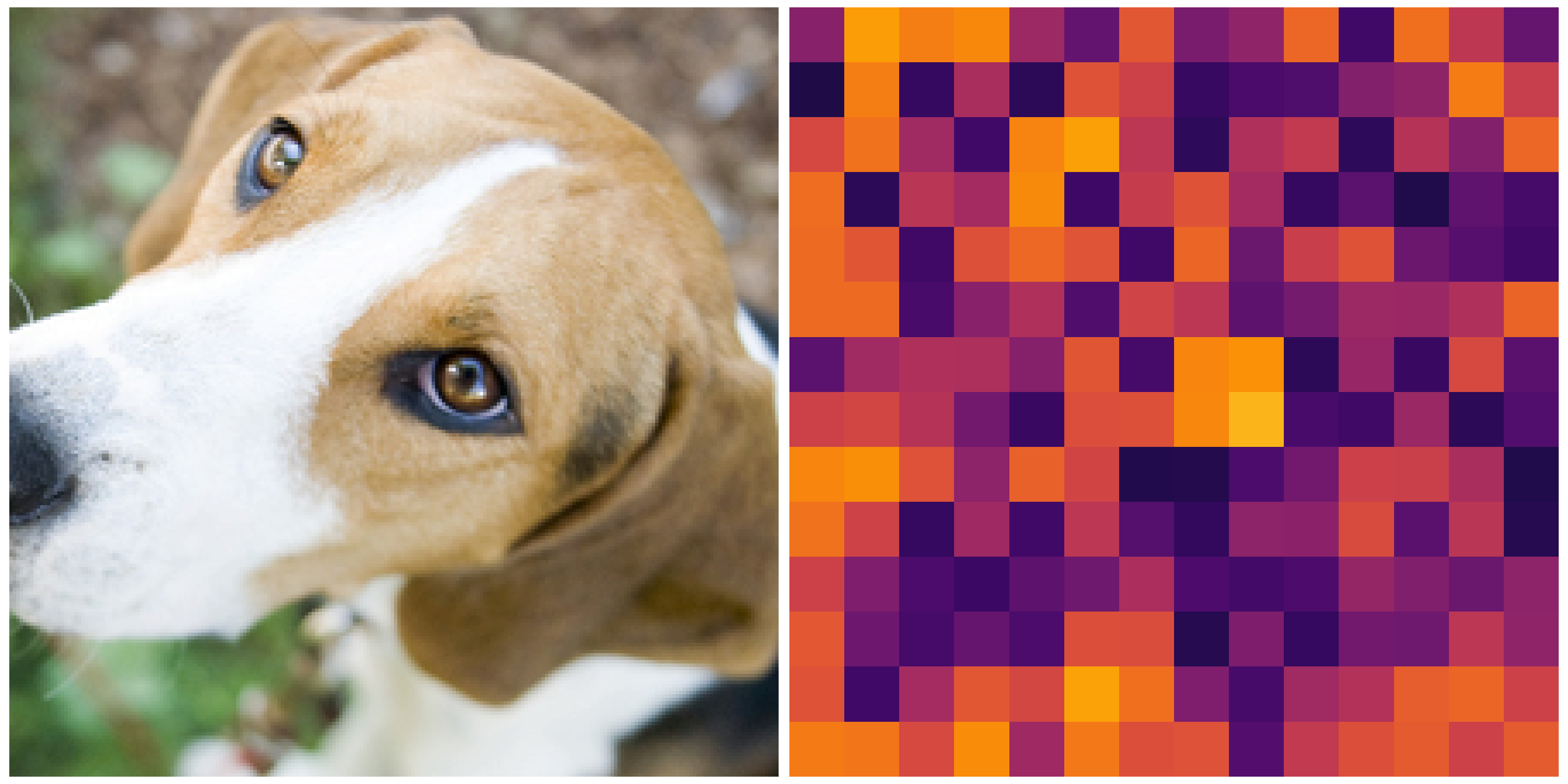}
        \hfill
        \includegraphics[width=0.243\linewidth]{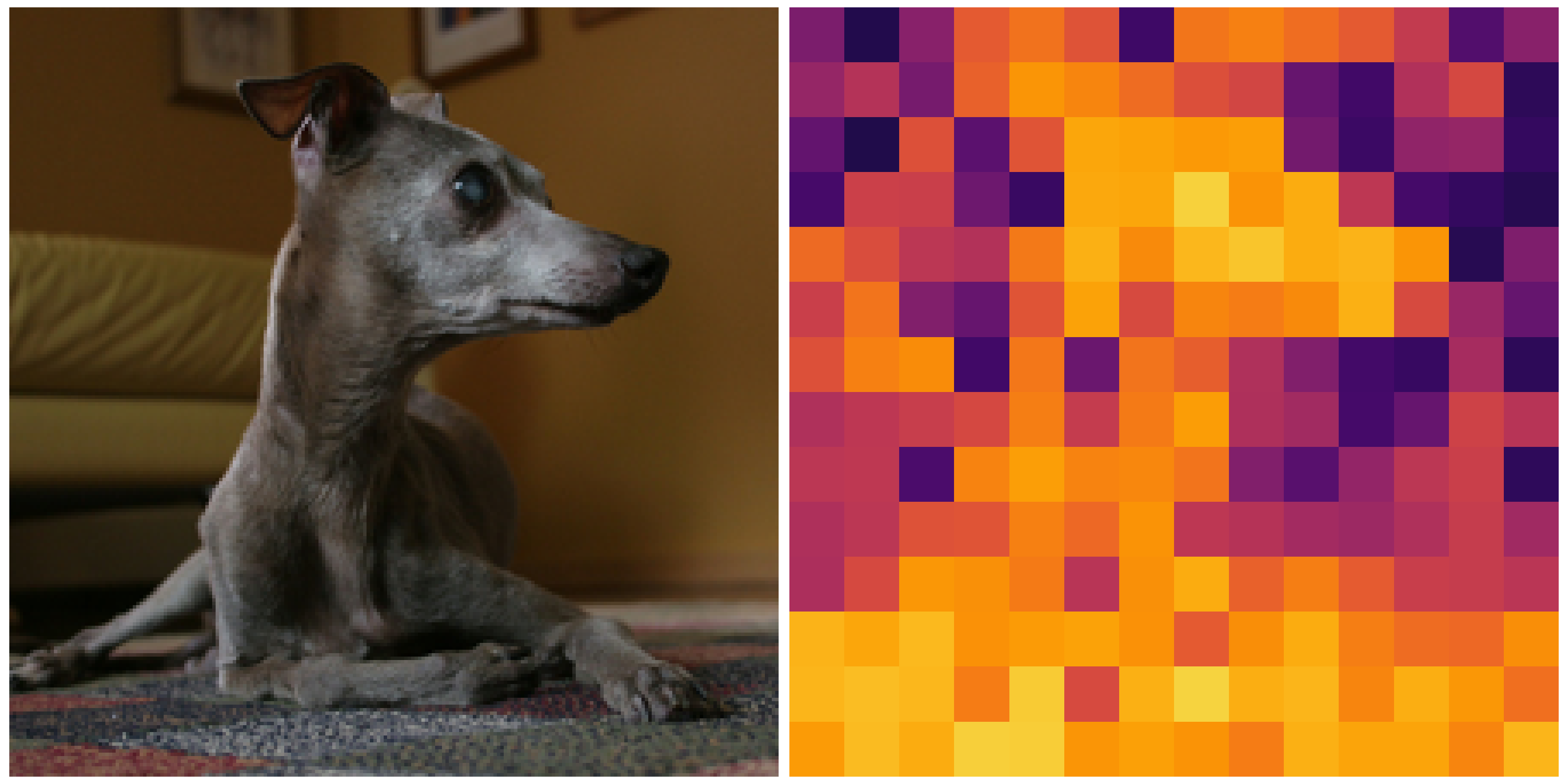}
        \hfill
        \includegraphics[width=0.243\linewidth]{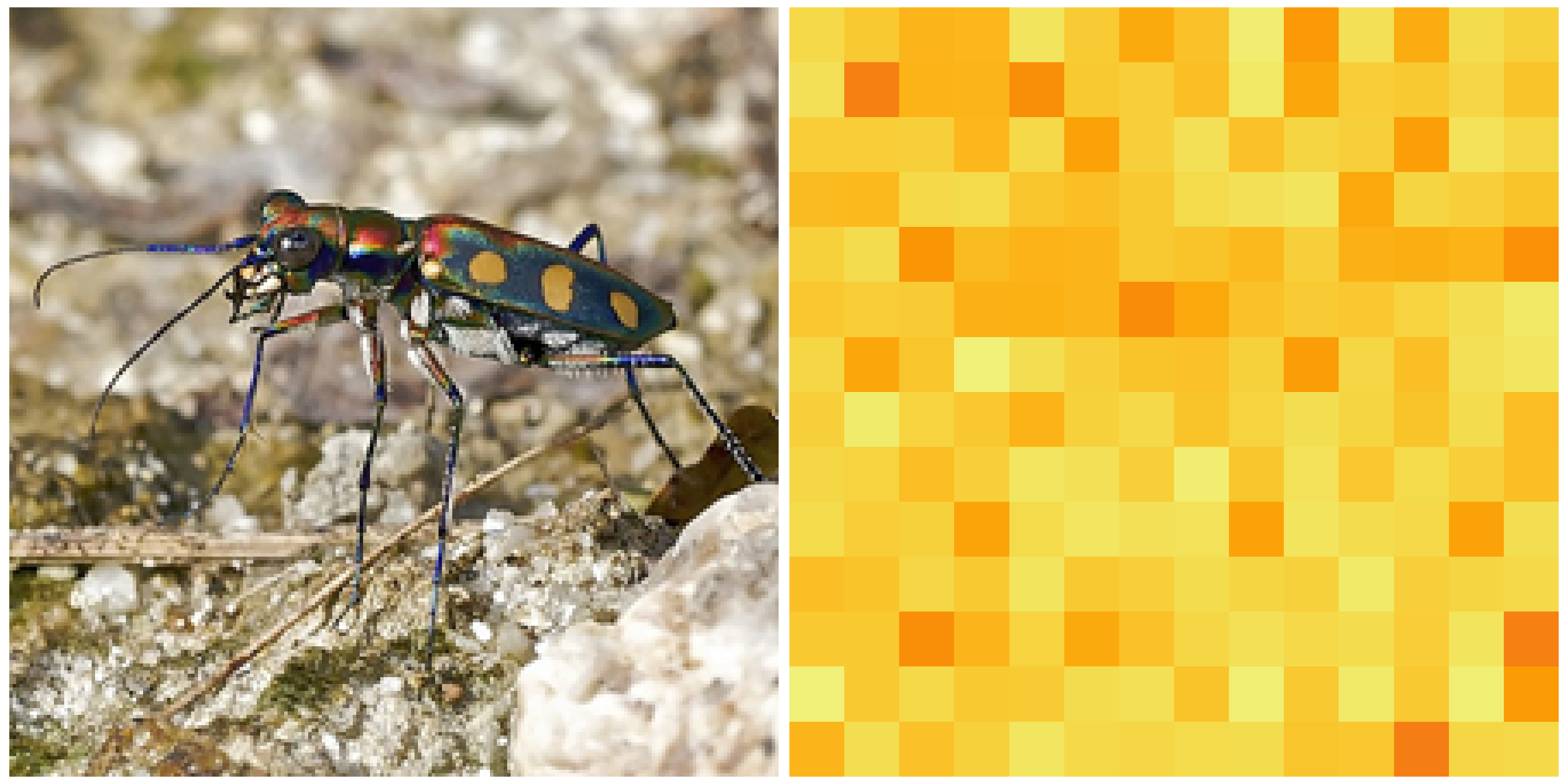}
        \hfill
        \includegraphics[width=0.243\linewidth]{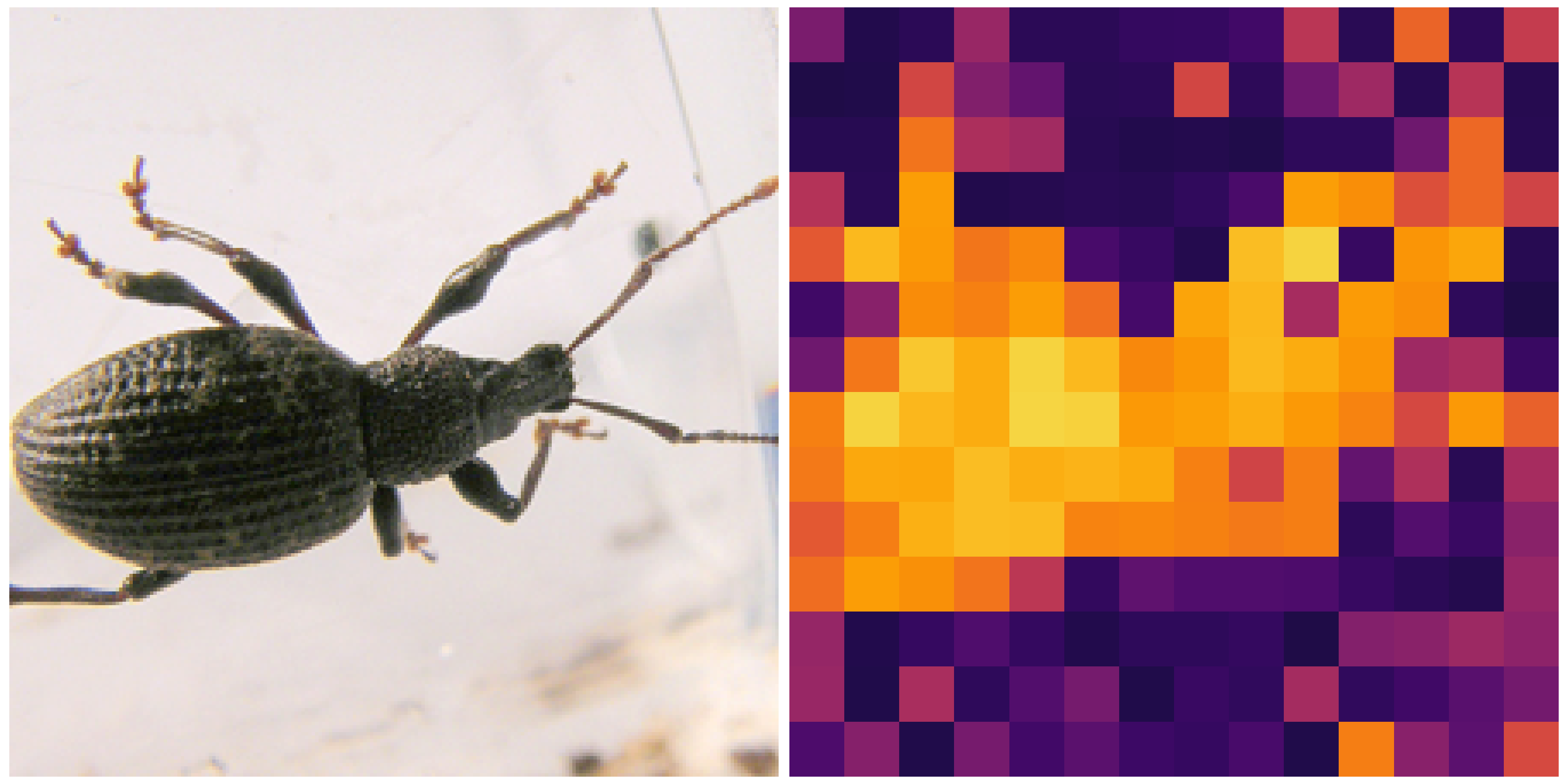}
    \end{minipage}
    \begin{minipage}[c]{\linewidth}
        \centering
        \includegraphics[width=0.243\linewidth]{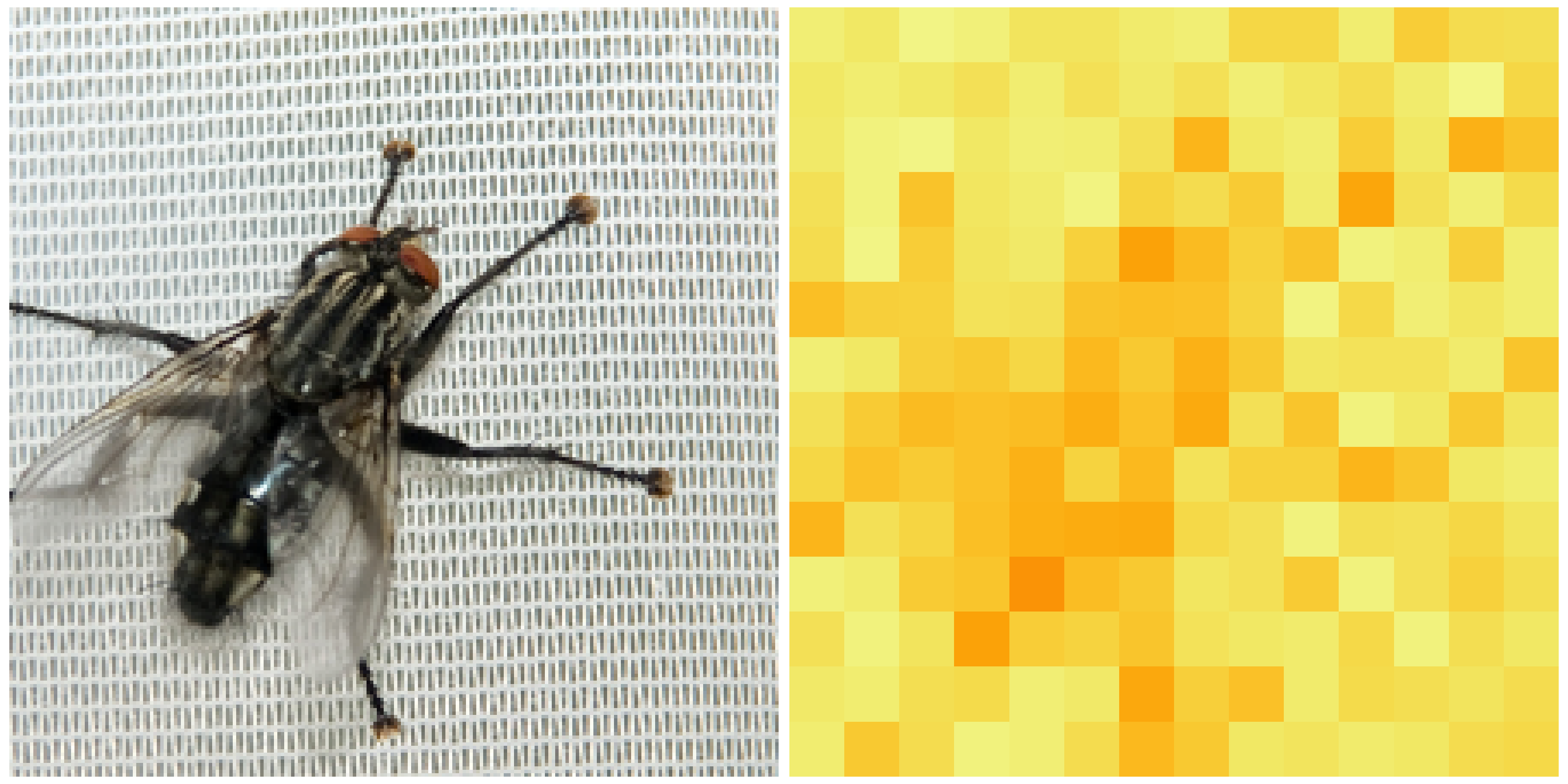}
        \hfill
        \includegraphics[width=0.243\linewidth]{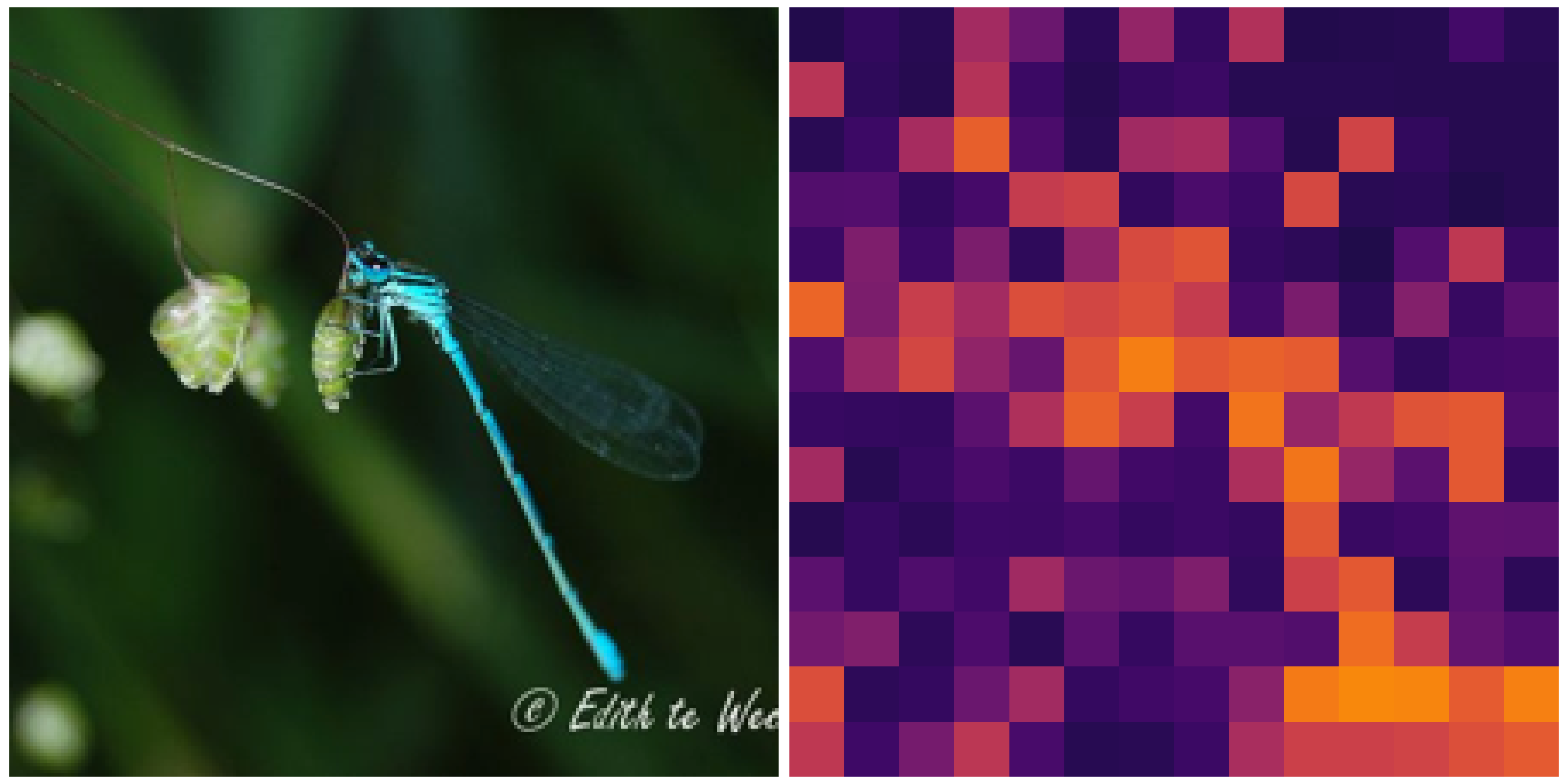}
        \hfill
        \includegraphics[width=0.243\linewidth]{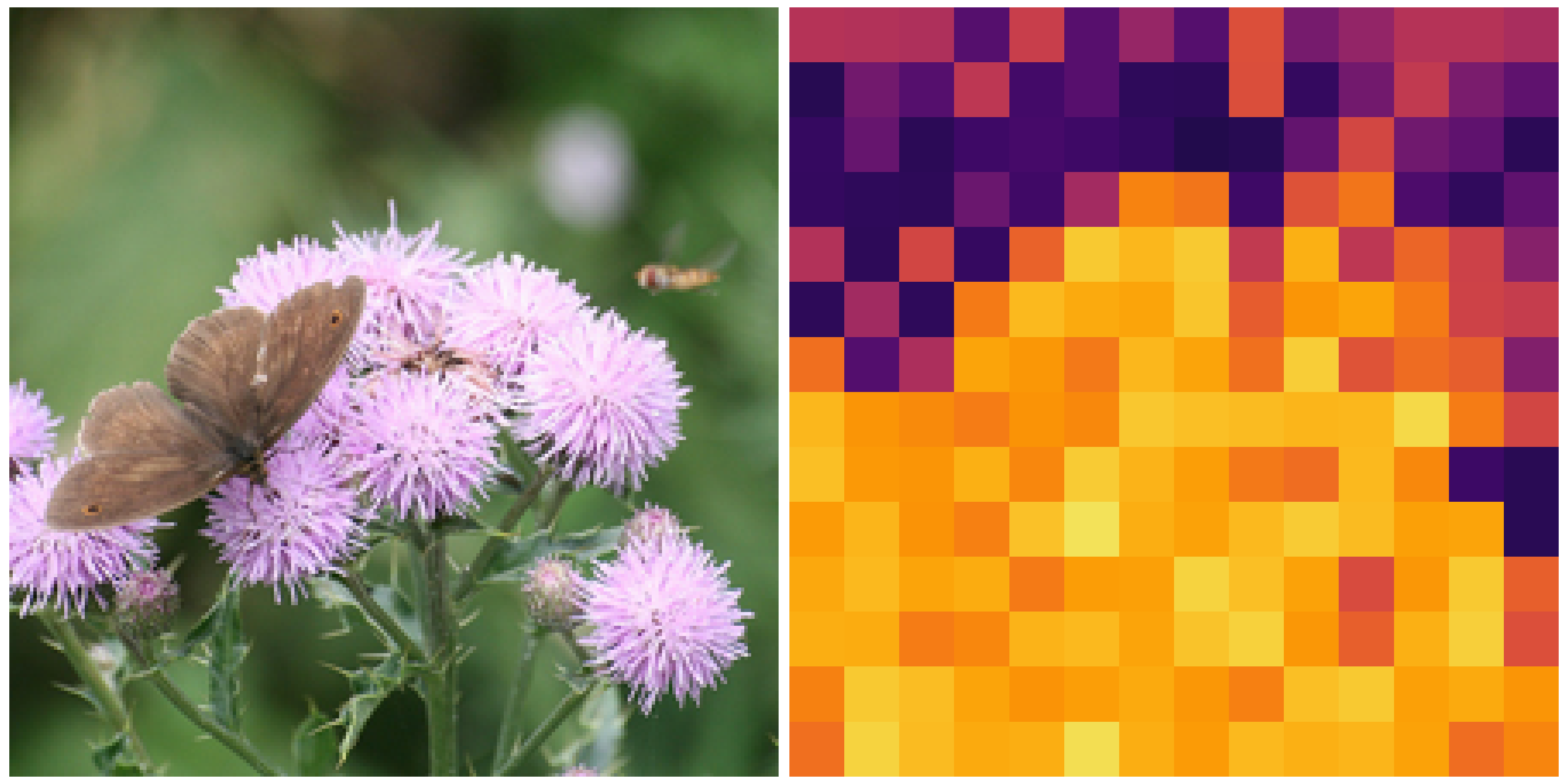}
        \hfill
        \includegraphics[width=0.243\linewidth]{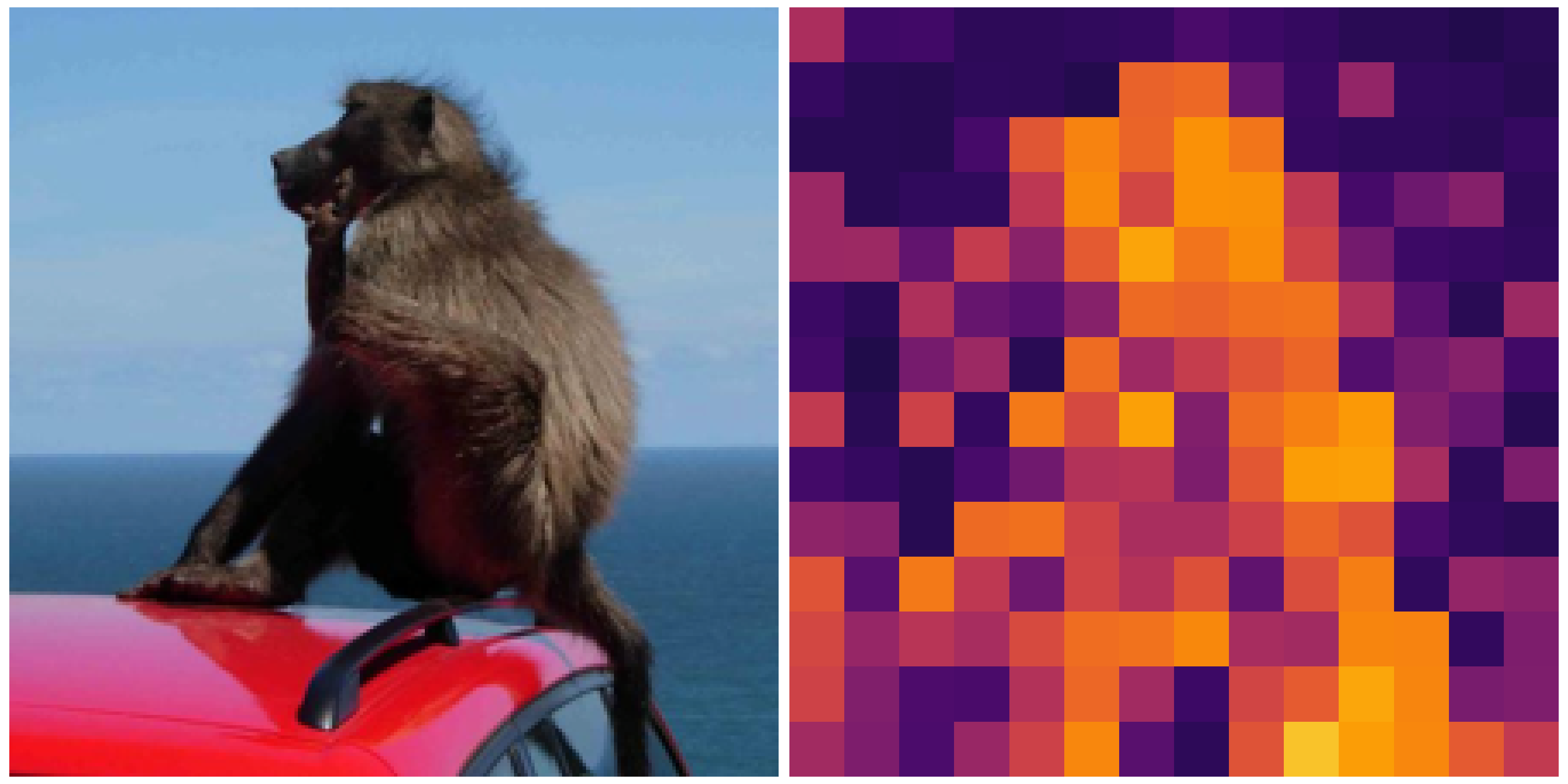}
    \end{minipage}
    \begin{minipage}[c]{\linewidth}
        \centering
        \includegraphics[width=0.243\linewidth]{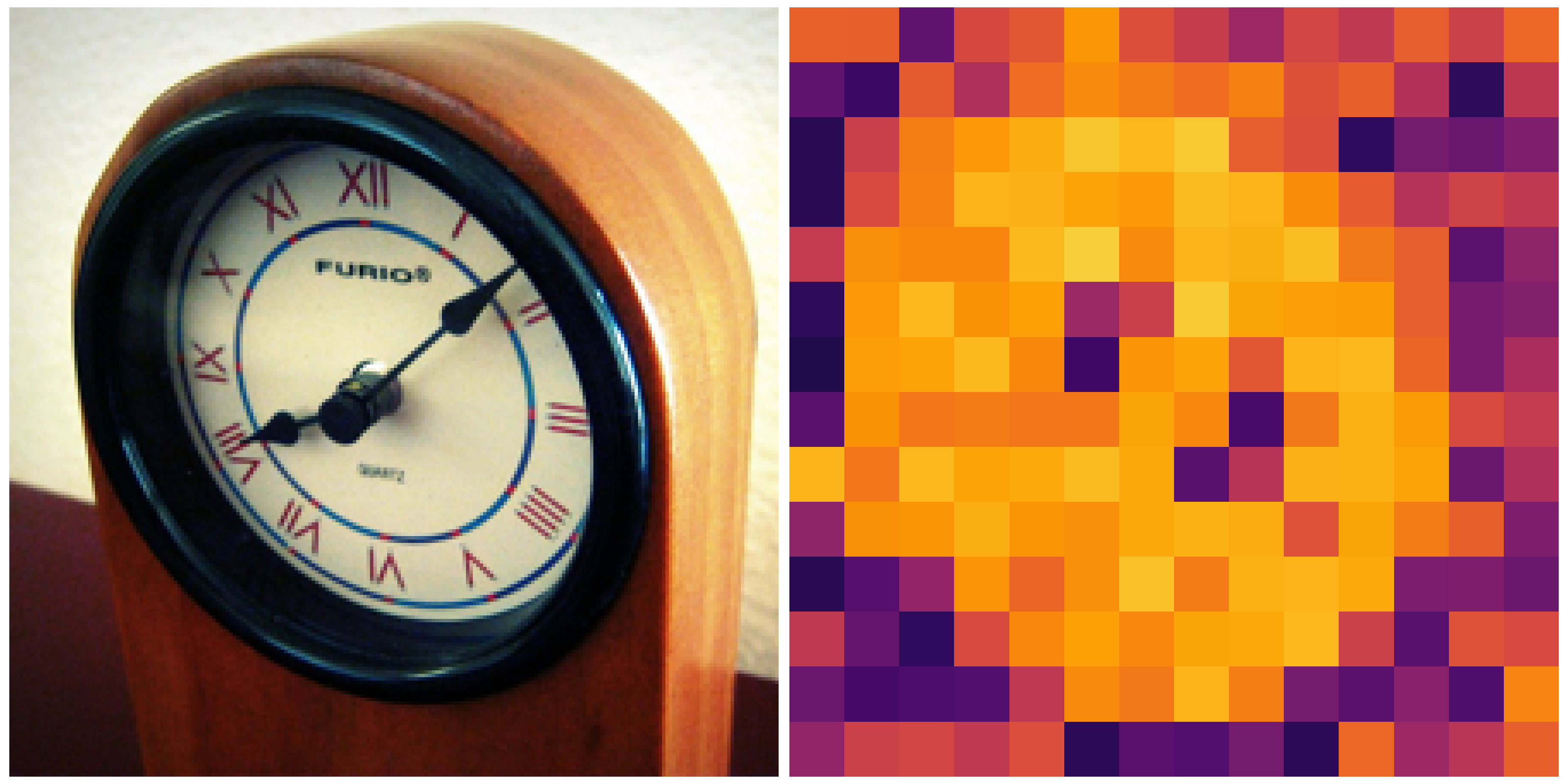}
        \hfill
        \includegraphics[width=0.243\linewidth]{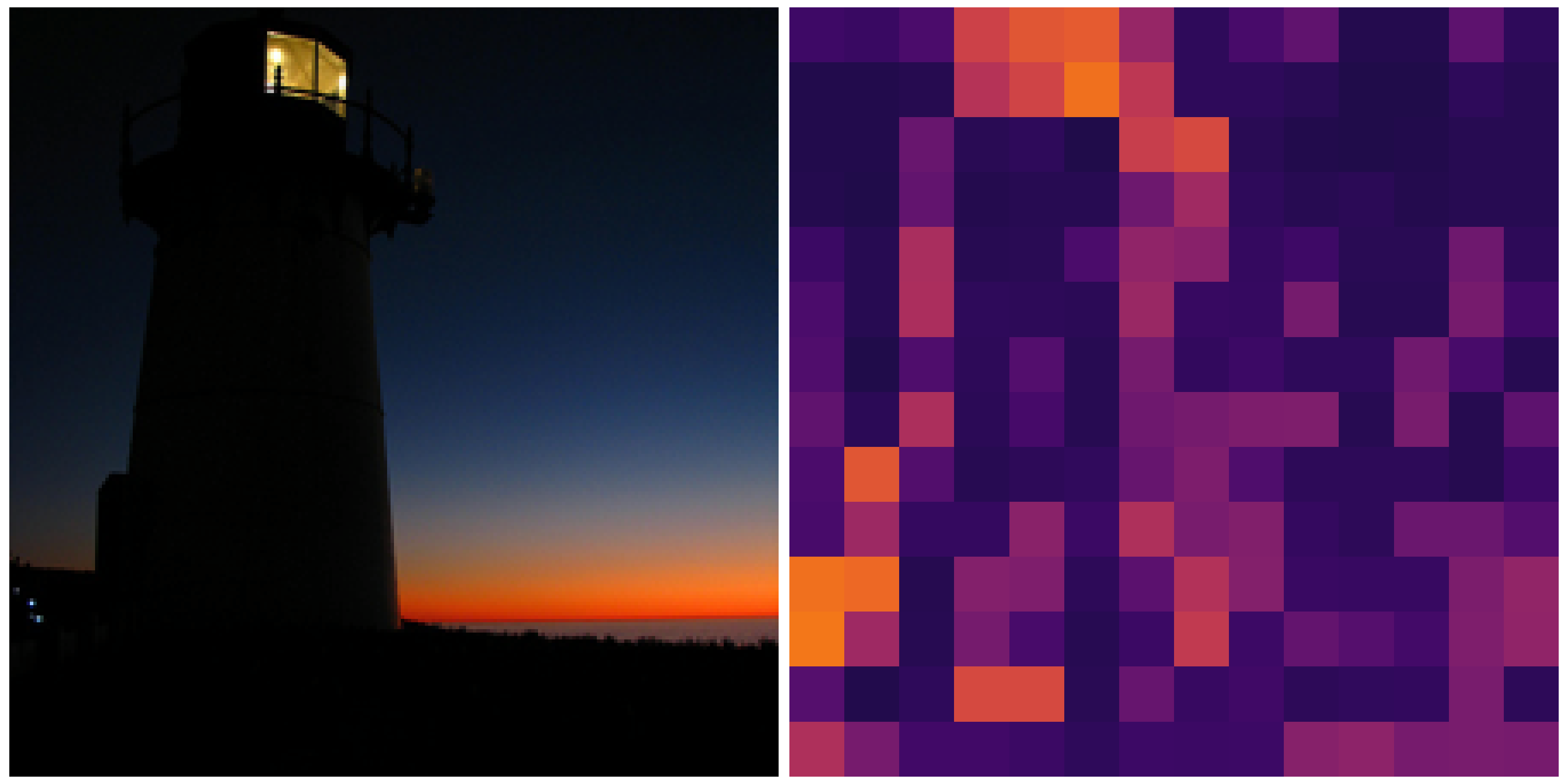}
        \hfill
        \includegraphics[width=0.243\linewidth]{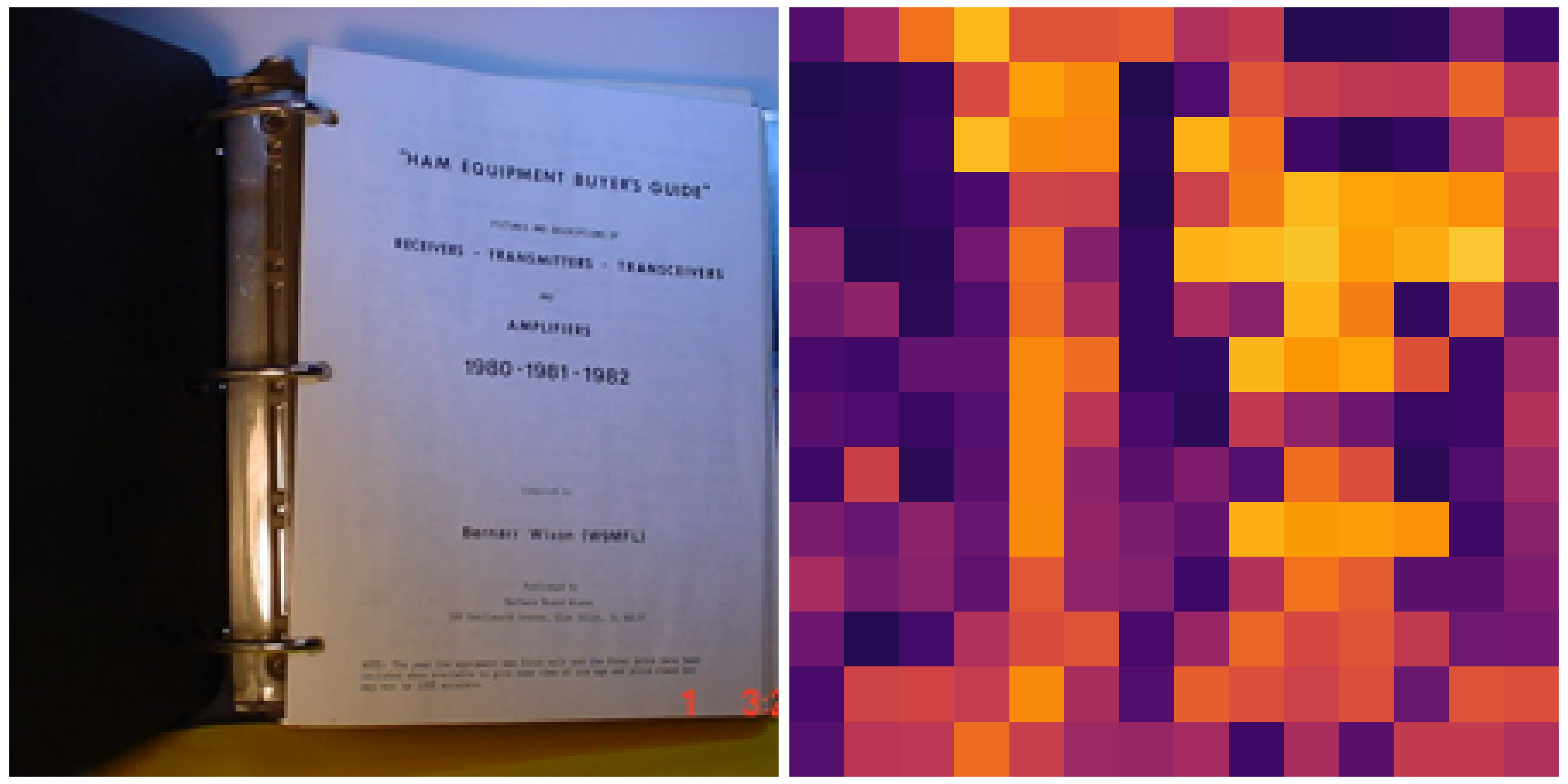}
        \hfill
        \includegraphics[width=0.243\linewidth]{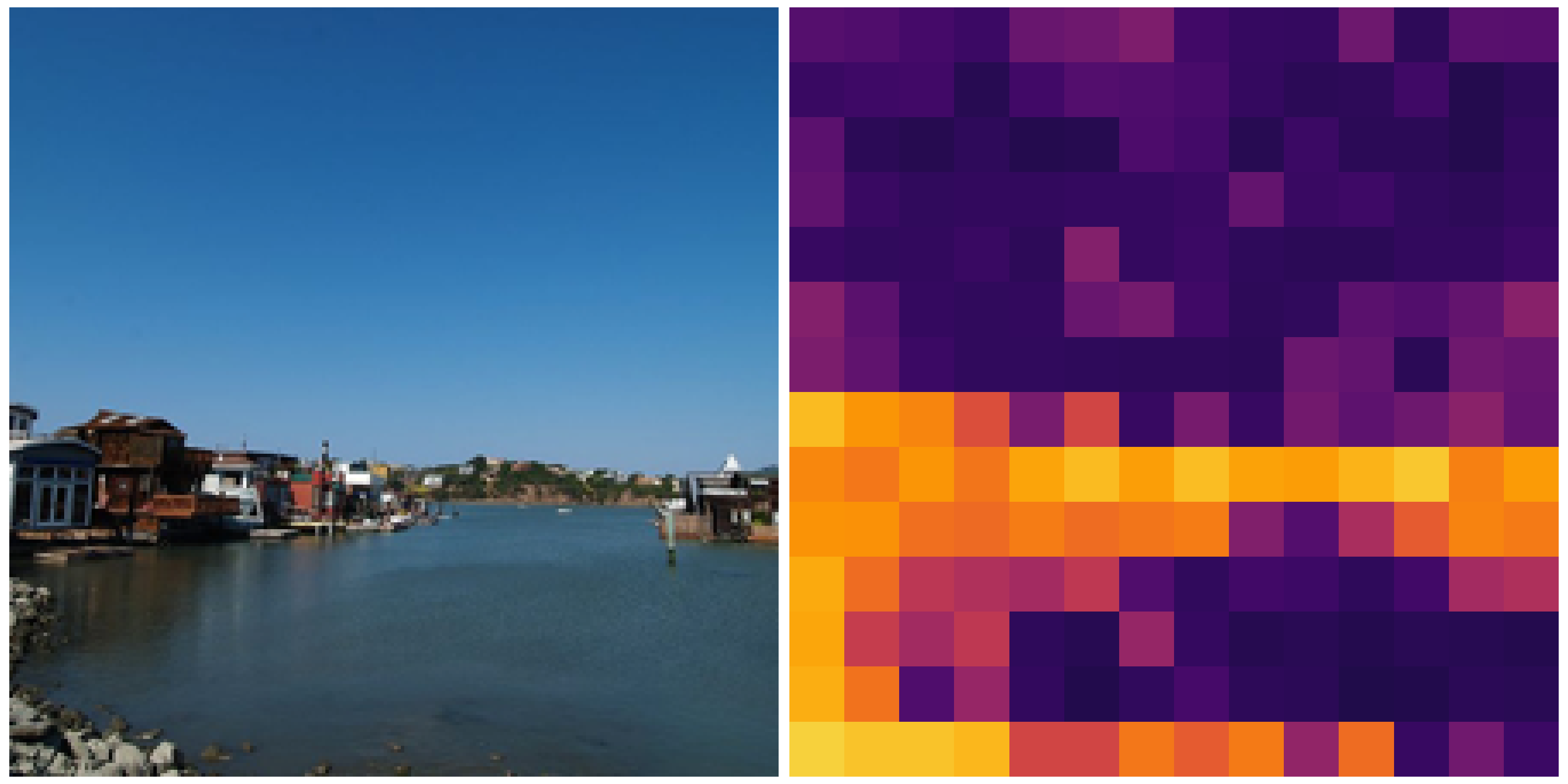}
    \end{minipage}
    \begin{minipage}[c]{\linewidth}
        \centering
        \includegraphics[width=0.243\linewidth]{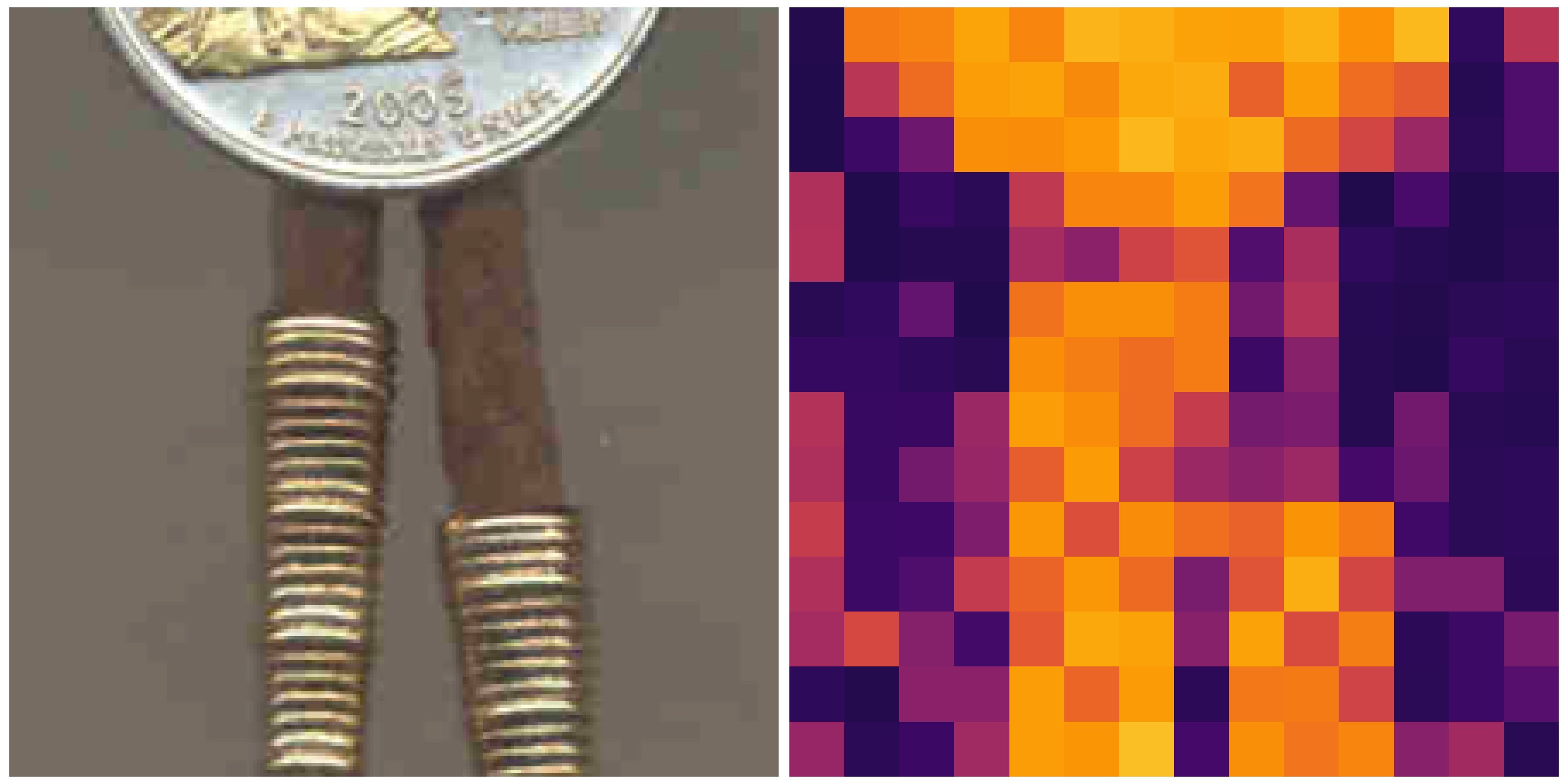}
        \hfill
        \includegraphics[width=0.243\linewidth]{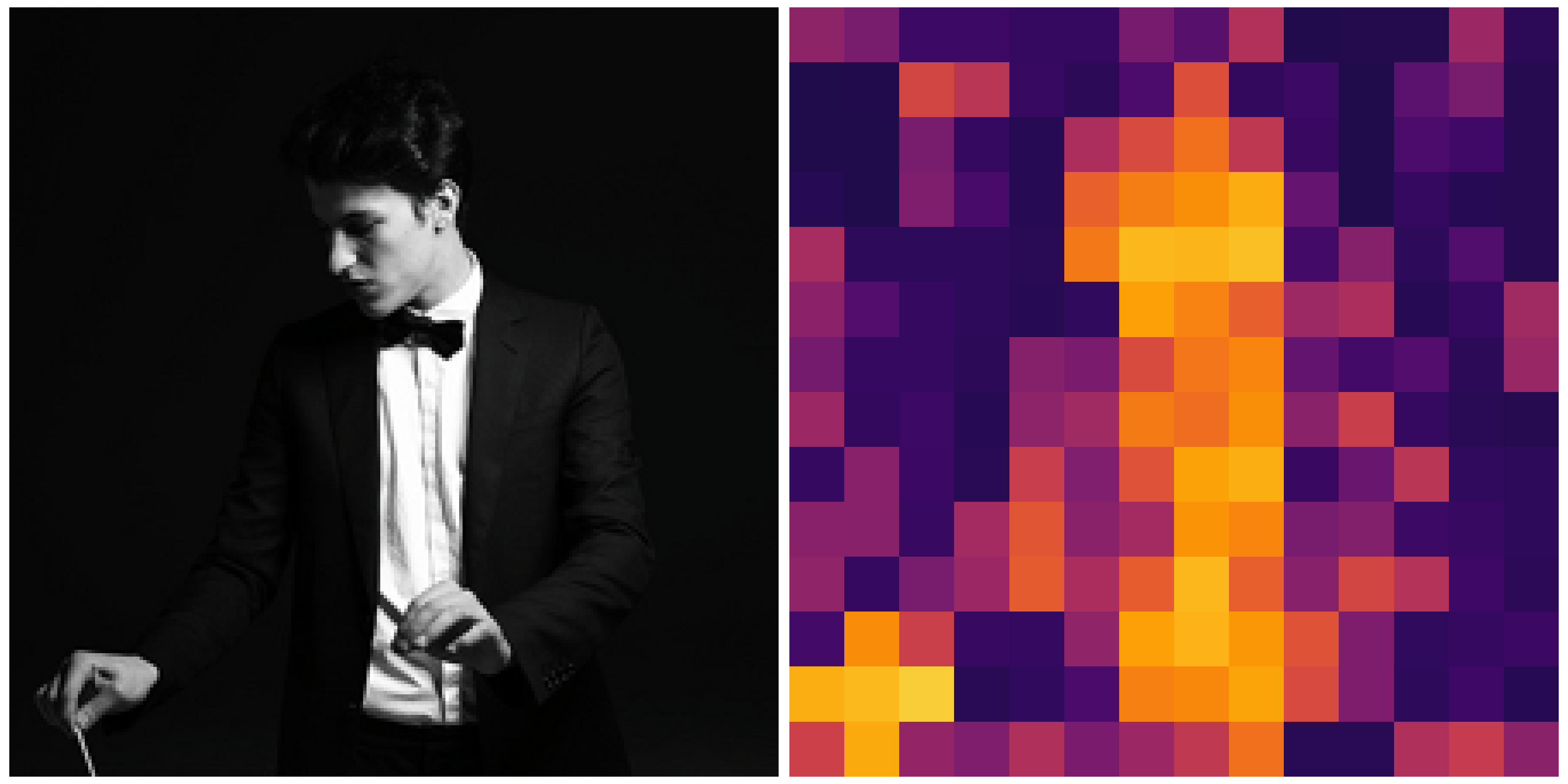}
        \hfill
        \includegraphics[width=0.243\linewidth]{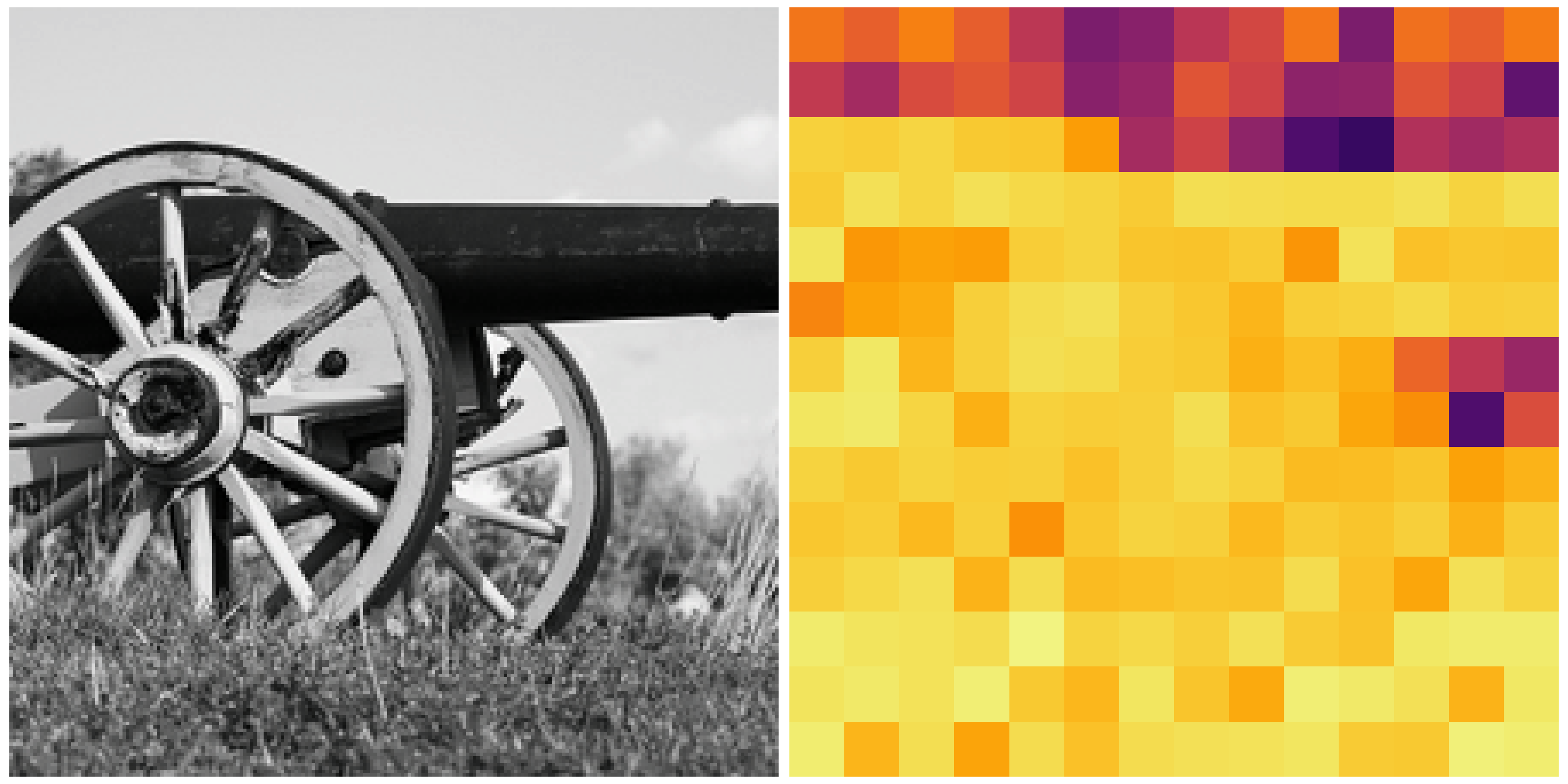}
        \hfill
        \includegraphics[width=0.243\linewidth]{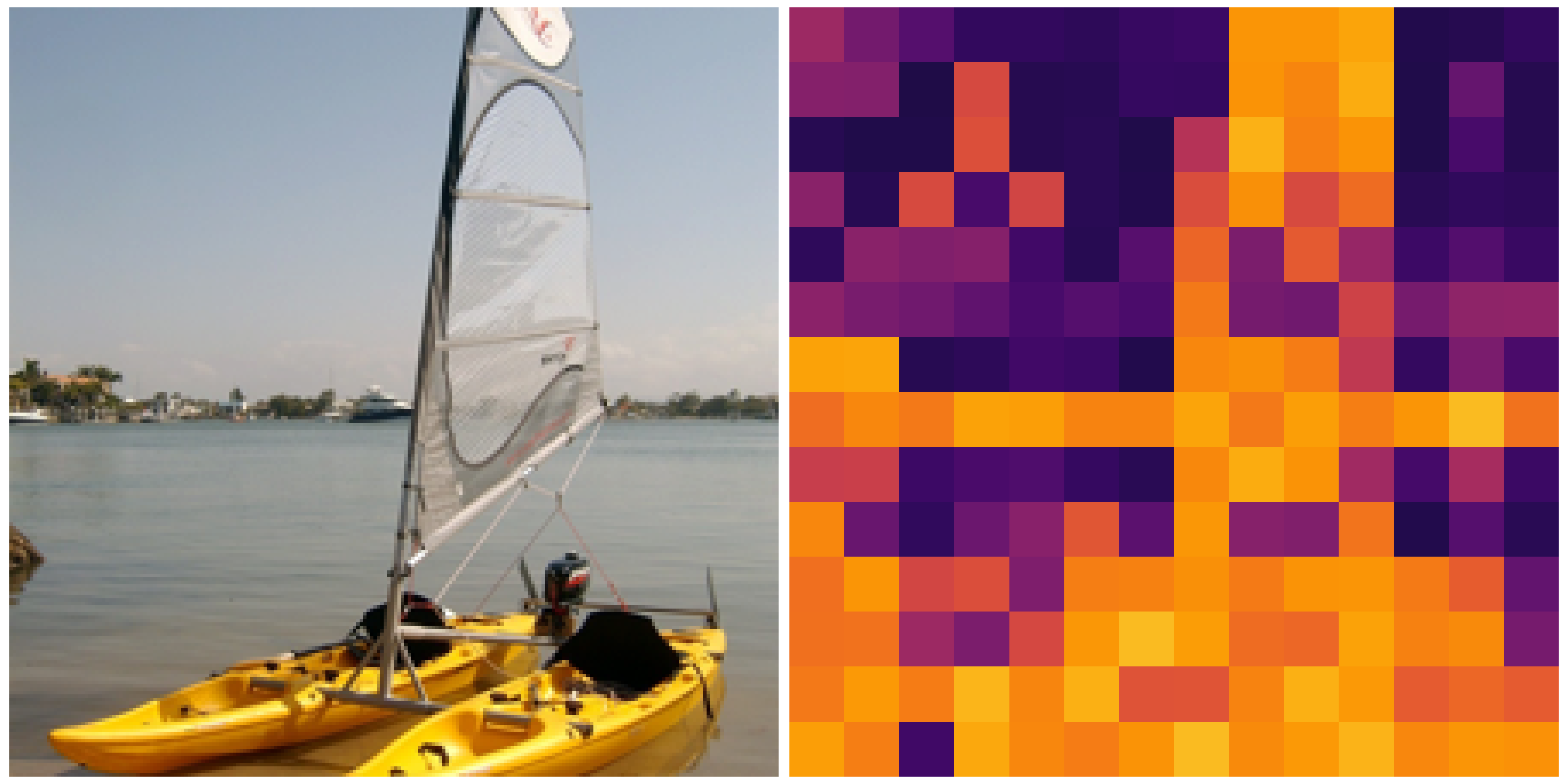}
    \end{minipage}
    \begin{minipage}[c]{\linewidth}
        \centering
        \includegraphics[width=0.243\linewidth]{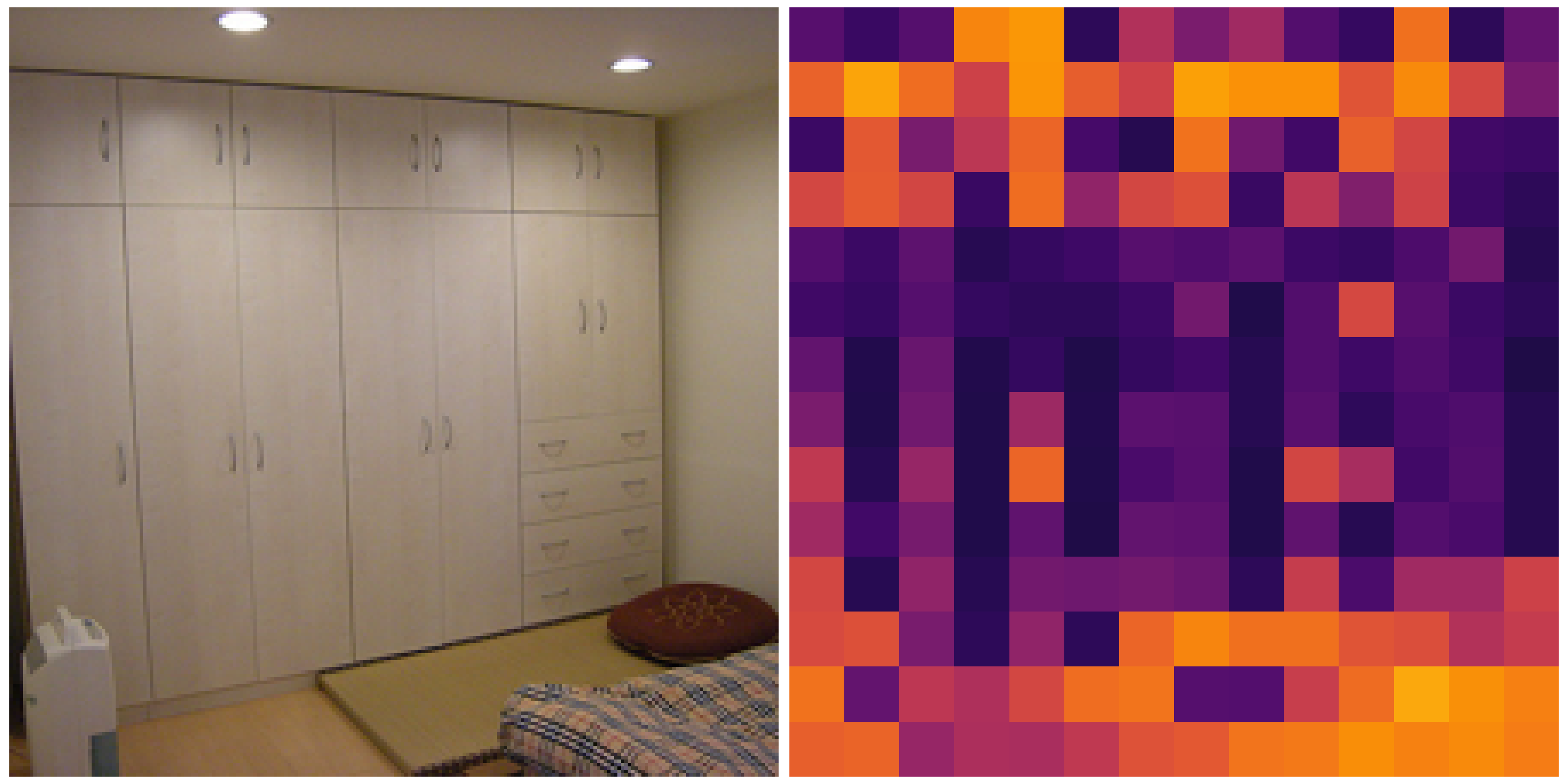}
        \hfill
        \includegraphics[width=0.243\linewidth]{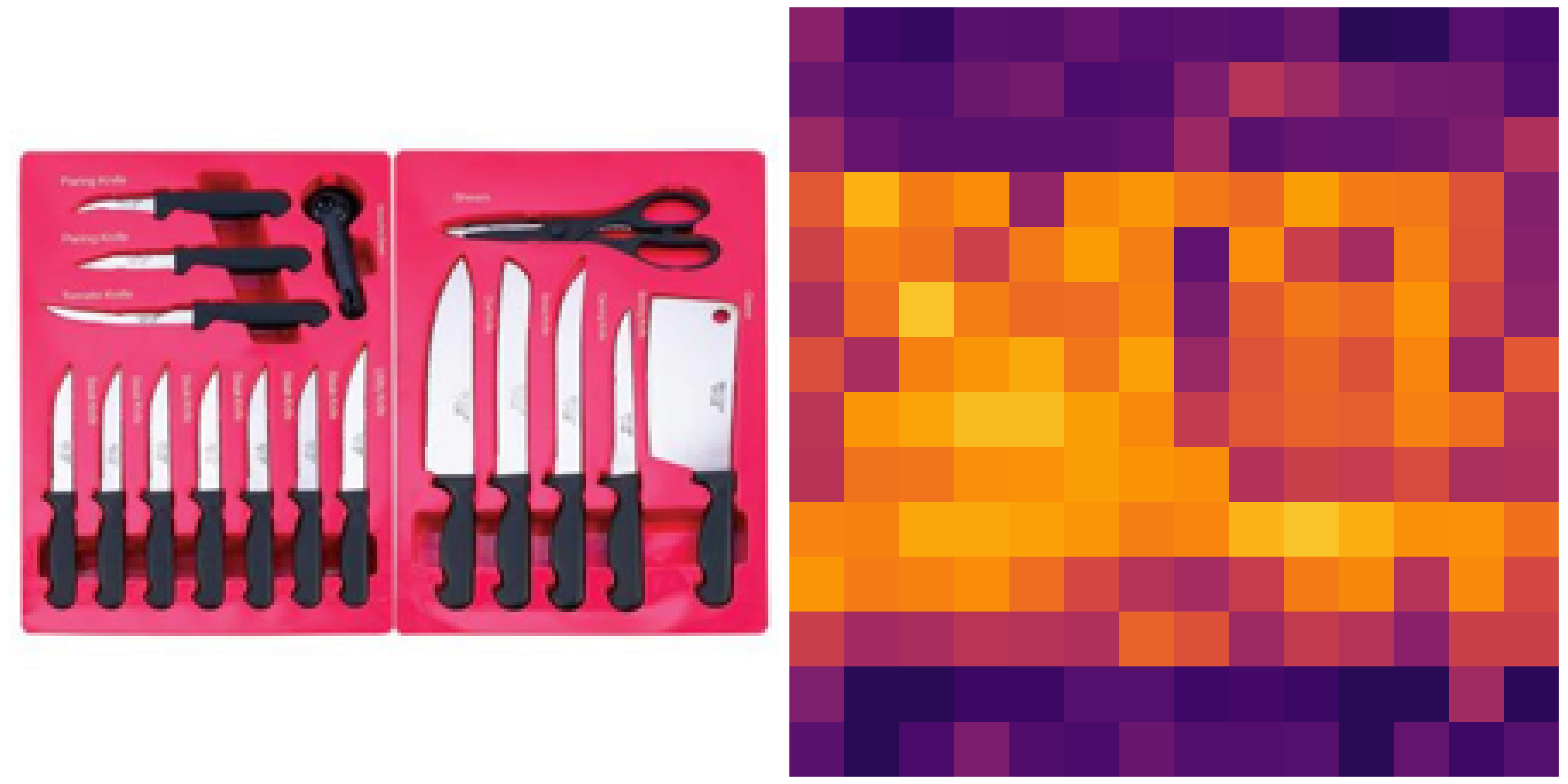}
        \hfill
        \includegraphics[width=0.243\linewidth]{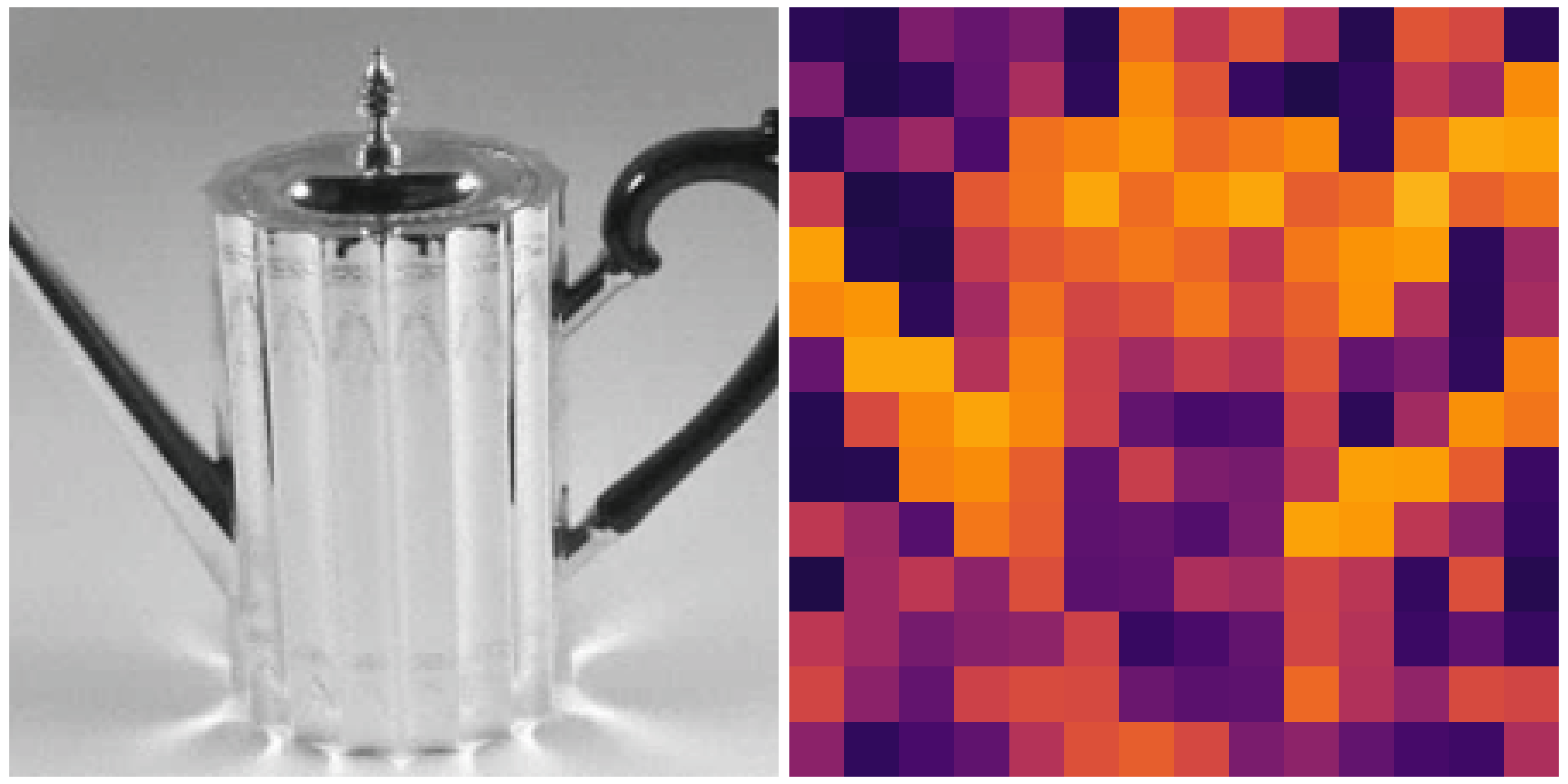}
        \hfill
        \includegraphics[width=0.243\linewidth]{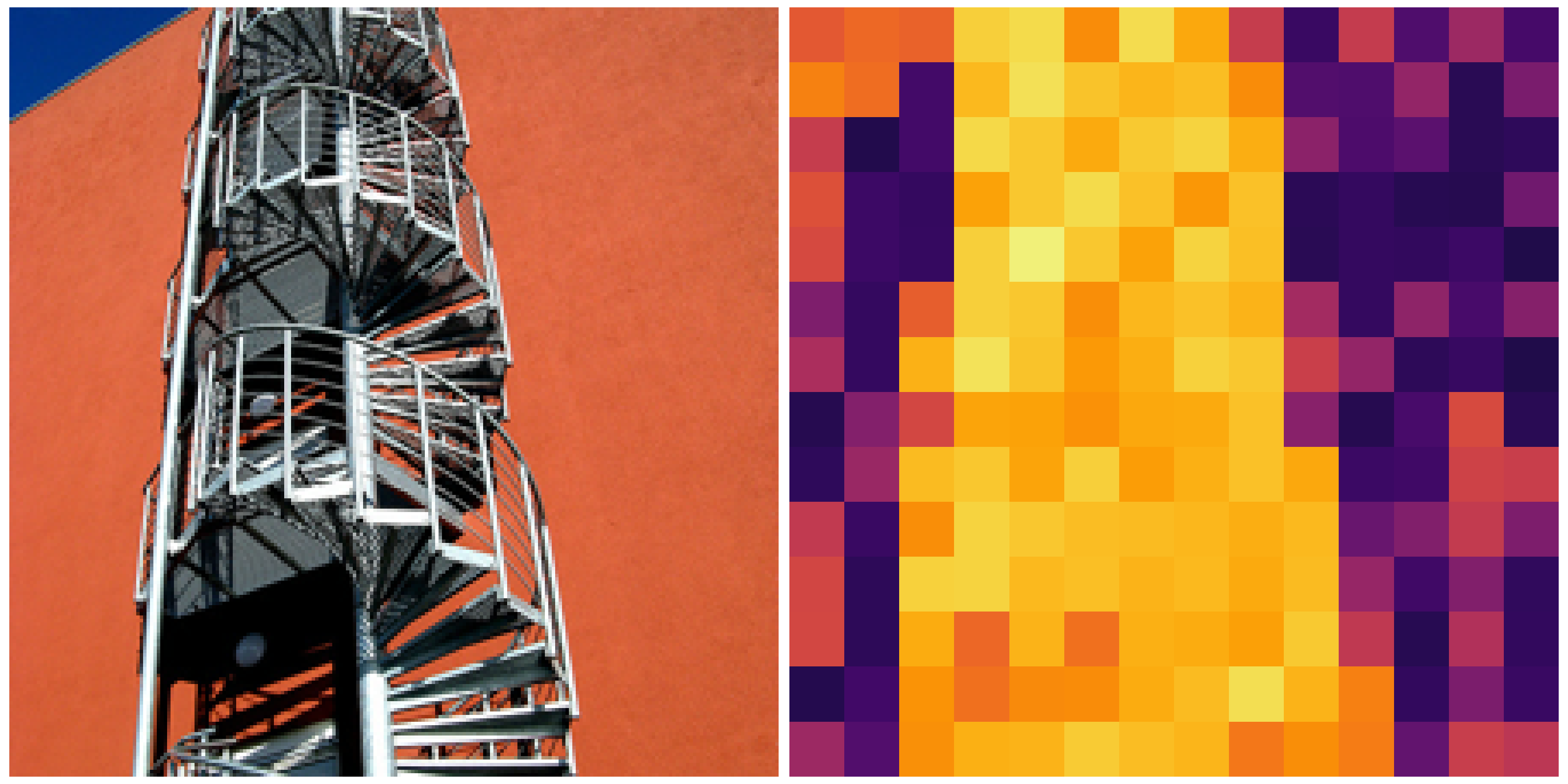}
    \end{minipage}
    \begin{minipage}[c]{\linewidth}
        \centering
        \includegraphics[width=0.243\linewidth]{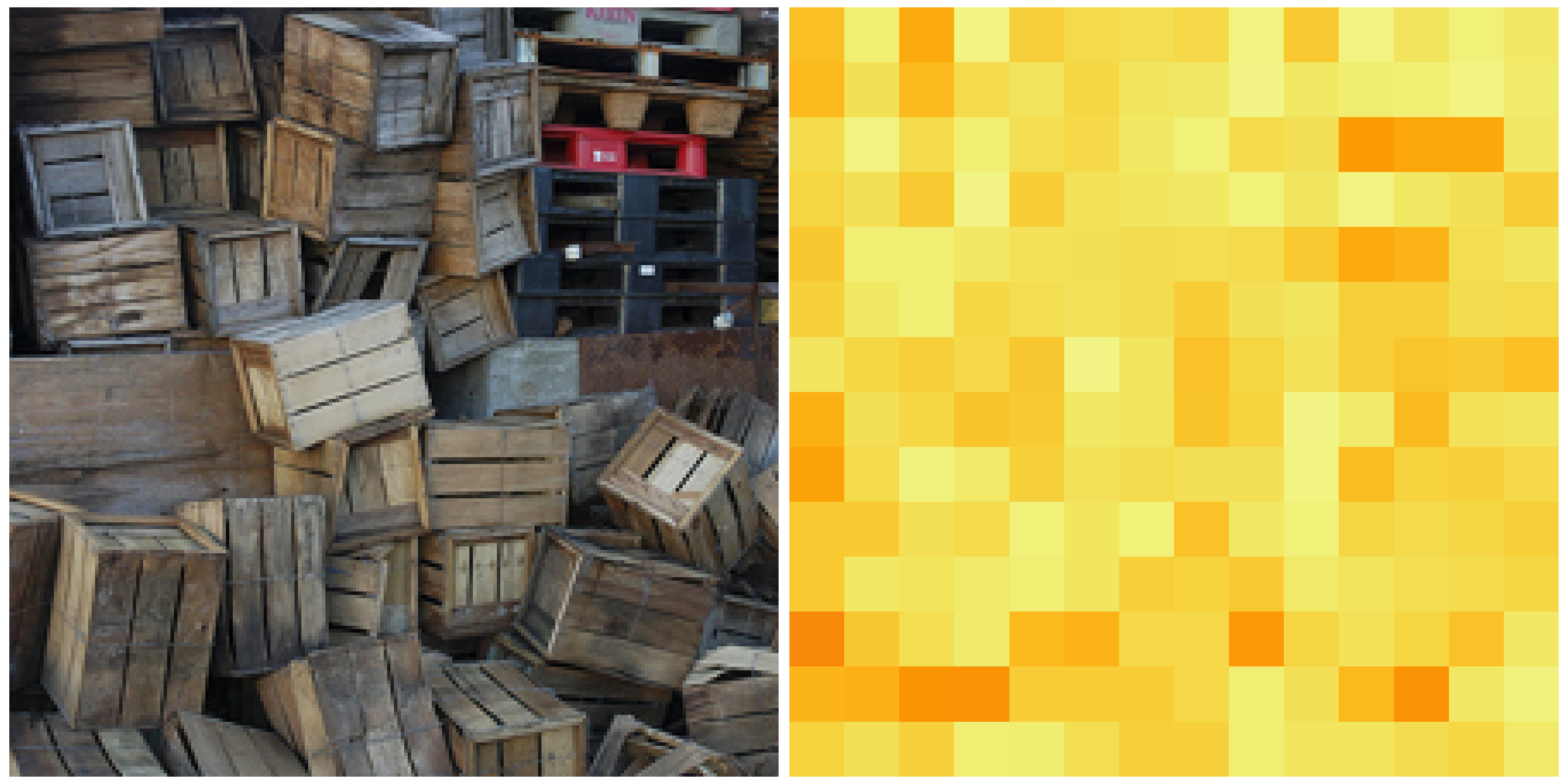}
        \hfill
        \includegraphics[width=0.243\linewidth]{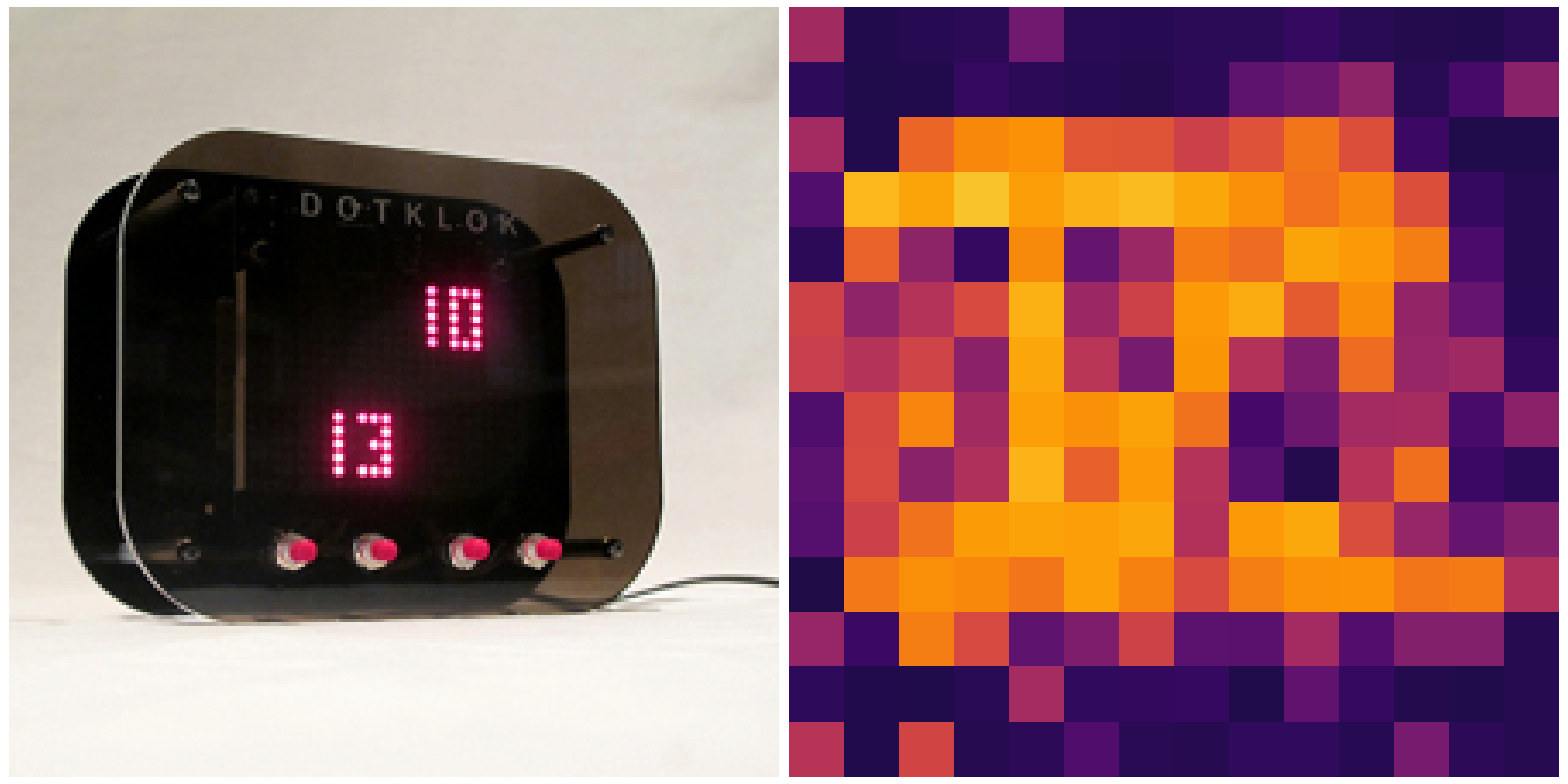}
        \hfill
        \includegraphics[width=0.243\linewidth]{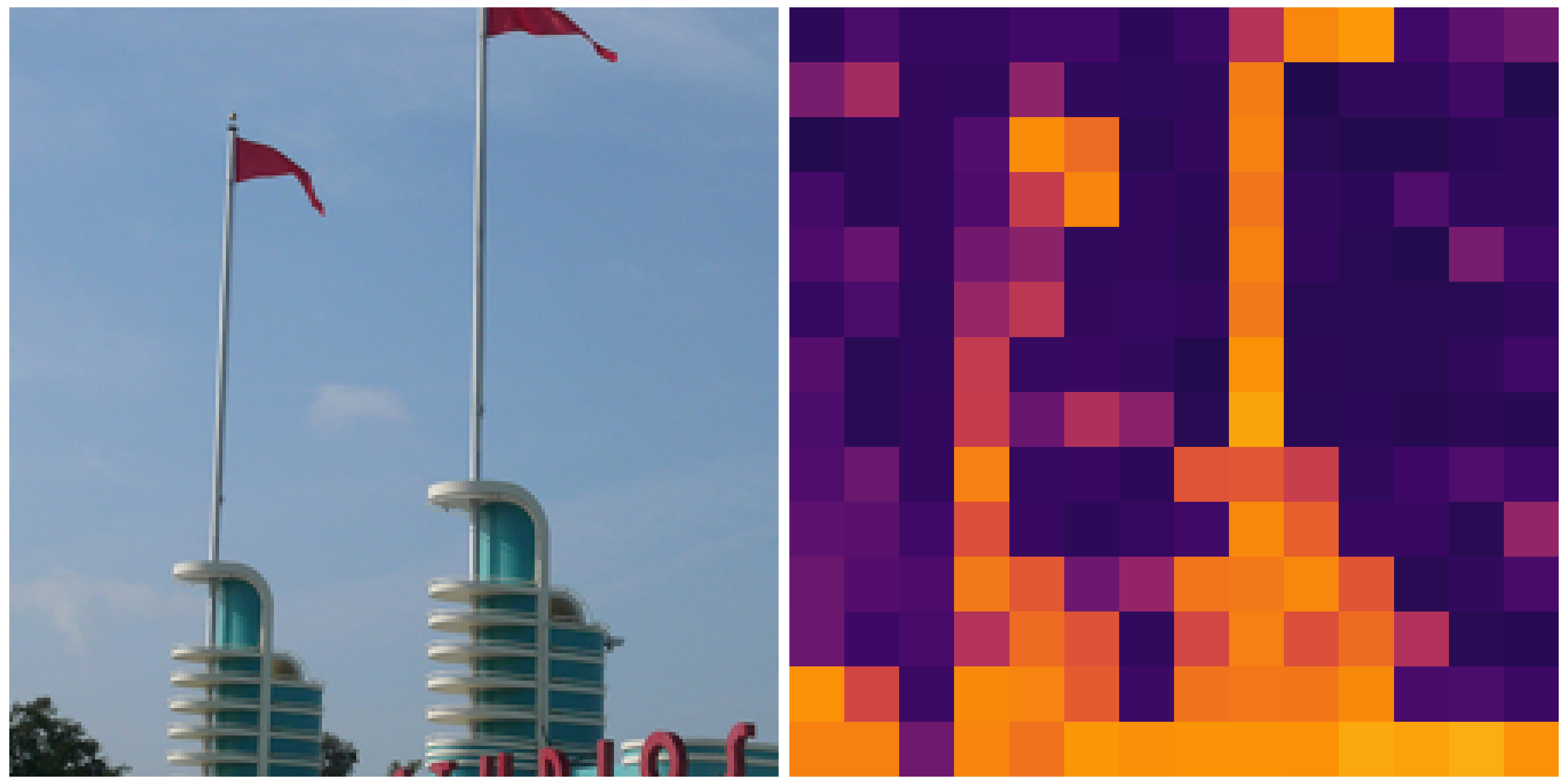}
        \hfill
        \includegraphics[width=0.243\linewidth]{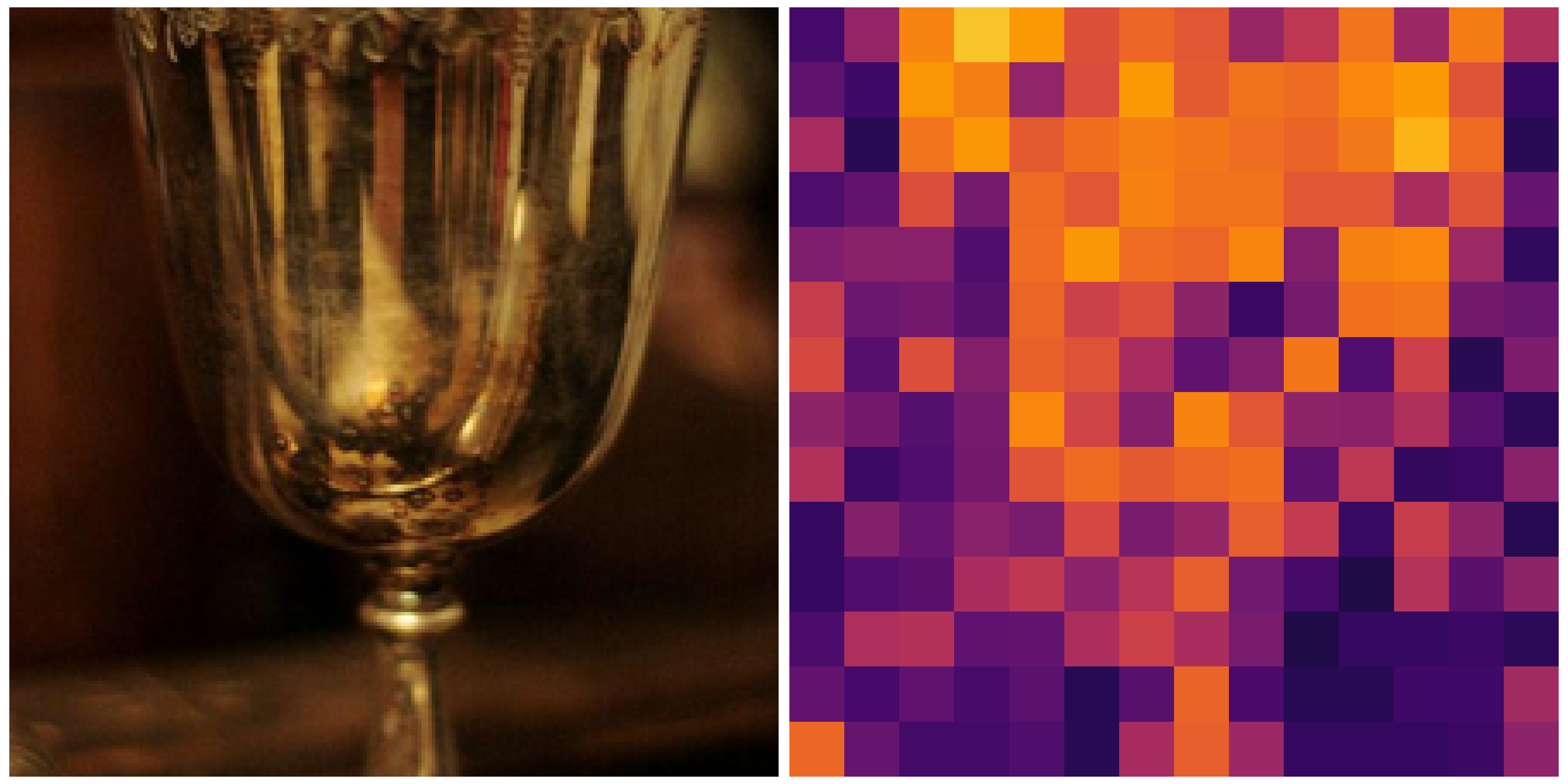}
    \end{minipage}
    \begin{minipage}[c]{\linewidth}
        \centering
        \includegraphics[width=0.243\linewidth]{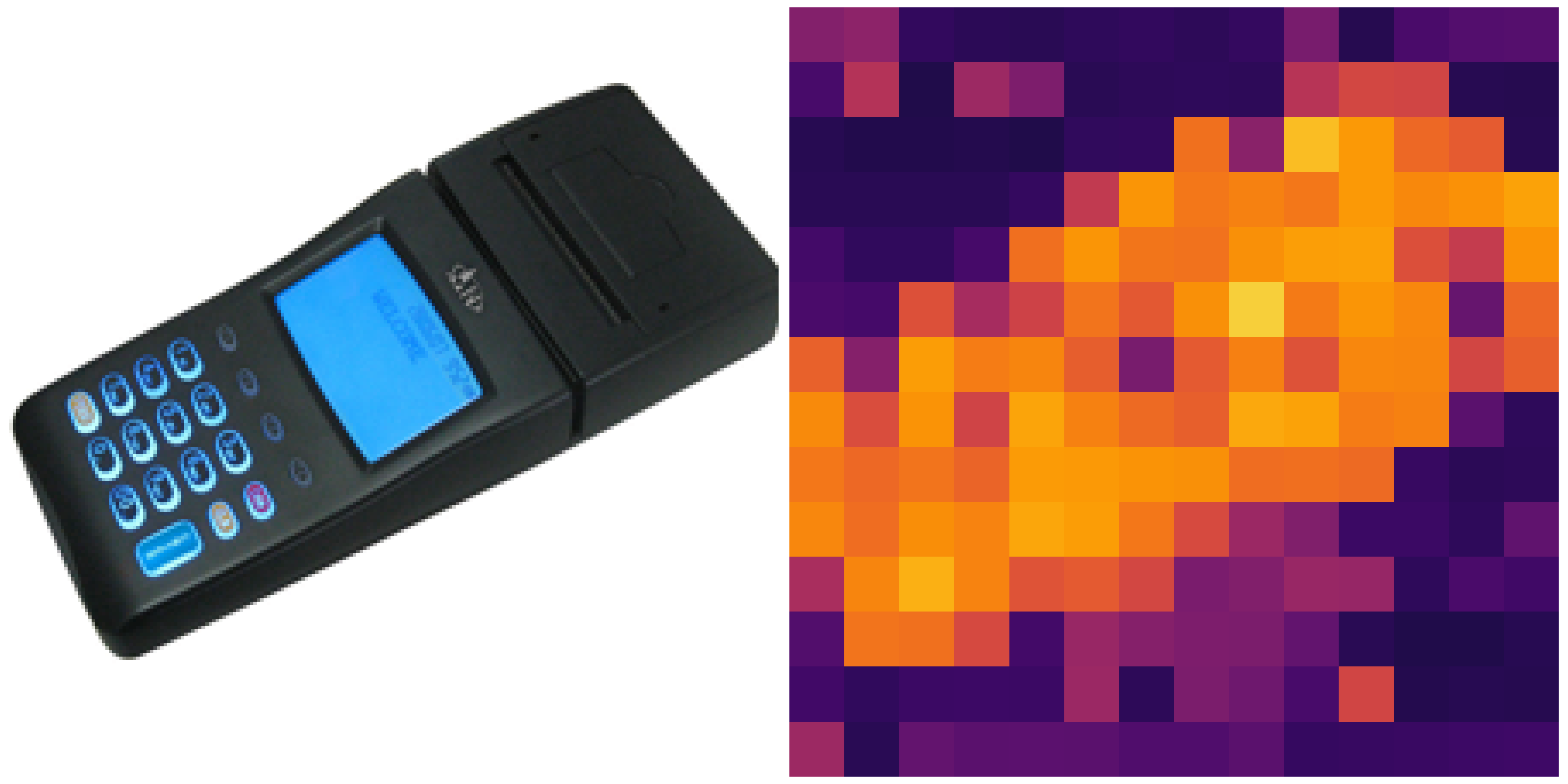}
        \hfill
        \includegraphics[width=0.243\linewidth]{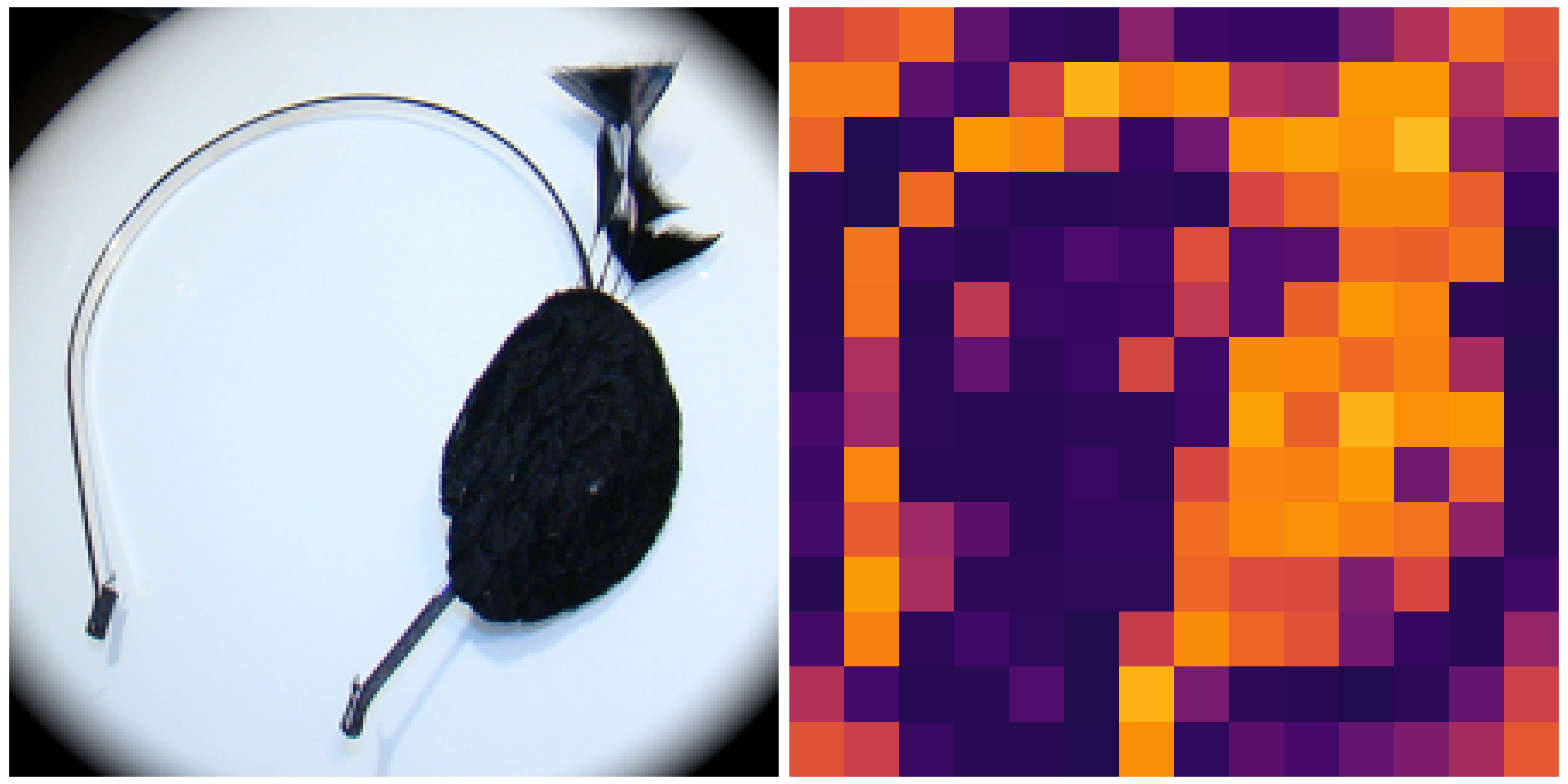}
        \hfill
        \includegraphics[width=0.243\linewidth]{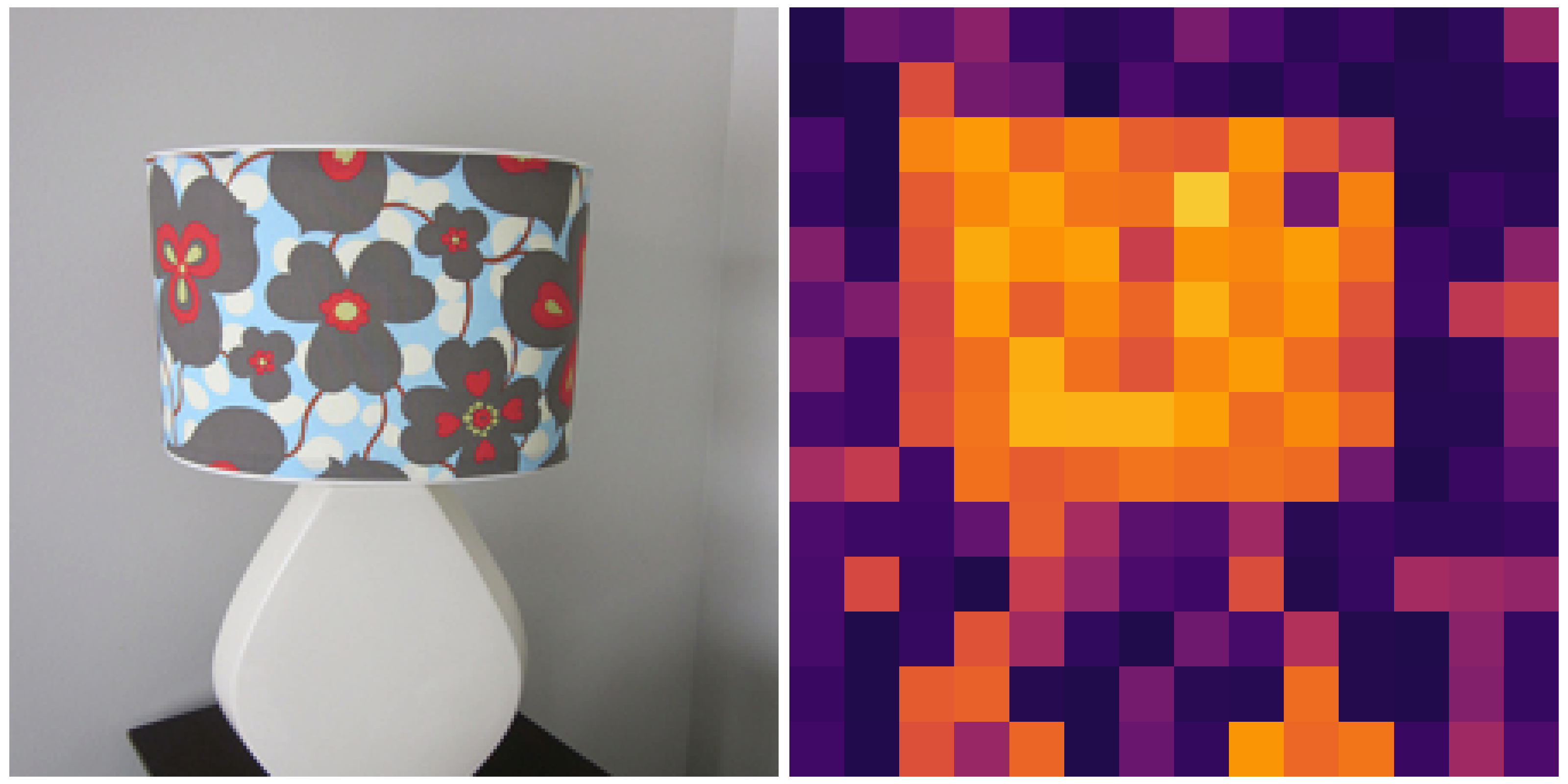}
        \hfill
        \includegraphics[width=0.243\linewidth]{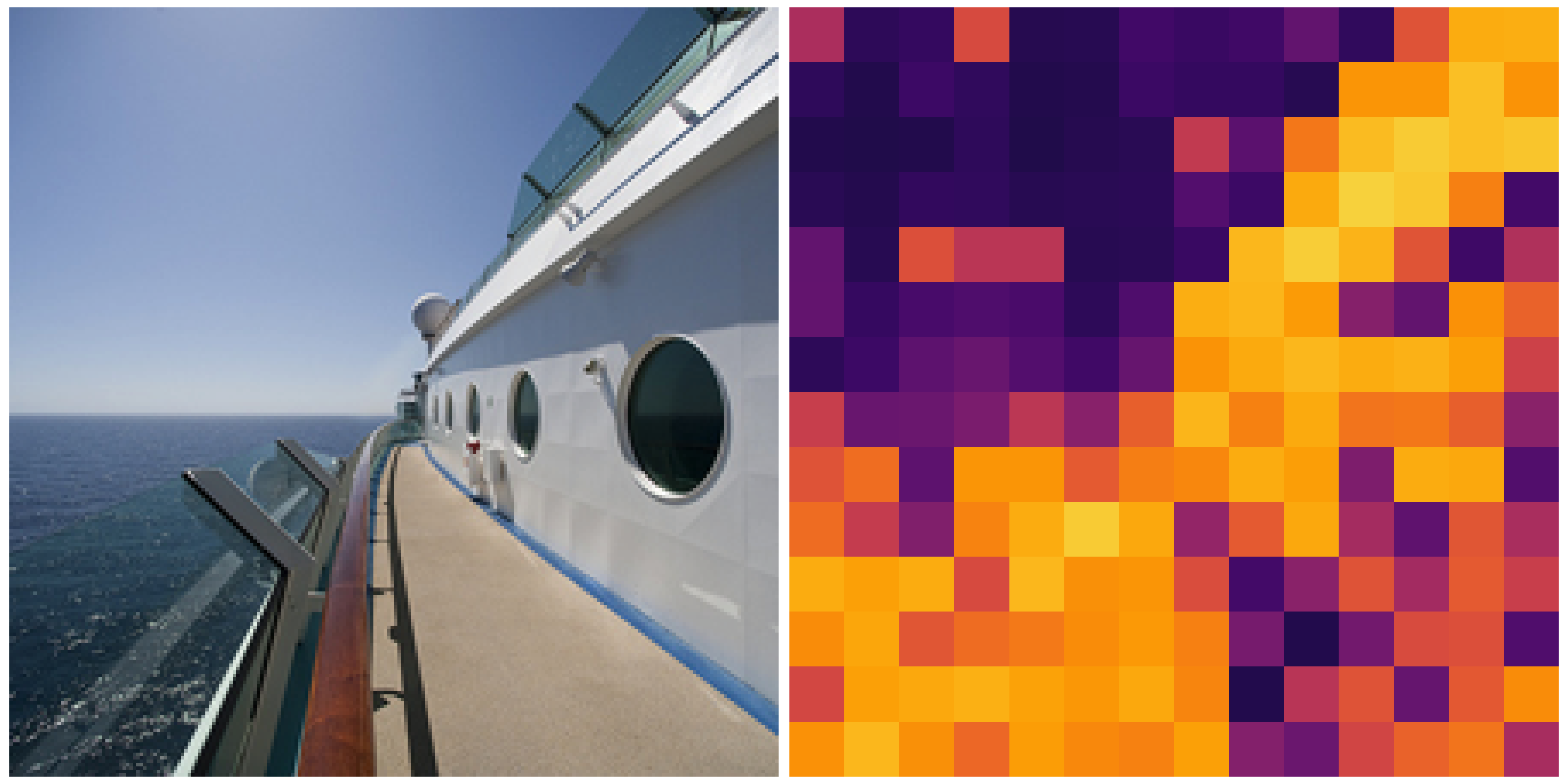}
    \end{minipage}
    \begin{minipage}[c]{\linewidth}
        \centering
        \includegraphics[width=0.243\linewidth]{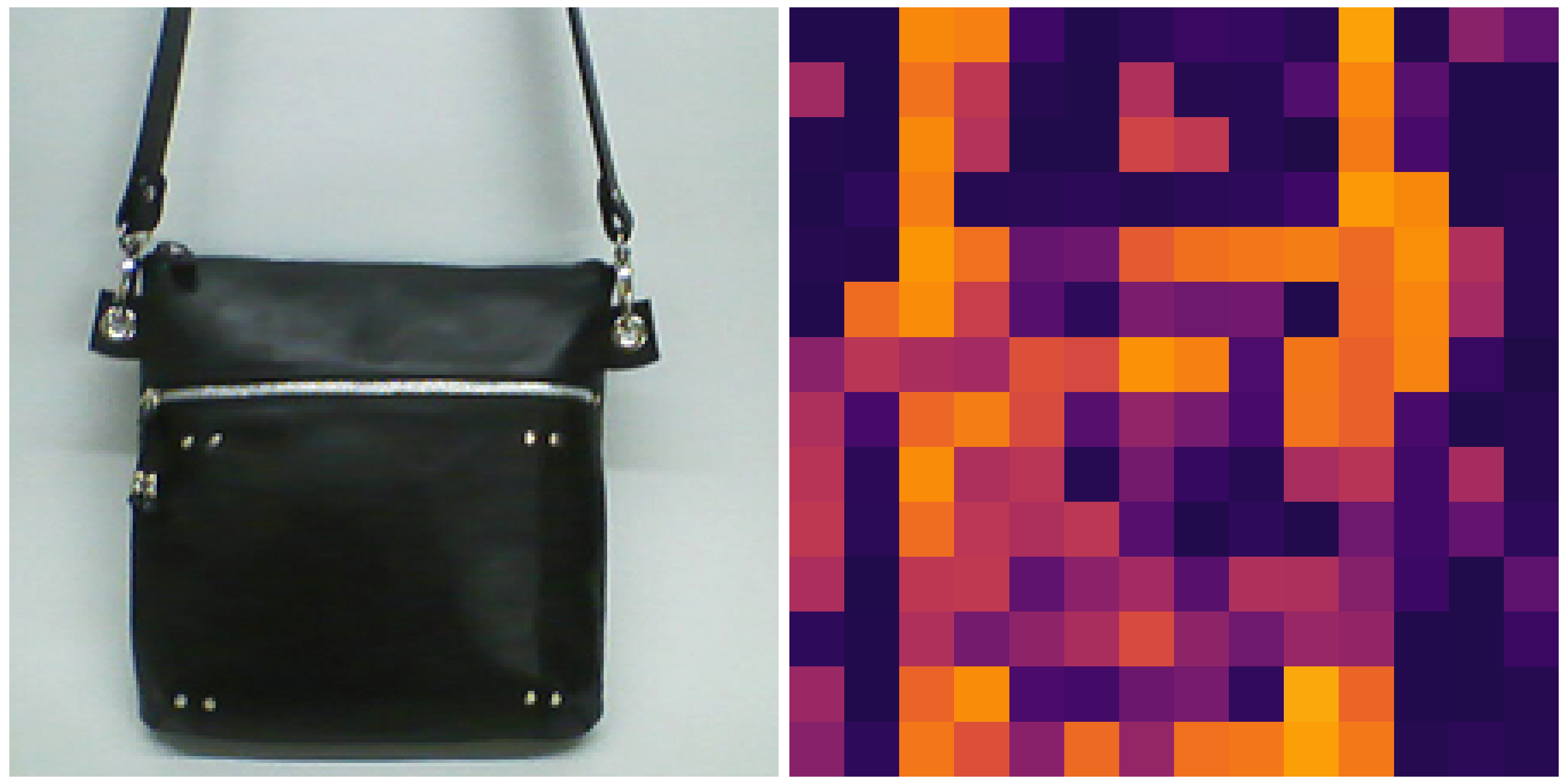}
        \hfill
        \includegraphics[width=0.243\linewidth]{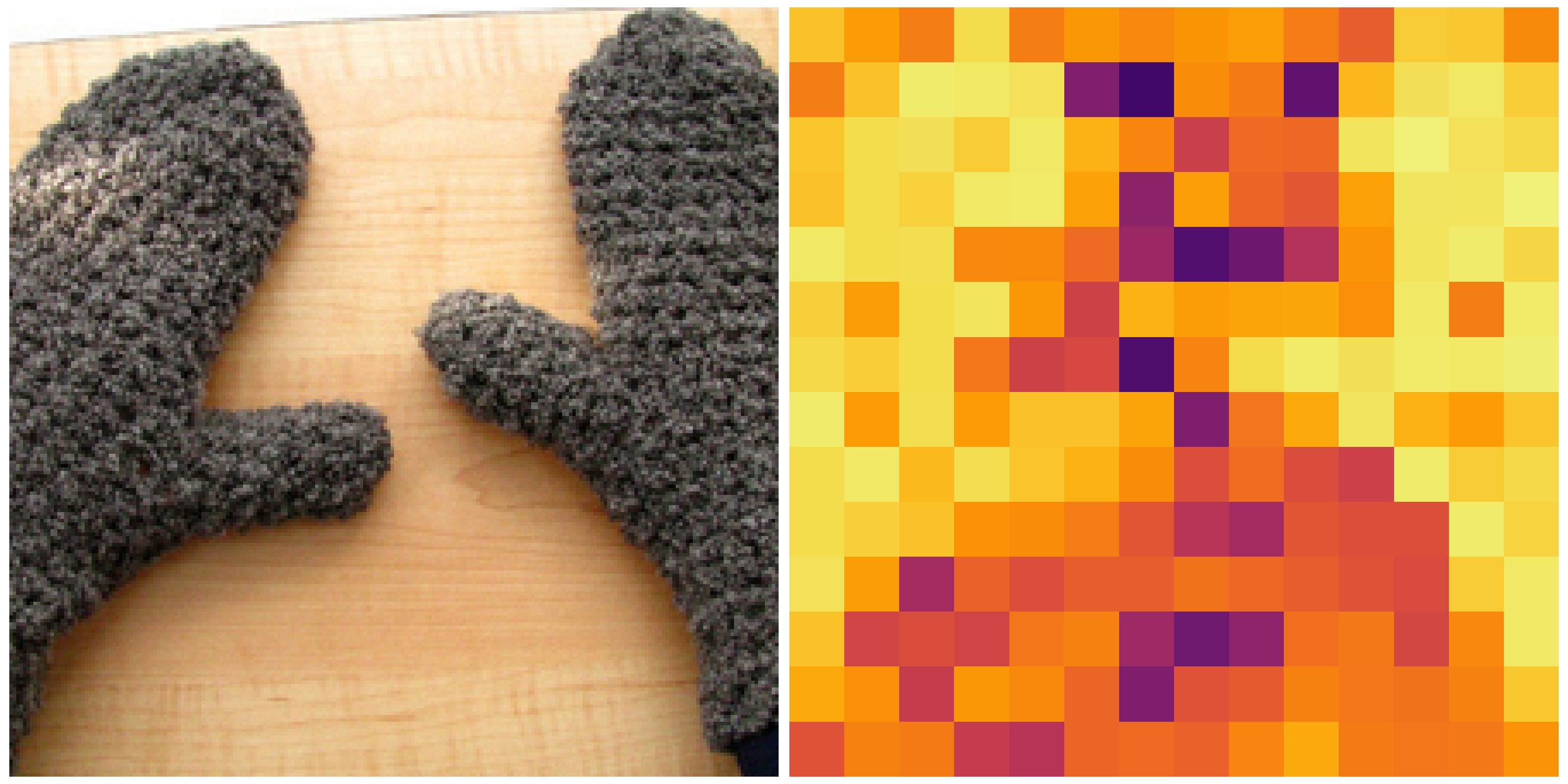}
        \hfill
        \includegraphics[width=0.243\linewidth]{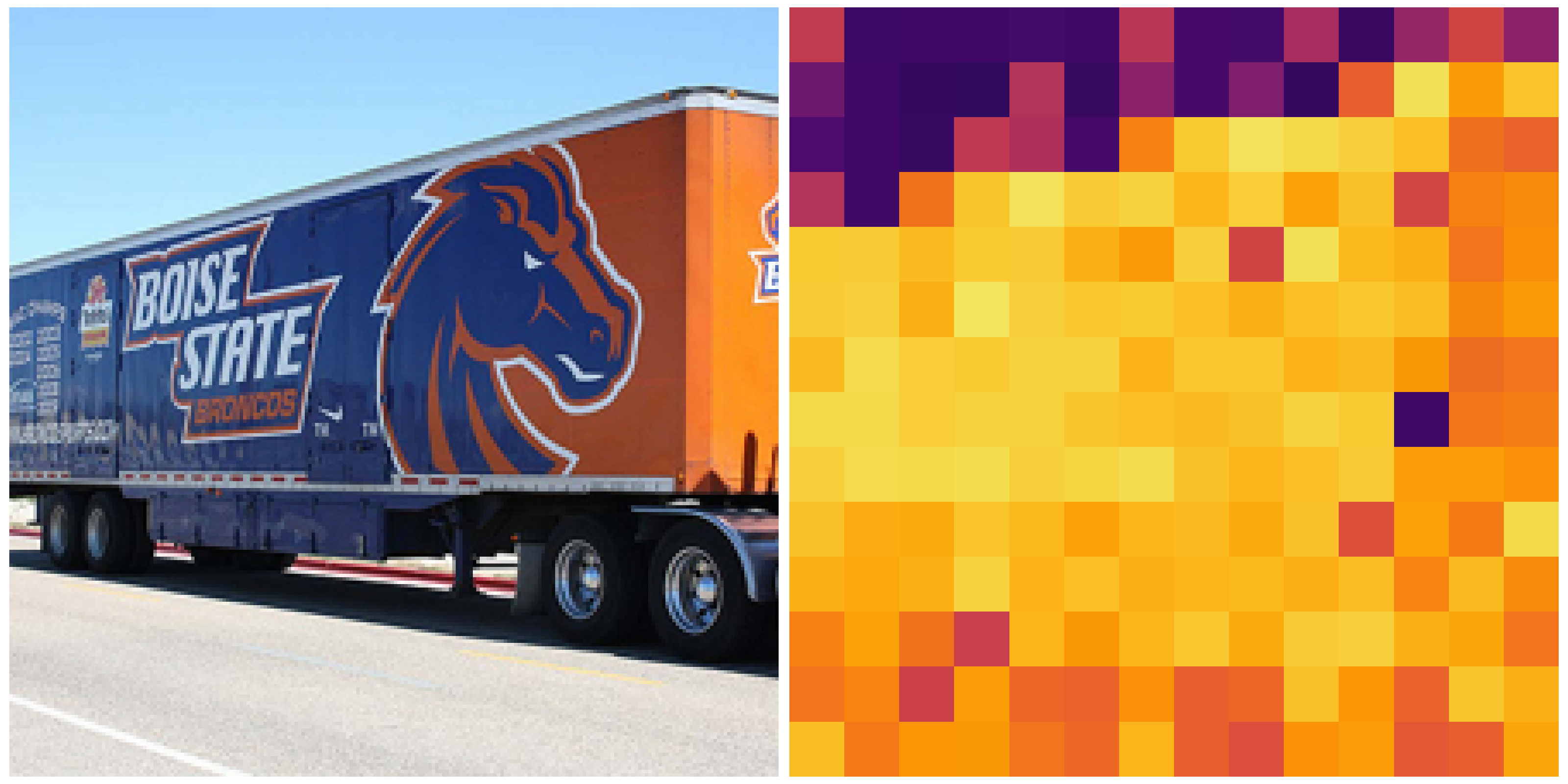}
        \hfill
        \includegraphics[width=0.243\linewidth]{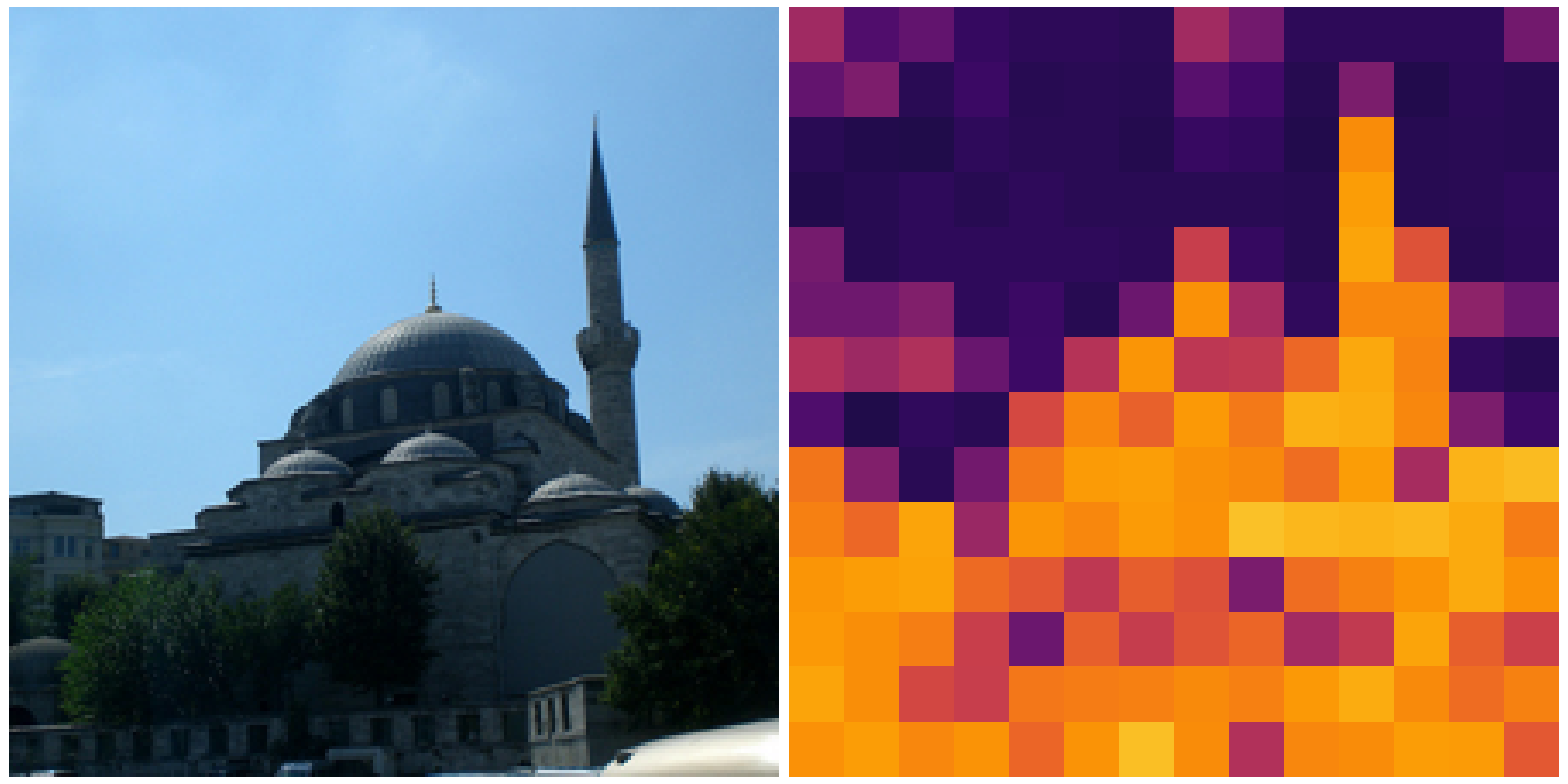}
    \end{minipage}
    \caption{Computational load heatmaps for additional samples from ImageNet-1k validation dataset. For each image, we show on the right the average fraction of learners that were activated in the ACMized model (the `computational load' of each token).}
    \label{fig:additional_heatmaps}
\end{figure*}

\begin{figure*}
    \centering
    \begin{subfigure}{\linewidth}
        \centering
        \includegraphics[width=0.12\linewidth]{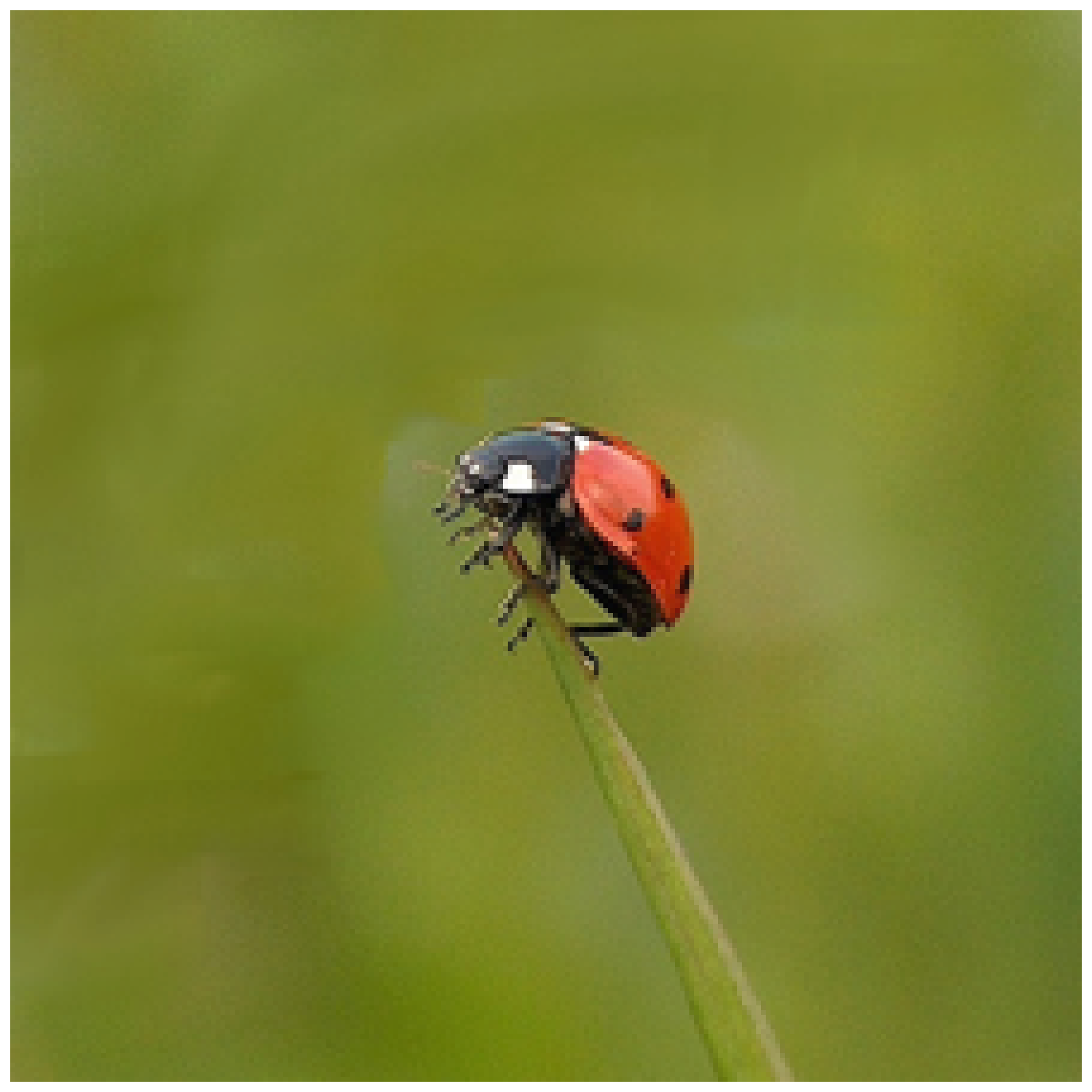}%
        \hfill
        \includegraphics[width=0.12\linewidth]{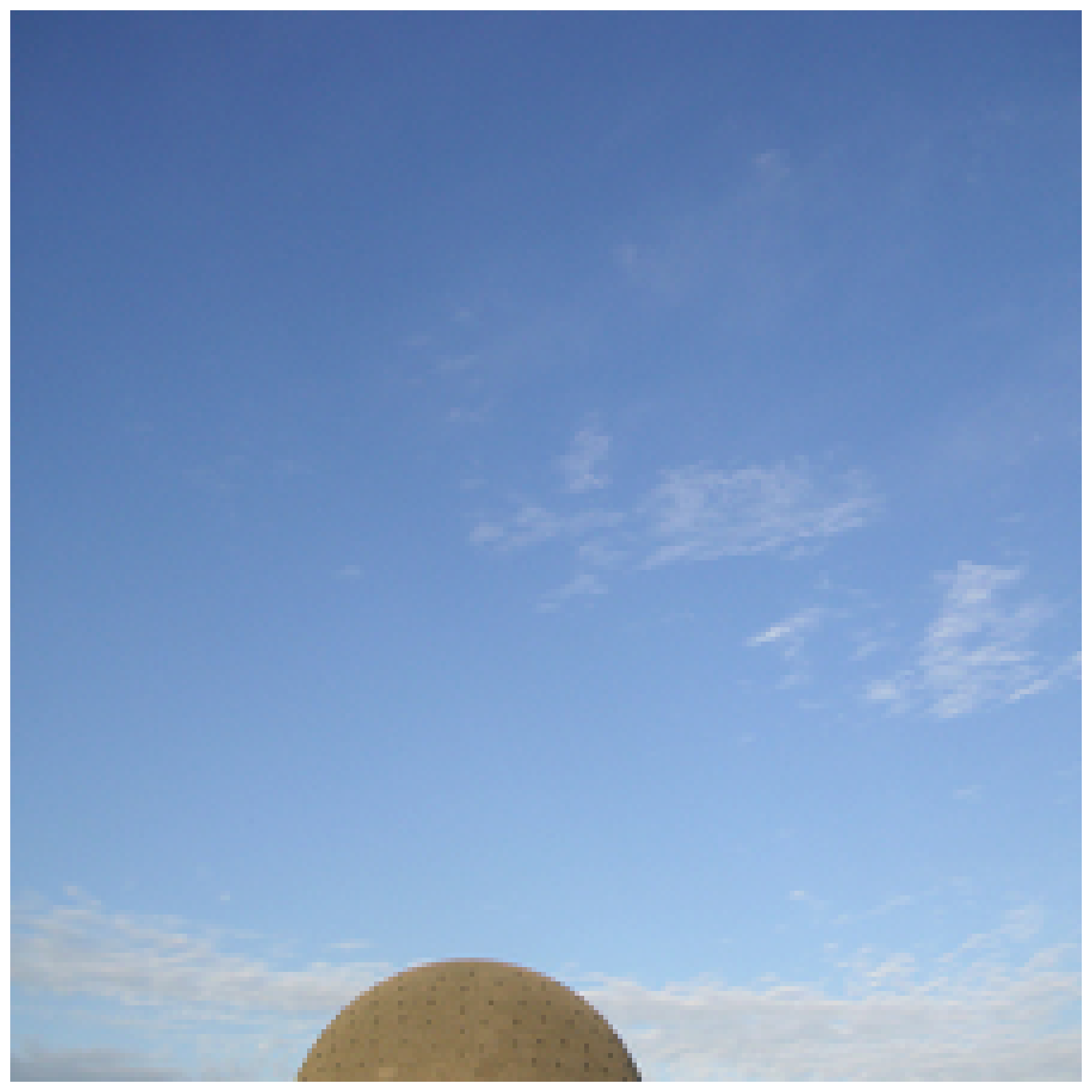}%
        \hfill
        \includegraphics[width=0.12\linewidth]{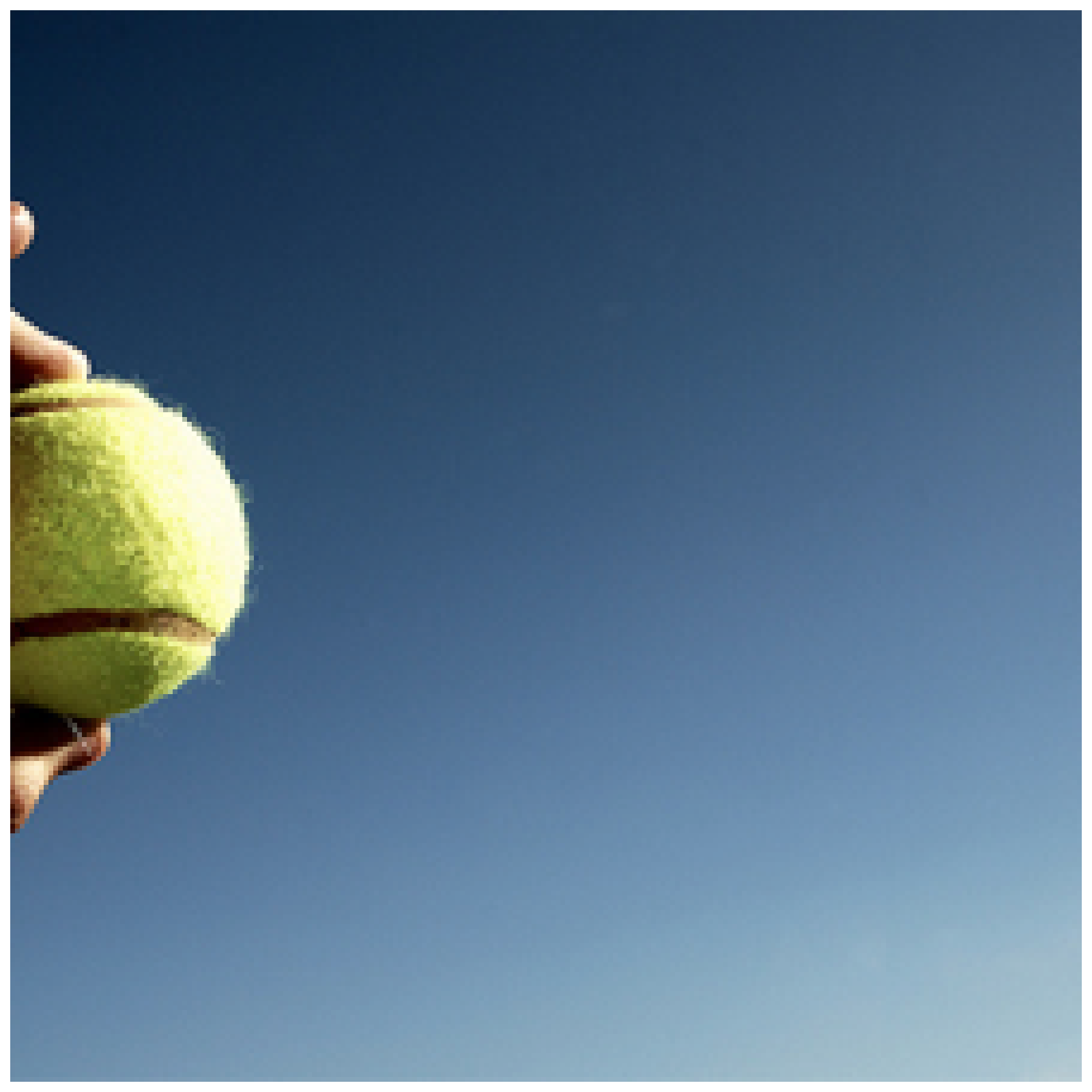}%
        \hfill
        \includegraphics[width=0.12\linewidth]{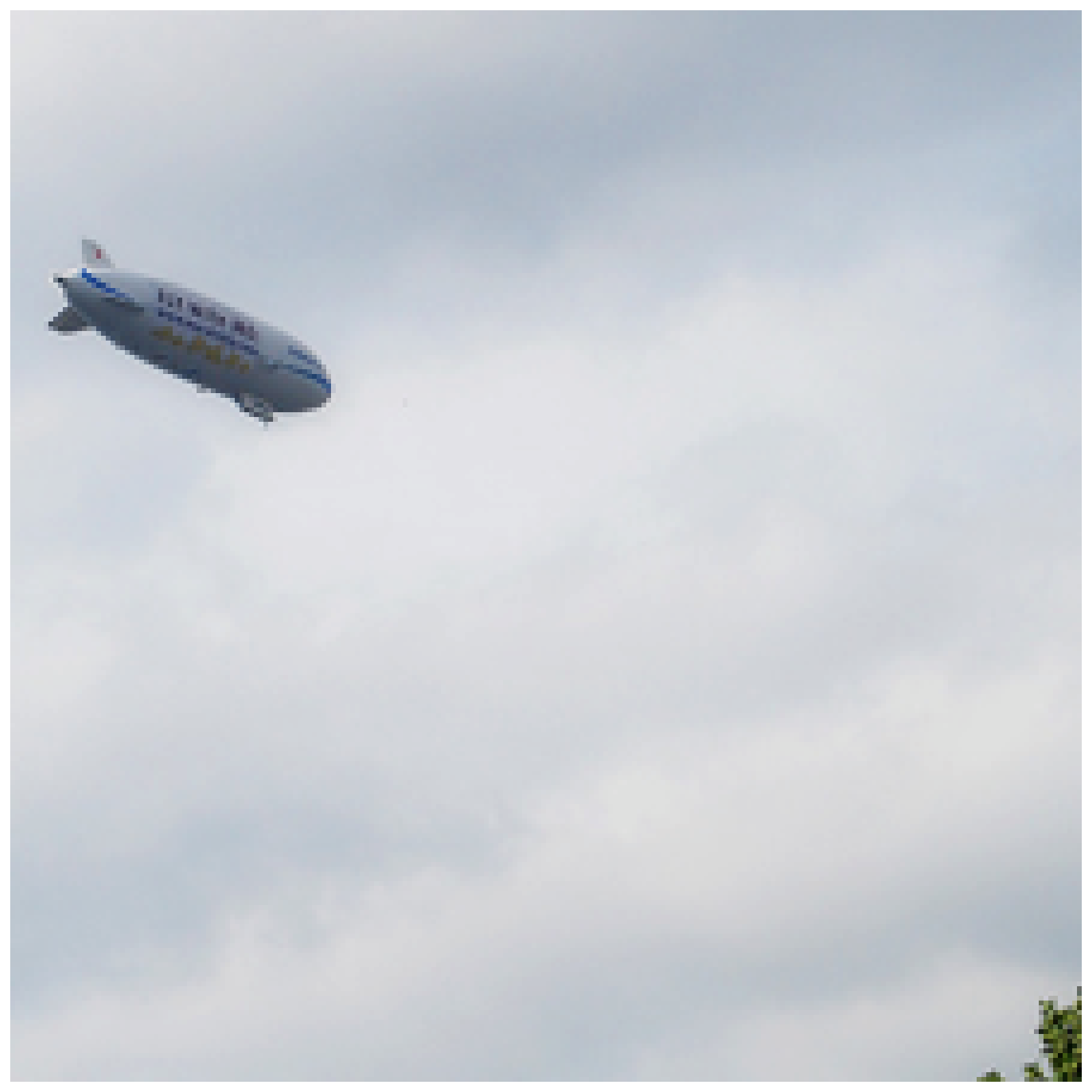}%
        \hfill
        \includegraphics[width=0.12\linewidth]{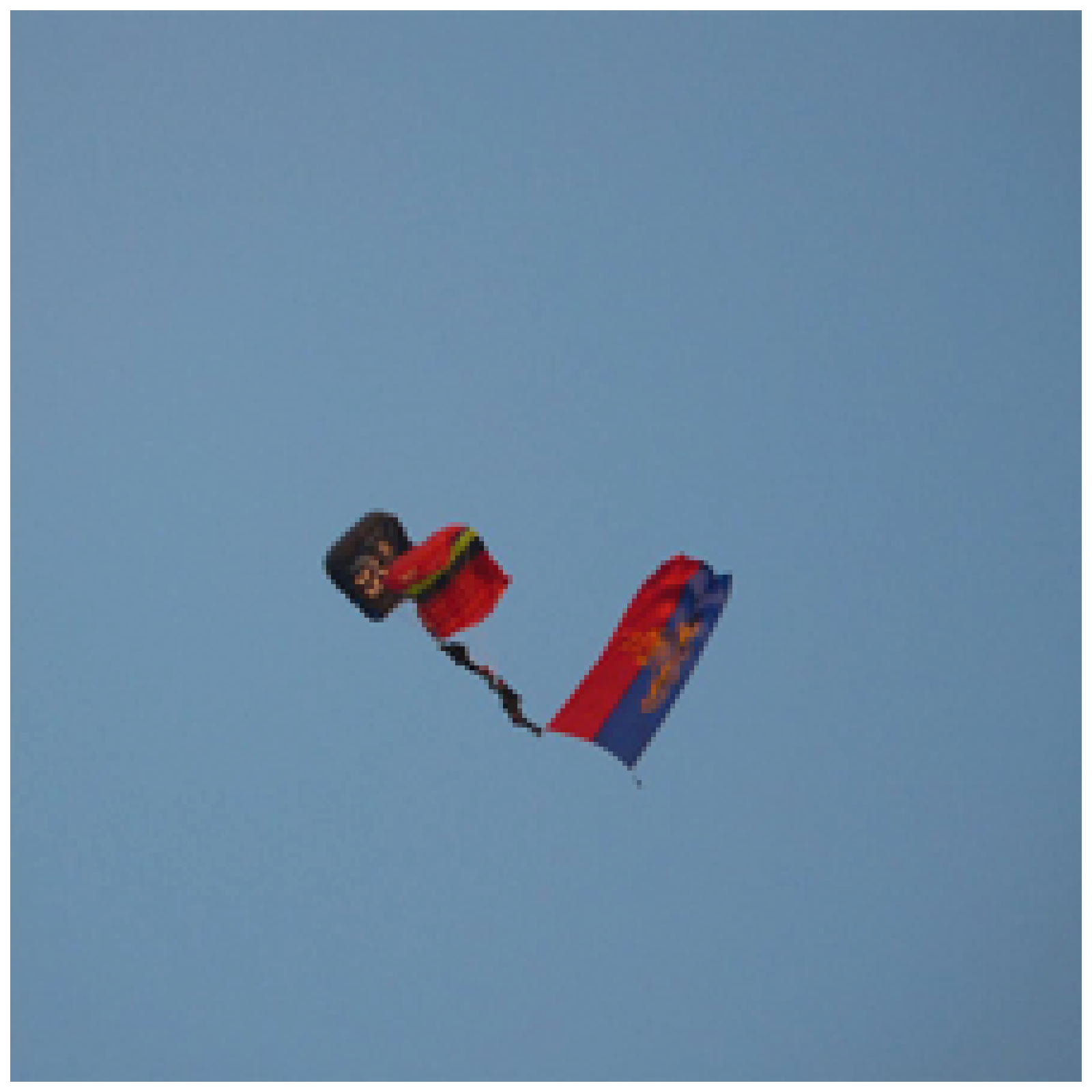}%
        \hfill
        \includegraphics[width=0.12\linewidth]{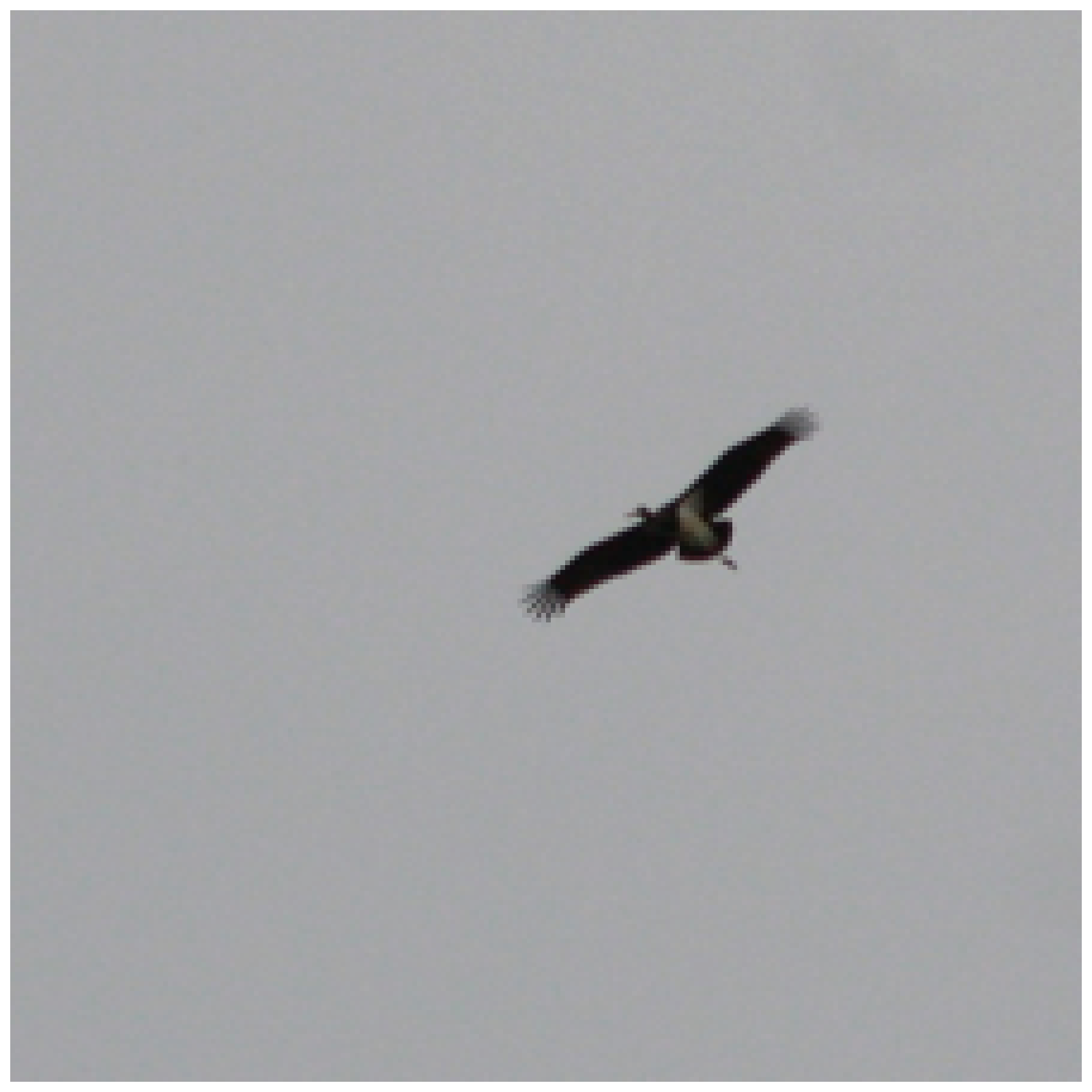}%
        \hfill
        \includegraphics[width=0.12\linewidth]{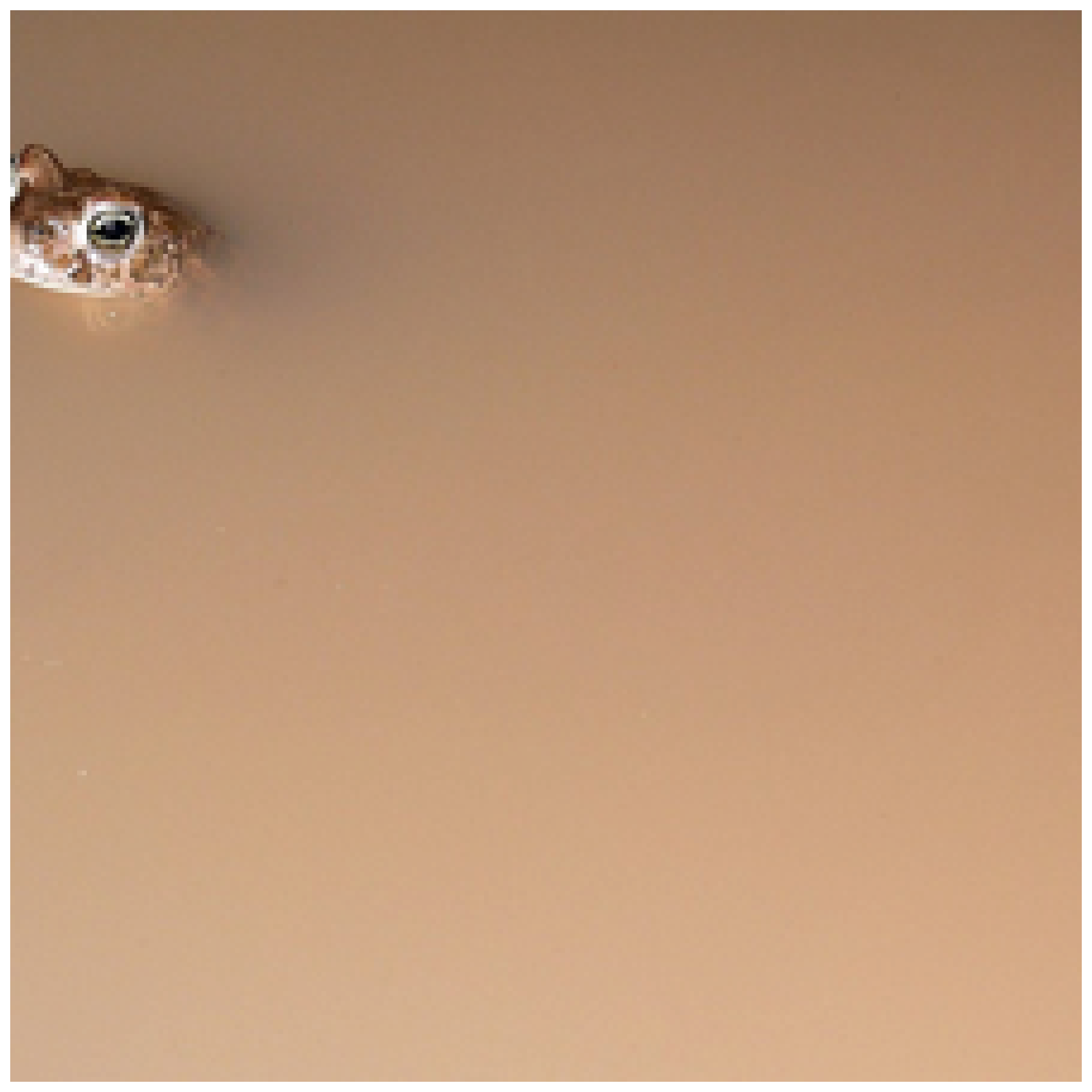}%
        \hfill
        \includegraphics[width=0.12\linewidth]{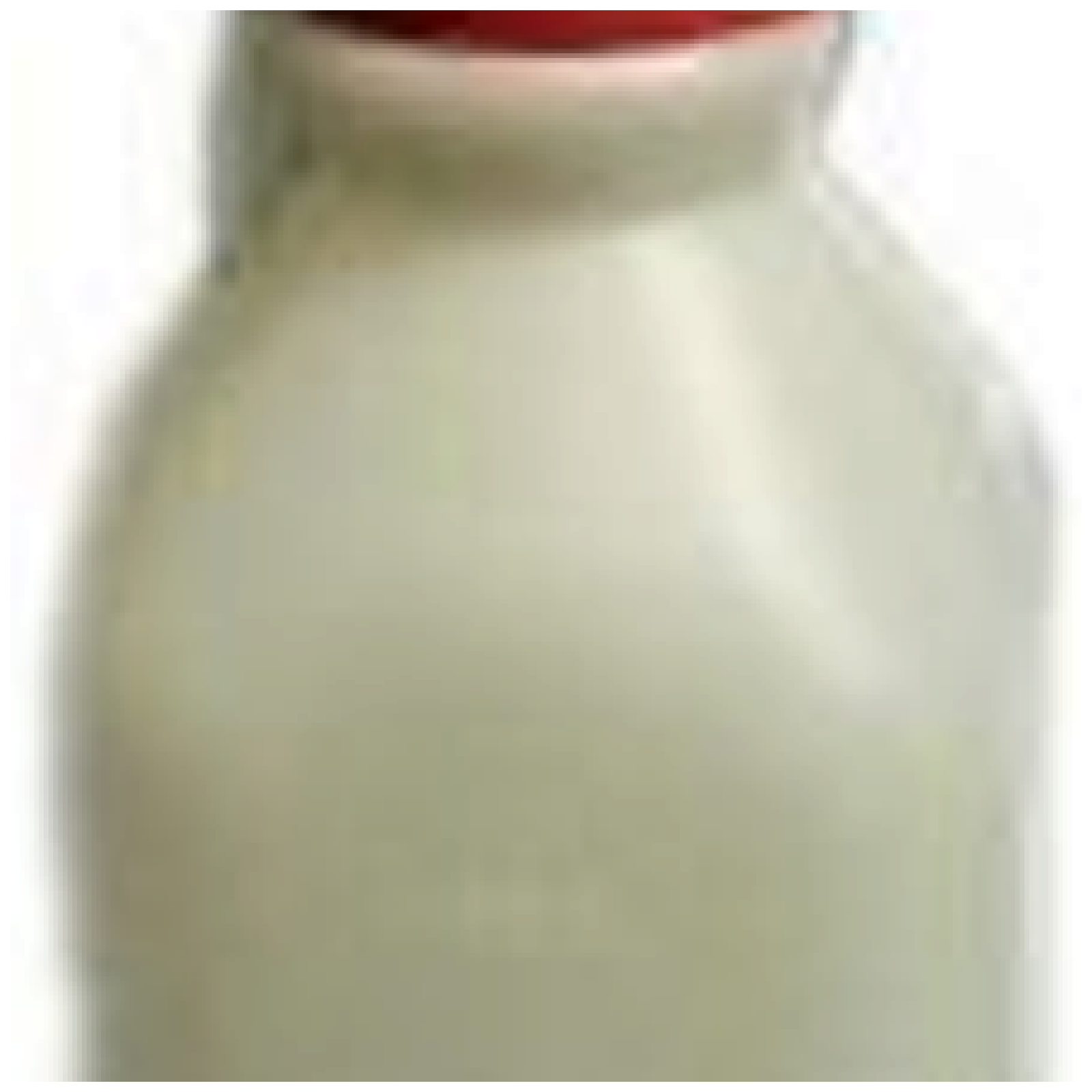}%
        \caption{Lowest allocated budget images}
    \end{subfigure}

    \begin{subfigure}{\linewidth}
        \centering
        \includegraphics[width=0.12\linewidth]{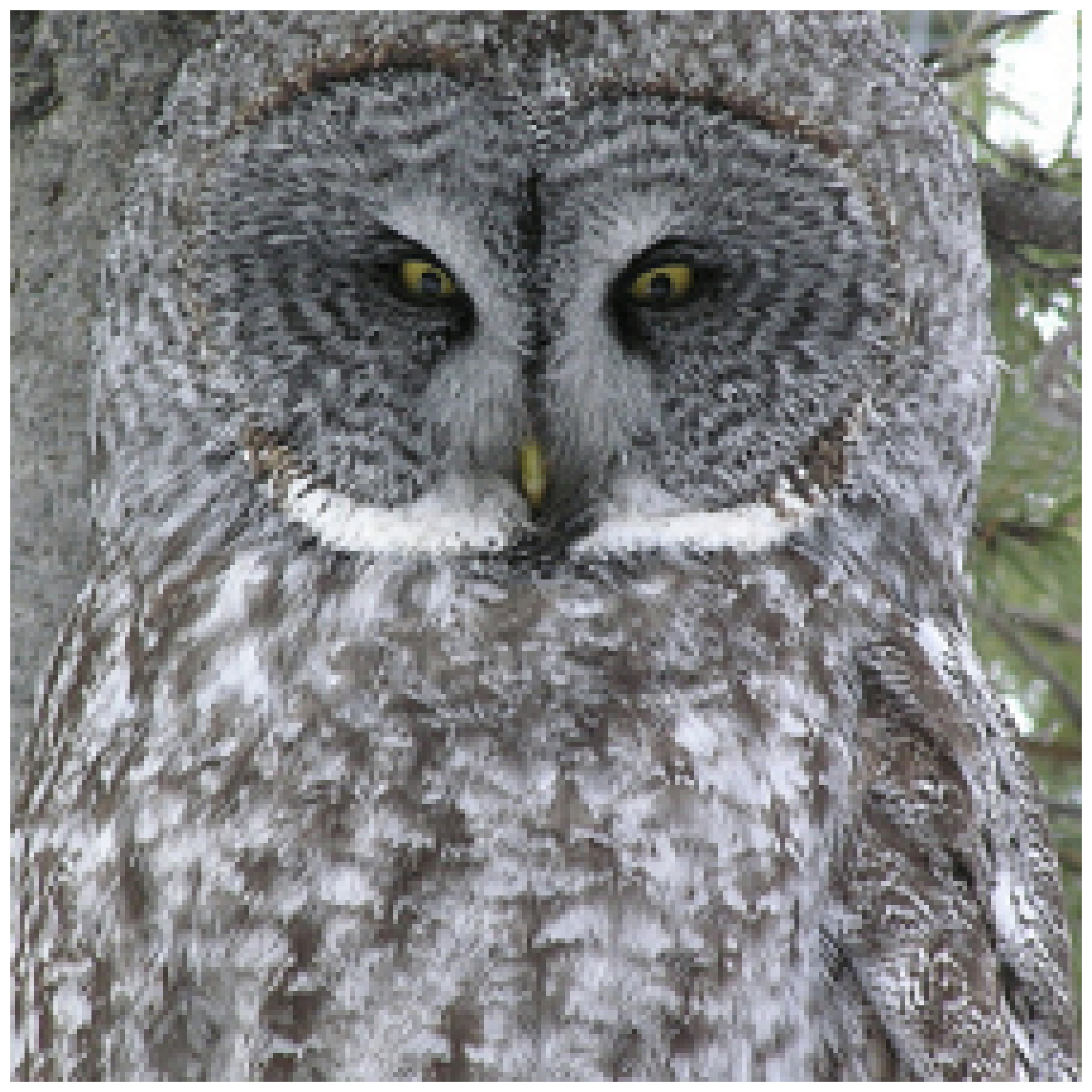}%
        \hfill
        \includegraphics[width=0.12\linewidth]{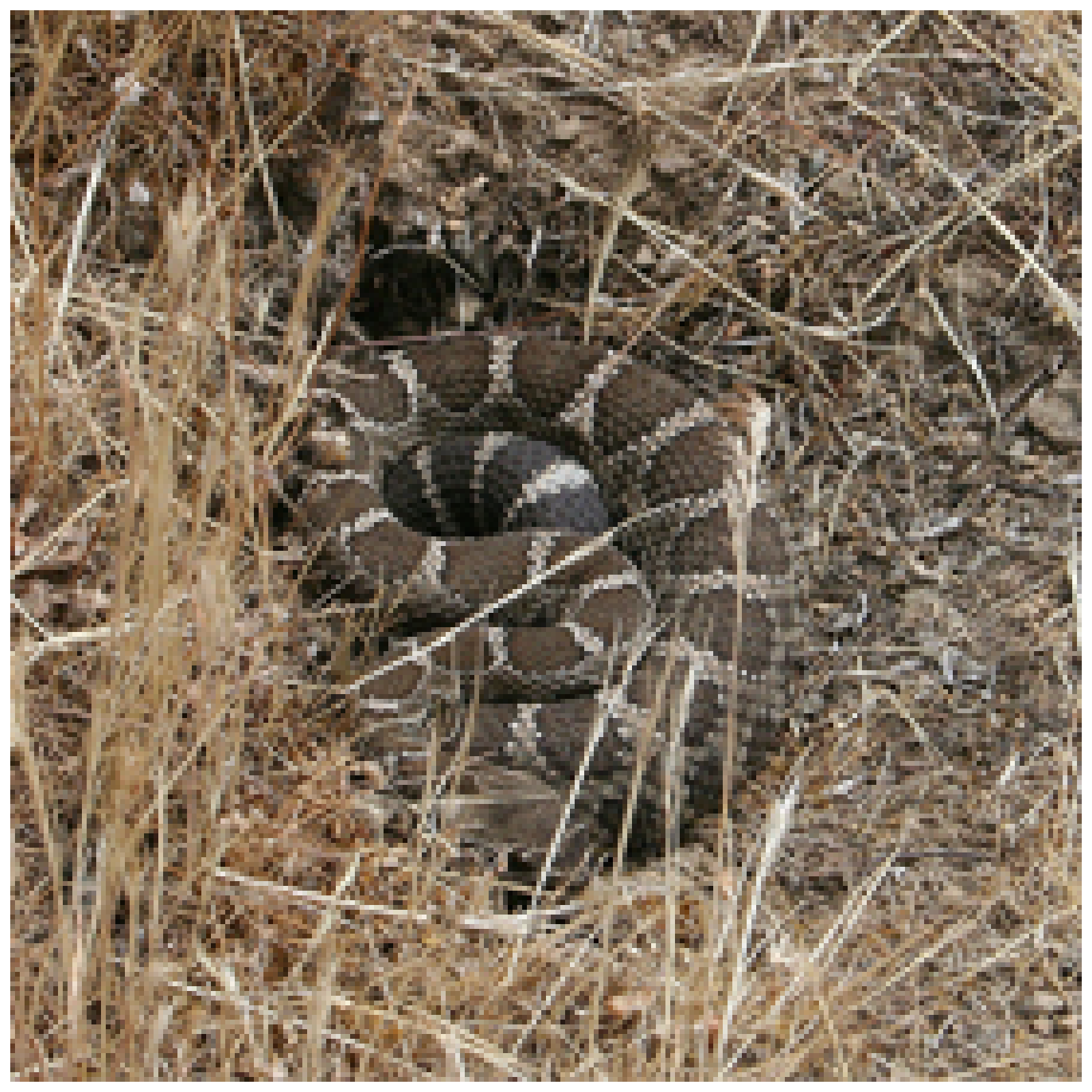}%
        \hfill
        \includegraphics[width=0.12\linewidth]{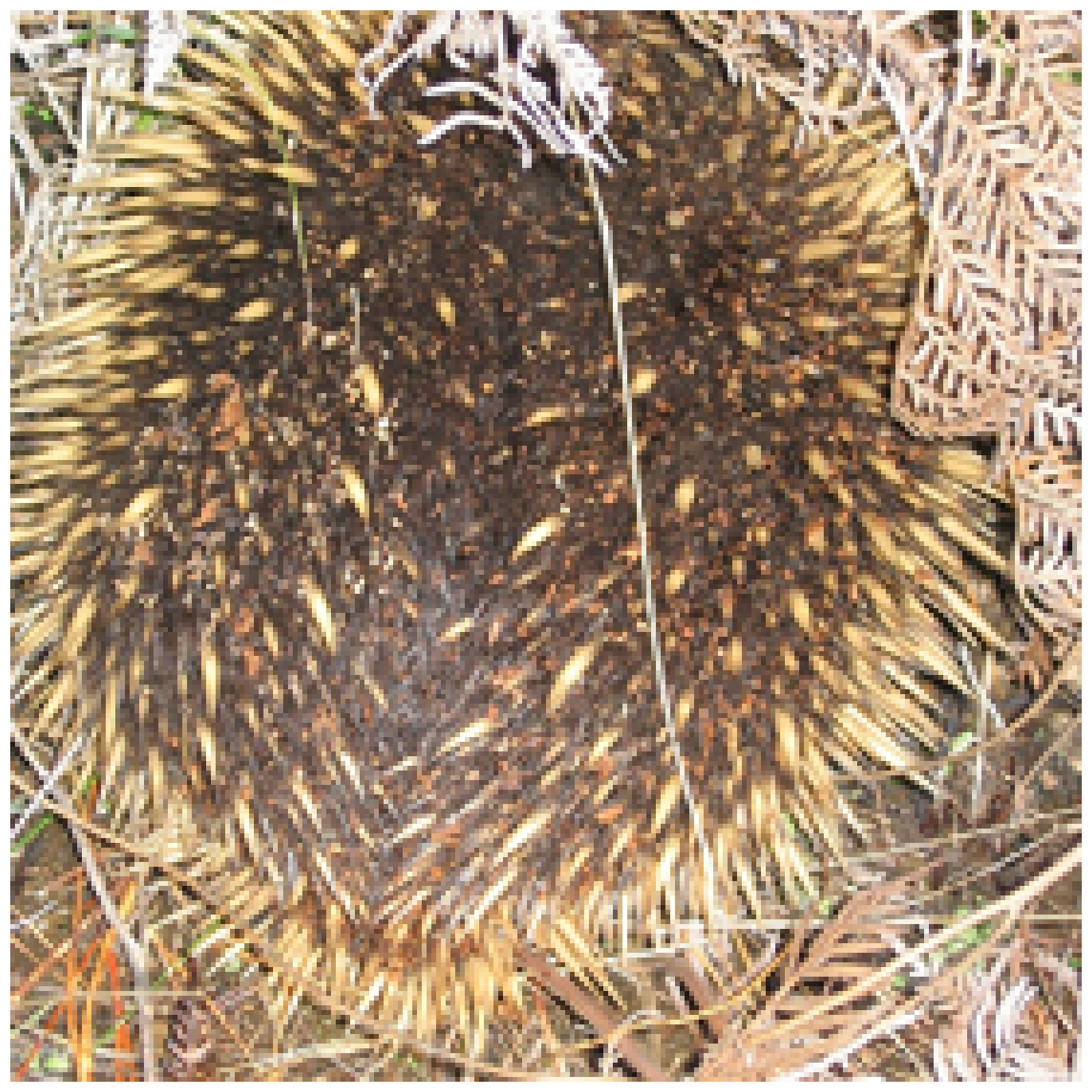}%
        \hfill
        \includegraphics[width=0.12\linewidth]{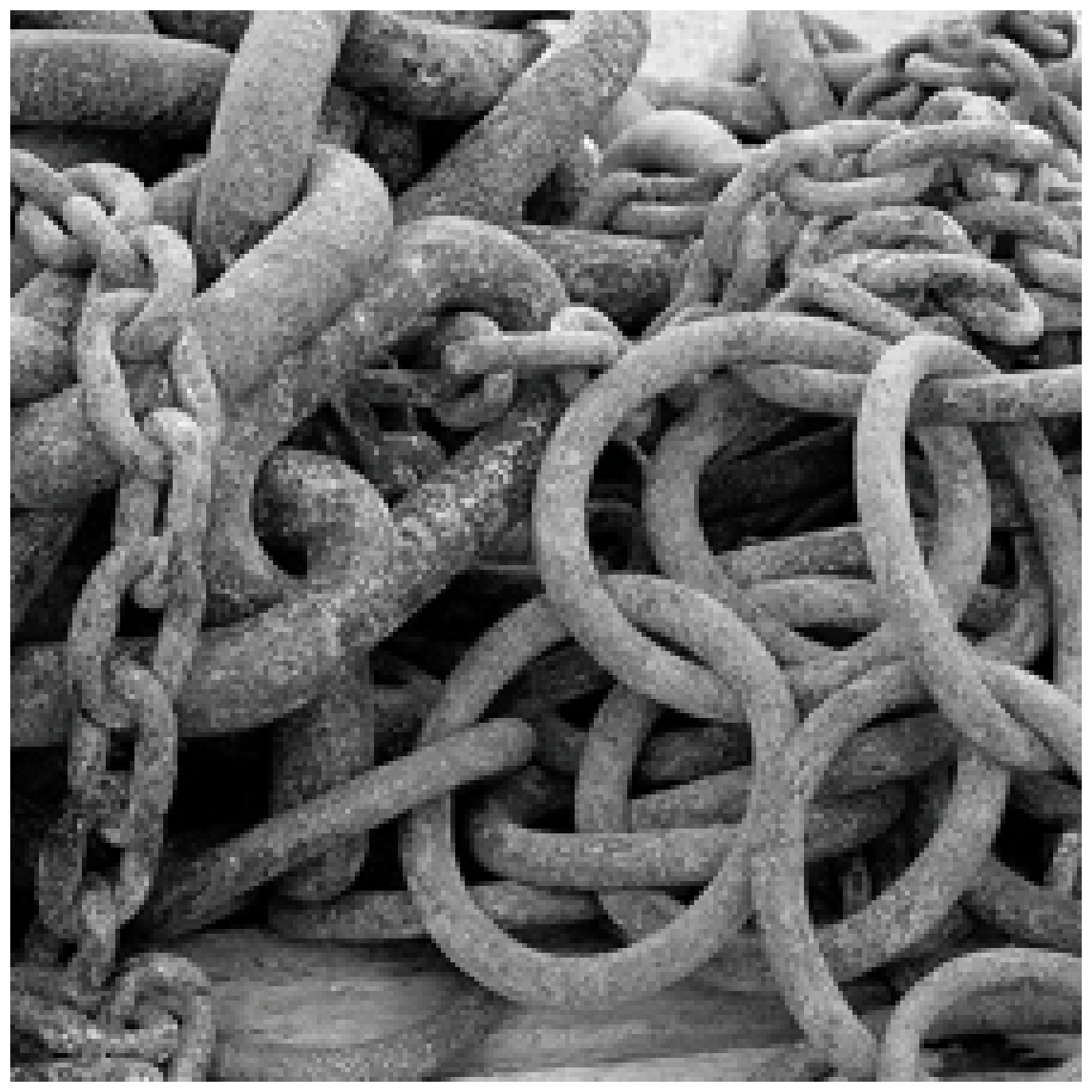}%
        \hfill
        \includegraphics[width=0.12\linewidth]{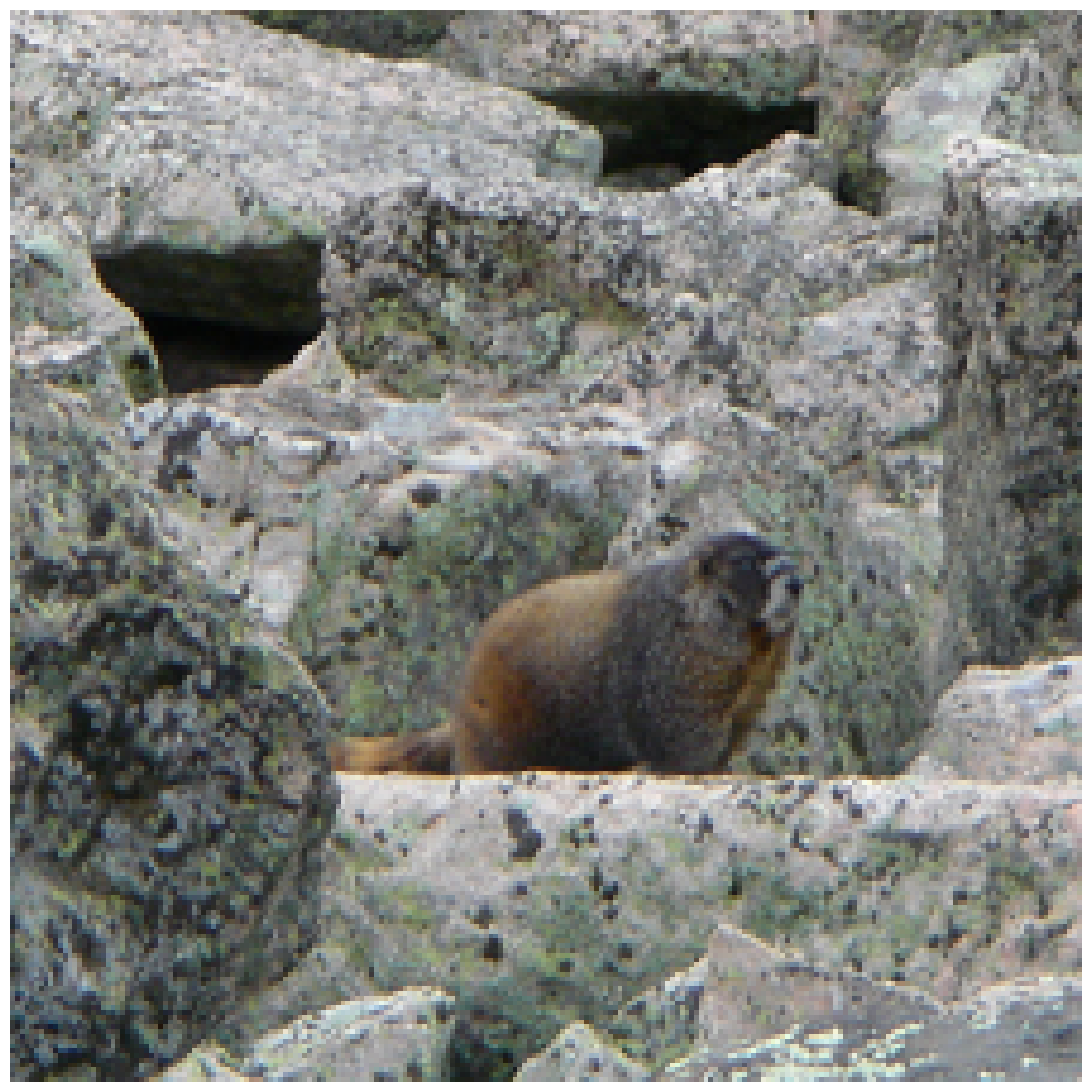}%
        \hfill
        \includegraphics[width=0.12\linewidth]{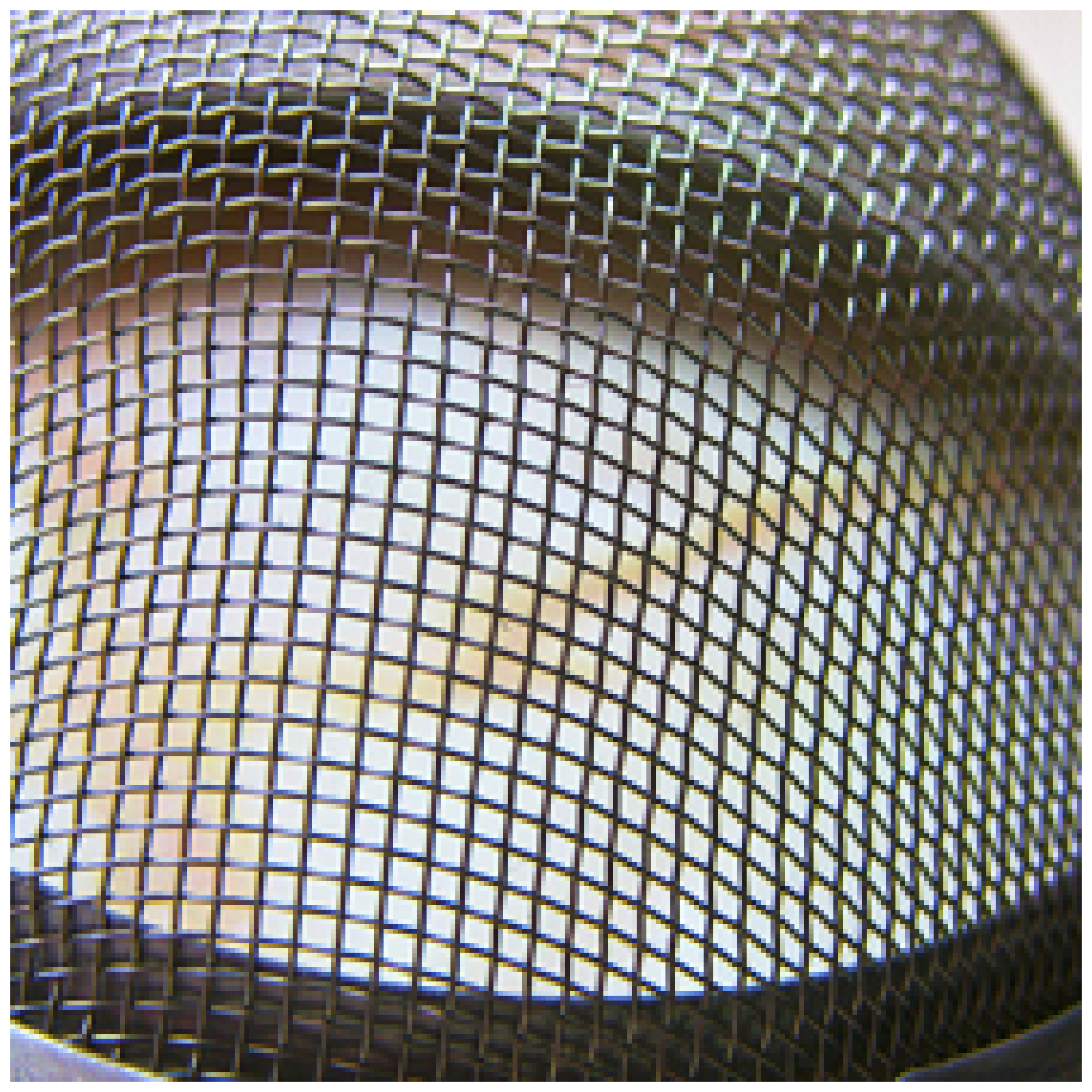}%
        \hfill
        \includegraphics[width=0.12\linewidth]{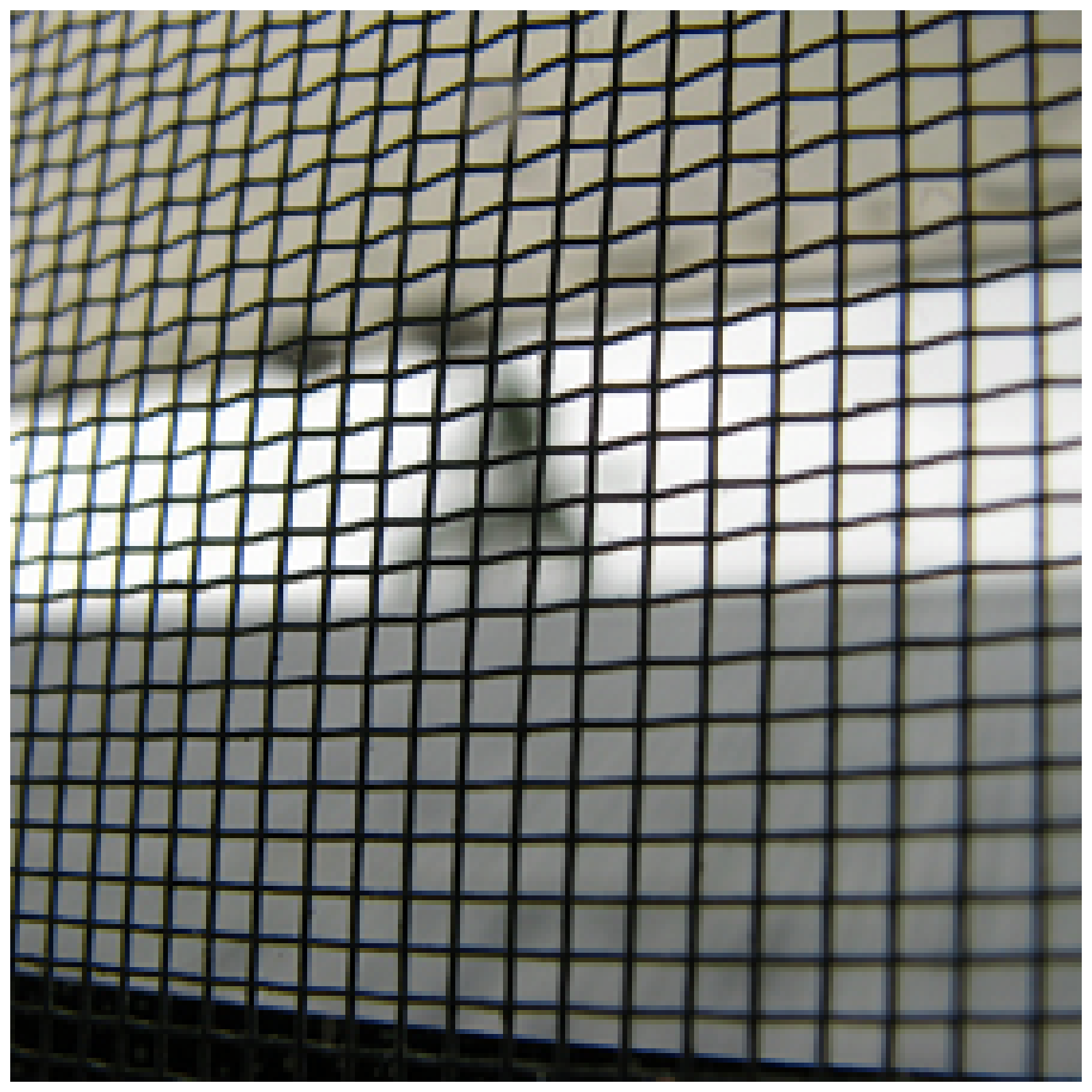}%
        \hfill
        \includegraphics[width=0.12\linewidth]{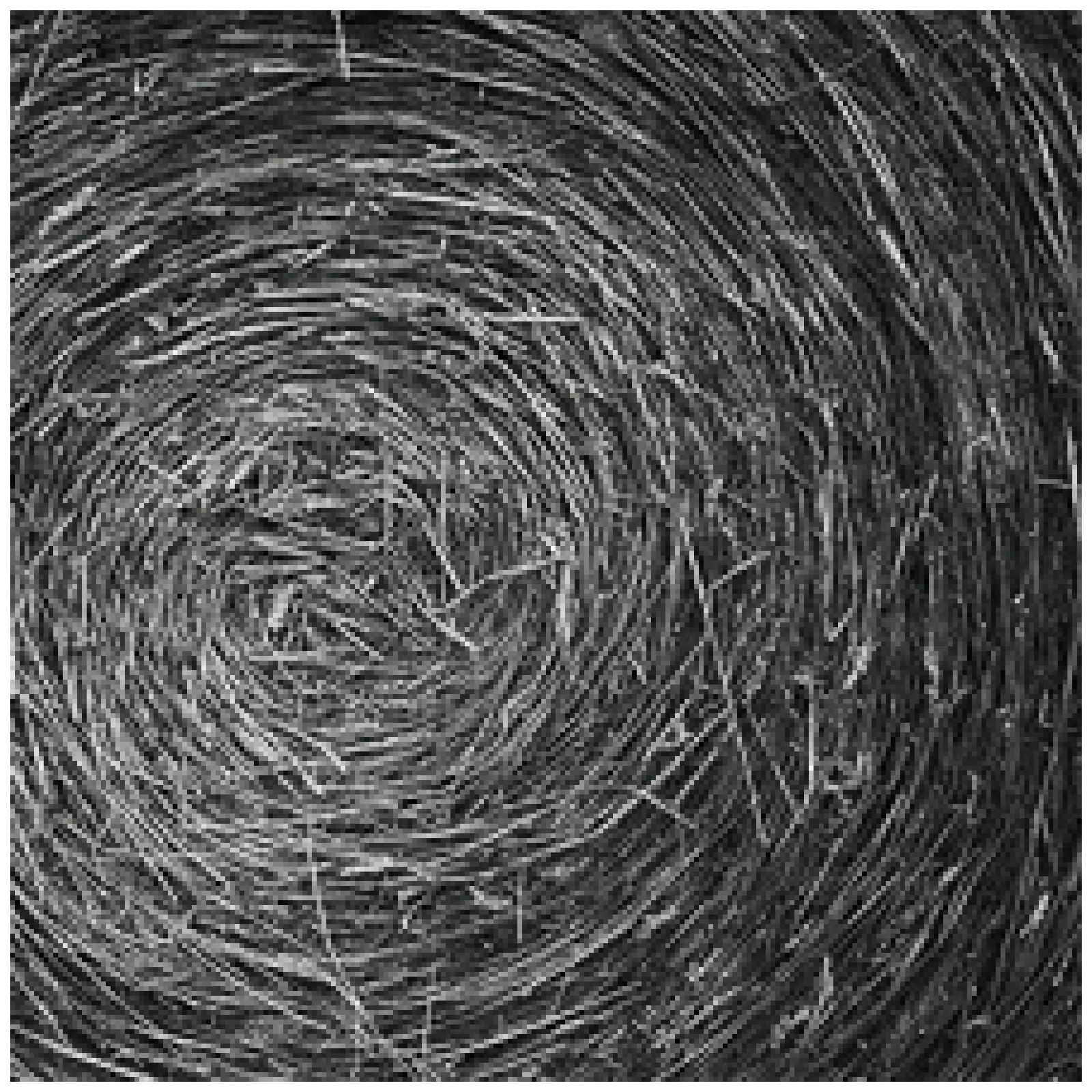}%
        \caption{Highest allocated budget images}
    \end{subfigure}
    \caption{Samples that consumed the least (top row) and most (bottom row) computation of the model trained with $\beta_{\text{target}} = 0.6$. We can see that the model perceives images containing high-frequency content as significantly harder.%
    \label{fig:cheapest_costliest_samples}}
\end{figure*}

We show in \Cref{fig:additional_heatmaps} additional heatmaps for our method, which highlight total amount of activated learners across all layers of the ACMized model for each patch (see Fig. 7 in the main paper). We can see the model allocates almost zero budget to simple background patches, while for the rest of the images the amount of activated learners correlates with the visual complexity of the objects.
We observe that the vision model mostly learned to ignore blurred and low-frequency content. To further confirm this, we procedurally find validation dataset samples that are the most and least computationally demanding for the model, and present them in Figure \Cref{fig:cheapest_costliest_samples}.

\section{Speech-to-text computational load}

\begin{figure*}
    \begin{minipage}[c]{\linewidth}
        \centering
        \includegraphics[width=0.44\linewidth]{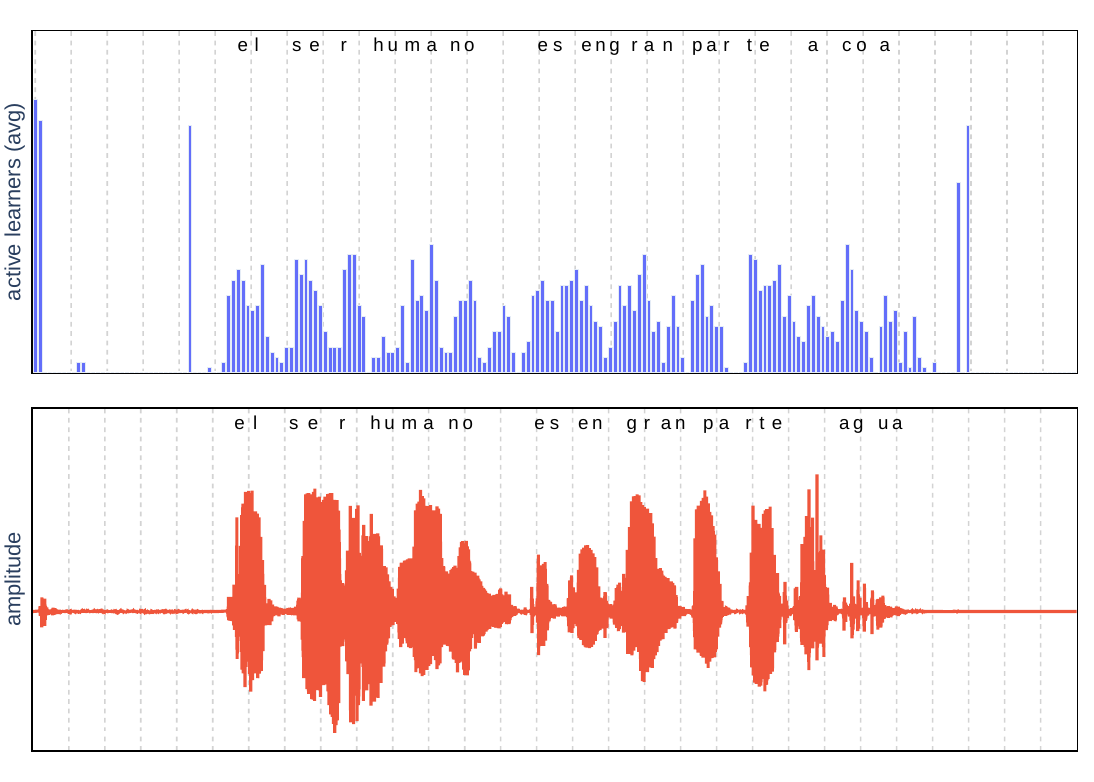}
        \hfill
        \includegraphics[width=0.44\linewidth]{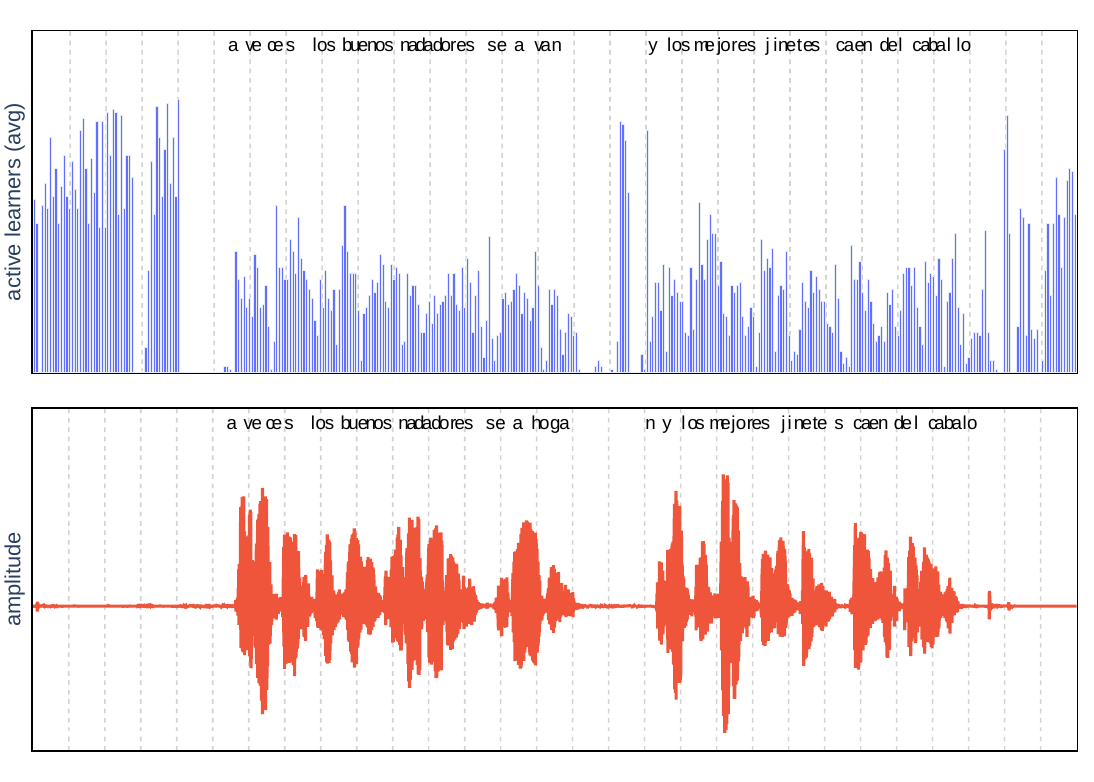}
    \end{minipage}
    \begin{minipage}[c]{\linewidth}
        \centering
        \includegraphics[width=0.44\linewidth]{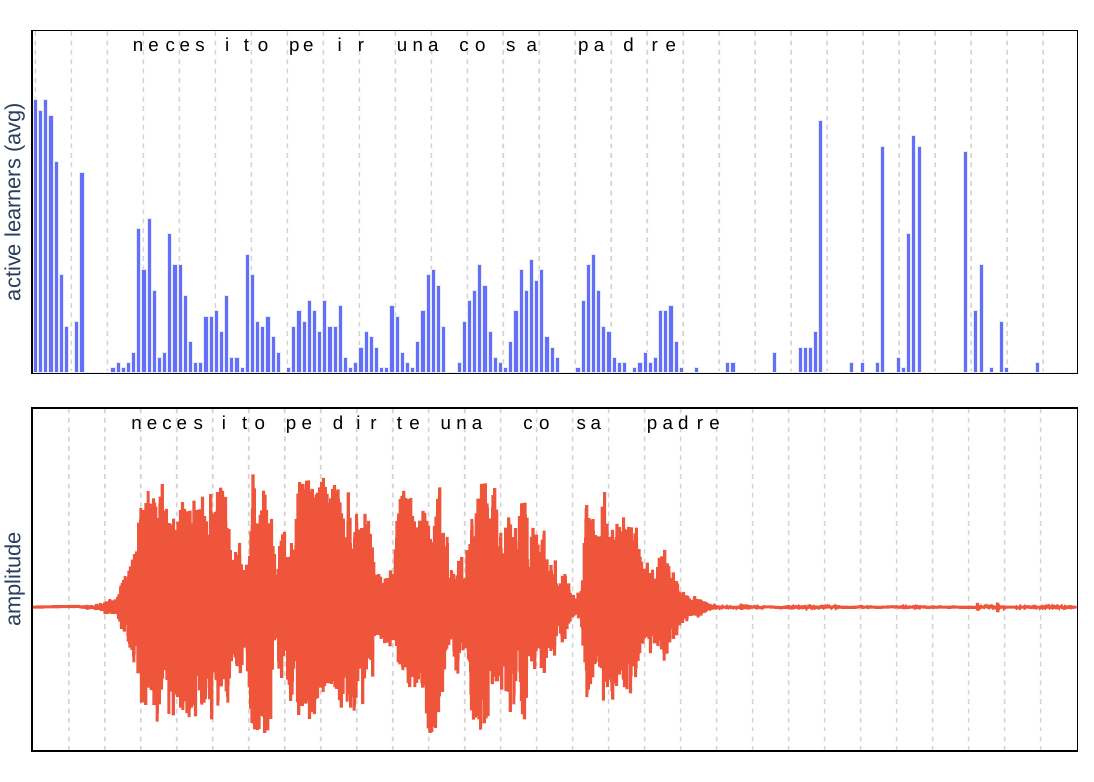}
        \hfill
        \includegraphics[width=0.44\linewidth]{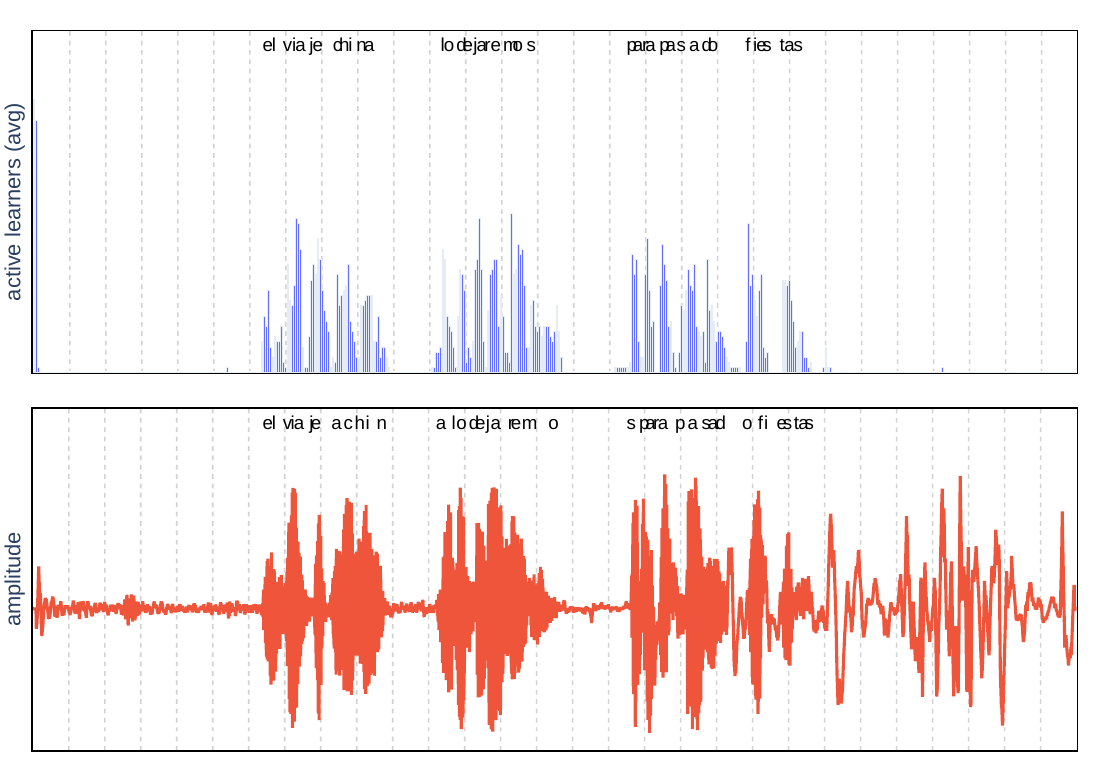}
    \end{minipage}
    \begin{minipage}[c]{\linewidth}
        \centering
        \includegraphics[width=0.44\linewidth]{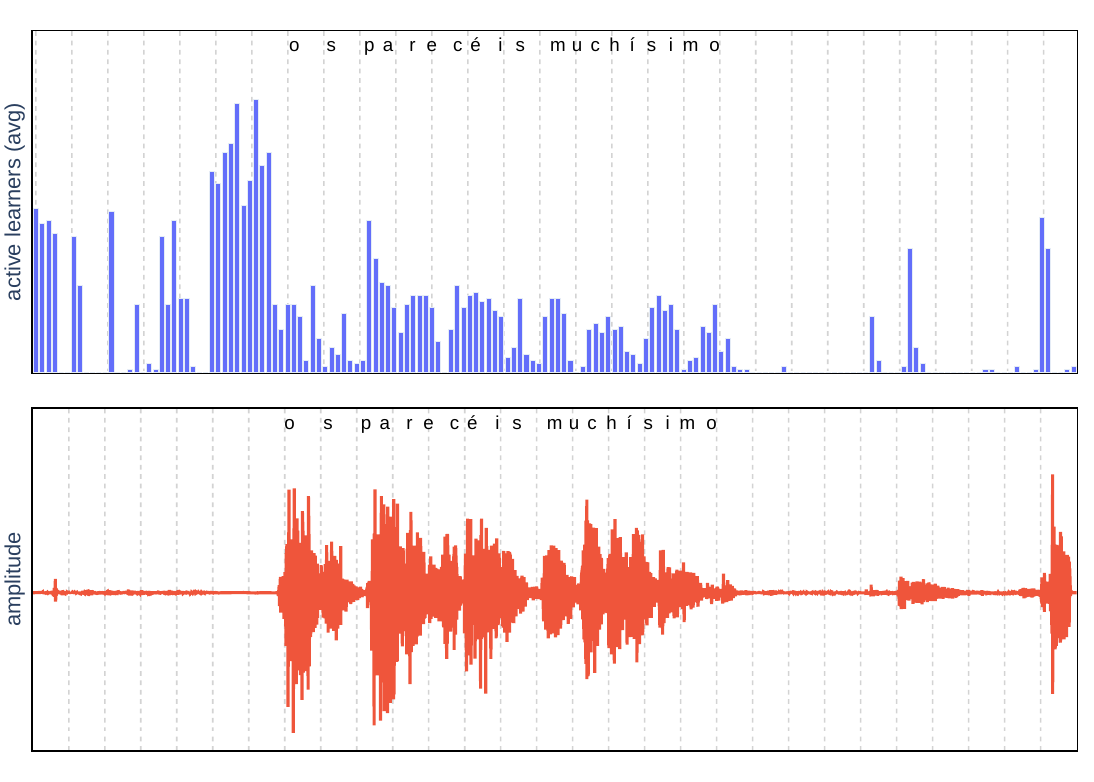}
        \hfill
        \includegraphics[width=0.44\linewidth]{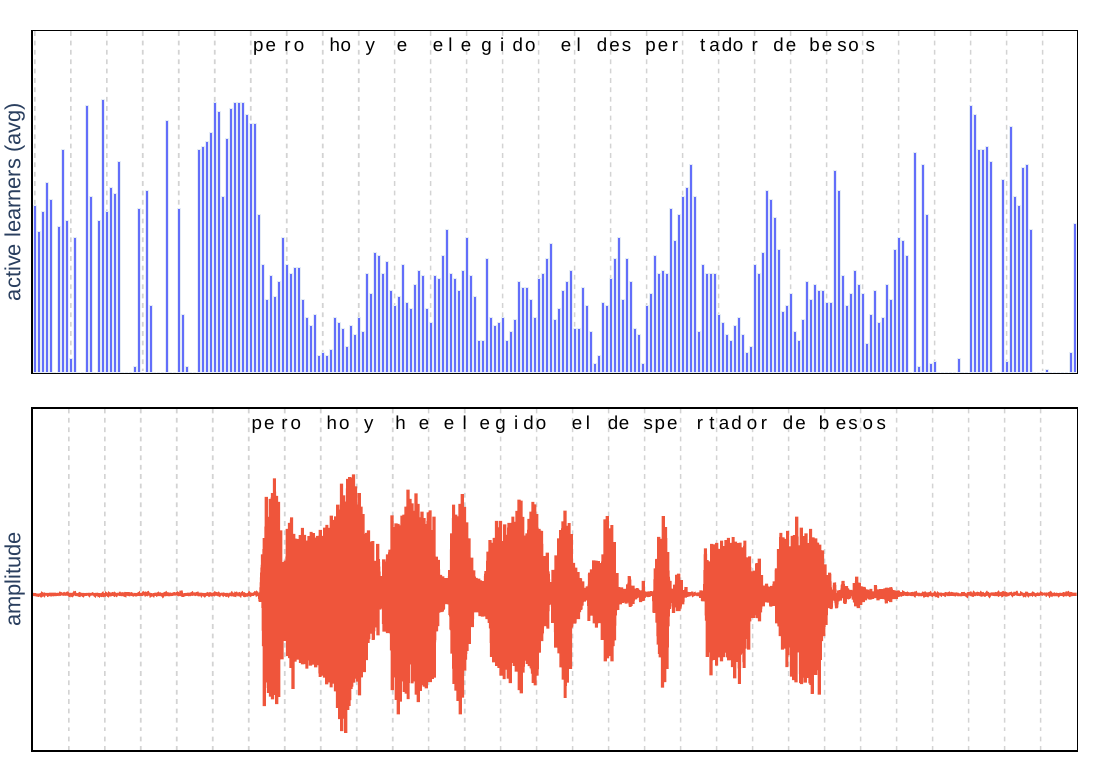}
    \end{minipage}
    \begin{minipage}[c]{\linewidth}
        \centering
        \includegraphics[width=0.44\linewidth]{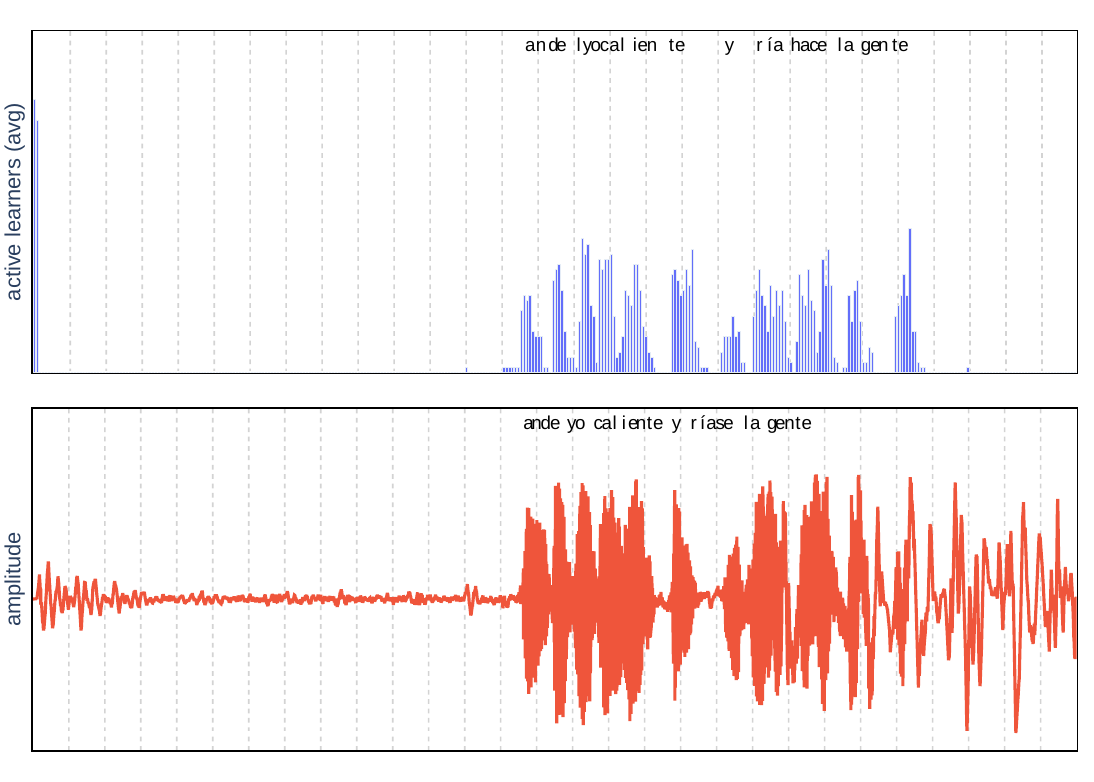}
        \hfill
        \includegraphics[width=0.44\linewidth]{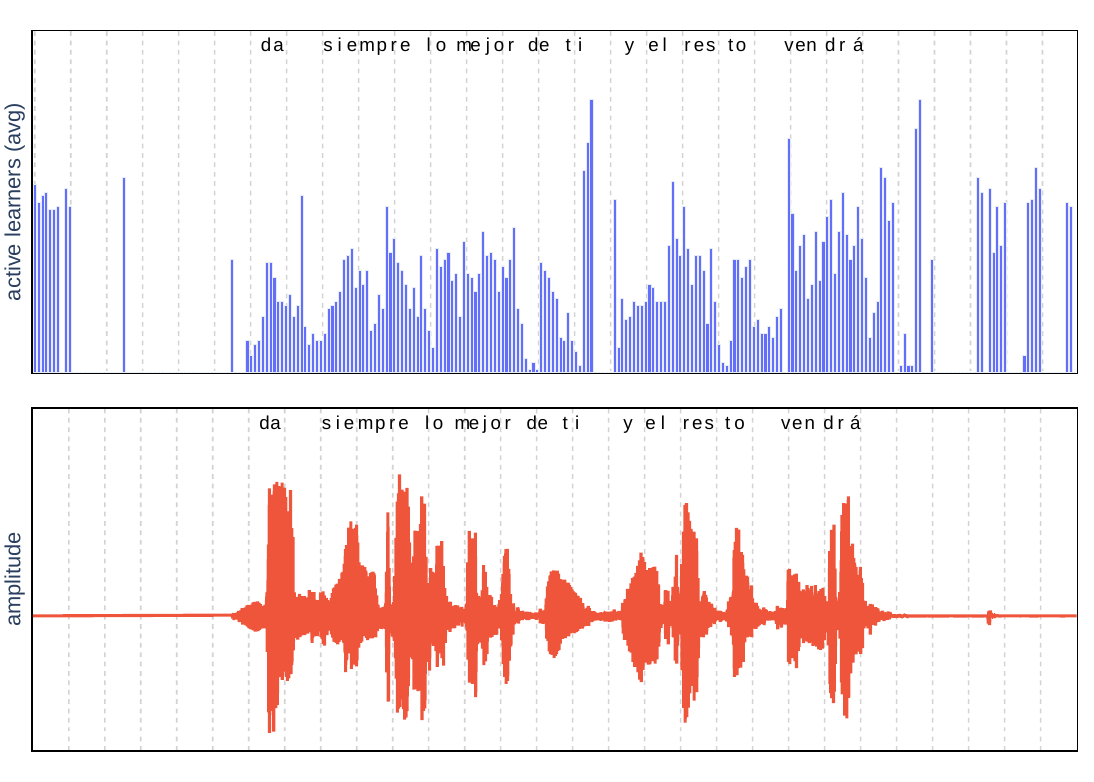}
    \end{minipage}
    \caption{Computational load for each input token for audio models. For each input speech recording (bottom), we show the average number of learners that were activated in the ACMized model (top).}
    \label{fig:audio_load}
\end{figure*}

In Fig. \ref{fig:audio_load} we present the average computational load of each token displayed along with the waveform for additional samples from the validation dataset. We can see that in general the model allocates more learners to tokens corresponding to speech, whereas it keeps a low budget for tokens corresponding to silence or background noise.

\section{Contributions}
\label{app:contrib}

Bartosz set the research direction of the project by proposing the ACM architecture, the ACM weight initialization scheme, and the $\mathcal{L}_{\text{b}}$ and $\mathcal{L}_{\text{d}}$ auxiliary losses. He wrote the shared codebase for the experiments, carried out the computer vision experiments, and implemented the fast Triton-based efficient implementation of the method. He also performed all the analyses in the paper with the exception of the audio computational load analysis. He played the primary role in the writing and editing of the article.\\

\noindent Alessio implemented and ran the audio experiments and audio computational load analysis, and contributed to the writing and editing of the article. \\

\noindent Karol wrote the initial re-implementation of the A-ViT method. \\

\noindent Pasquale provided valuable advice on gradients estimators for k-subset sampling. He also contributed to the writing and editing of the article. \\

\noindent Simone contributed significantly to this work by proposing the auxiliary loss $\mathcal{L}_{\text{e}}$ and authoring the majority of the related work section. Additionally, he assisted in writing and revising the manuscript and significantly improved the visual quality of the figures.

\clearpage

\bibliography{refs}